\documentclass[journal]{IEEEtran}
\usepackage{tabularx,booktabs}

\usepackage{ifpdf}
\usepackage{algorithmic}
\usepackage{algorithm}

\usepackage{array}
\usepackage{multirow}
\usepackage{longtable}
\usepackage{rotating}
\usepackage[justification=centering]{caption}
\usepackage{array} 
\usepackage{graphicx}
\newcommand\sbullet[1][.5]{\mathbin{\vcenter{\hbox{\scalebox{#1}{$\bullet$}}}}}
\makeatletter
\let\NAT@parse\undefined
\makeatother

\usepackage[english]{babel}
\usepackage{amsmath}
\usepackage{graphicx}
\usepackage{xcolor}
\usepackage{enumitem}
\usepackage{threeparttable}
 \usepackage{longtable}
\usepackage{subfloat} 
\usepackage{subfig} 
\usepackage{url}
\usepackage{amssymb}
\usepackage{pifont}

\newcommand{\subsubsubsection}[1]{\paragraph{#1}\mbox{}\\}
\newcommand{\cmark}{\ding{51}}%
\newcommand{\xmark}{\ding{55}}%
\begin{document}
\title{Adversarial Attacks and Defenses in Machine Learning-Powered Networks: A Contemporary Survey}

\author{
	Yulong Wang, \emph{Member}, \emph{IEEE},
	Tong Sun, 
	Shenghong Li, \emph{Member}, \emph{IEEE},
	Xin Yuan, \emph{Member}, \emph{IEEE},\\
	Wei Ni, \emph{Senior Member}, \emph{IEEE},
 Ekram Hossain, \emph{Fellow}, \emph{IEEE}, 
    and H. Vincent Poor, \emph{Life Fellow}, \emph{IEEE}
	\thanks{Y.~Wang and T.~Sun are with the School of Computer Science, Beijing University of Posts and Telecommunications, Beijing 100876, China (e-mail: \{wyl, suntong\}@bupt.edu.cn).}
	\thanks{S.~Li, X.~Yuan, and W.~Ni are with the Commonwealth Science and Industrial Research Organisation (CSIRO), Marsfield, Sydney, New South Wales, 2122, Australia (e-mail: \{shenghong.li, xin.yuan, wei.ni\}@data61.csiro.au)}
    \thanks{E. Hossain is with the Department of Electrical and Computer Engineering, University of Manitoba, Canada (e-mail: ekram.hossain@umanitoba.ca).}
    \thanks{H. V. Poor is with the Department of Electrical and Computer Engineering, Princeton University, Princeton, NJ 08544, USA (e-mail: poor@princeton.edu).}
}

\maketitle
\begin{abstract}
Adversarial attacks and defenses in machine learning and deep neural network have been gaining significant attention due to the rapidly growing applications of deep learning in the Internet and relevant scenarios. This survey provides a comprehensive overview of the recent advancements in the field of adversarial attack and defense techniques, with a focus on deep neural network-based classification models. Specifically, we conduct a comprehensive classification of recent adversarial attack methods and state-of-the-art adversarial defense techniques based on attack principles, and present them in visually appealing tables and tree diagrams. This is based on a rigorous evaluation of the existing works, including an analysis of their strengths and limitations. We also categorize the methods into counter-attack detection and robustness enhancement, with a specific focus on regularization-based methods for enhancing robustness. New avenues of attack are also explored, including search-based, decision-based, drop-based, and physical-world attacks, and a hierarchical classification of the latest defense methods is provided, highlighting the challenges of balancing training costs with performance, maintaining clean accuracy, overcoming the effect of gradient masking, and ensuring method transferability. At last, the lessons learned and open challenges are summarized with future research opportunities recommended.
\end{abstract}



\IEEEpeerreviewmaketitle

\begin{table}[!htbp]
\begin{tabular*}{\linewidth}{ll}
\toprule
\textbf{Abbreviation} & \textbf{Full form} \\ \midrule
AAAM & Adversarial Attack Attention Module \\
adMRL & adversarial MRL\\

AdvCam & Adversarial Camouflage \\

AdvLB & Adversarial Laser Beam \\

AdvRush & Adversarially robust architecture Rush \\

AGKD-BML & Attention Guided Knowledge Distillation and \\
 & Bi-directional Metric Learning \\

AI & Artificial Intelligence\\

AL & Adversarial Learning\\

AMM & Adversarial Margin Maximization network \\

ANP & Adversarial Noise Propagation \\

AoA & Attack on Attention\\

APE-GAN & Adversarial Perturbation Elimination GAN\\

APR & Amplitude-Phase Recombination \\

ART & Adaptive ReTraining \\

ASR & Attack Success Rate \\

AT&Adversarial Training\\

AtkSE & Attacking by Shrinking Errors \\

ATTA & Adversarial Transformation-enhanced \\
 & Transfer Attack \\
 
AUC& Area Under Curve\\ 

BN & Batch Normalization \\

BIM & Basic Iterative Method\\

BLF & Bounded Logit Function \\

BFGS & Broyden Fletcher Goldfarb Shanno\\

CAFA& Class Activation Feature Loss\\

CAP-GAN & Cycle-consistent Attentional Purification GAN\\

CAT &  Contrastive Adversarial Training\\

C-BCE & Conditional Binary Cross-Entropy \\

CD&Constellation Diagram\\

CE& Cross-Entropy\\ 

CAFD &  Class Activation Feature-based Denoiser \\

CMAG & Cascade Model-Aware Generative\\

CNN & Convolutional Neural Network\\

COLT & COnvex Layer-wise Adversarial Training\\

CSA & Cost-Sensitive Adversarial learning model \\

CSE & Cost-Sensitive Adversarial Extension \\

CSM & Cross-Spectral Mapping \\

C\&W &  Carlini and Wagne attacking algorithm\\

DAC & Degree Assortativity Change \\

DGA & Domain Generating Algorithms \\

DH-AT & Dual Head Adversarial Training \\

DM& Data Manifold\\

DNN & Deep Neural Network\\

DSNGD & Dynamically Sampled Nonlocal Gradient Descent \\

D2Defend & Dual-Domain based Defense \\

DWT &  Discrete Wavelet Transform \\

ER-Classifier & Embedding Regularized Classifier \\

ERF & Effective Receptive Field \\

FeaCP & Feature-wise Convex Polytope attack\\

FGSM &  Fast Gradient Sign Method \\

FMR & Feature-Map Reconstructor\\

FNC & Feature Norm Clipping \\

FP& False Positive\\

FR & Face Recognition  \\

FS & Feature Scattering \\

GAE & Graph AutoEncoder\\

GAN & Generative Adversarial Network \\

GAT & Generative Adversarial Training \\

GEM & Graph Embedding Model \\

GF-Attack & Generalized adversarial attack Framework\\

GNN & Graphic Neural Network  \\

GR & Global Reconstructor \\

GraphAT & Graph Adversarial Training\\

H\&G &  Hendrycks and Gimpel\\

HAG & Hash Adversary Generation \\

HFT &  High Frequency Trading \\
HMC & Hamiltonian Monte Carlo \\

HMCAM & Hamiltonian Monte Carlo with Accumulated \\
& Momentum \\

HSI & Hyper-Spectral Image \\

IBA & Iterative Black-box Attack\\

ICAT & Induced Class Adversarial Training \\

IGA & Iterative Gradient Attack \\

IoT & Internet-of-Things \\

\bottomrule
\end{tabular*}
\end{table}

\begin{table}[ht!]
\begin{tabular}{ll}
\toprule
\textbf{Abbreviation} & \textbf{Full form} \\ \midrule
IoU & Intersection over Union \\

IPW & Iterative Partially-White-box subspace attack\\

JSMA & Jacobian-based Saliency Map Attack \\

KD & Knowledge Distillation \\
KL & Kullback Leibler divergence\\  

LAFEAT & LAtent FEAture Attack \\

L-BFGS & Limited-memory BFGS \\

LF & Low-Frequency component distortion \\

LID& Local Intrinsic Dimensionality\\

LPIPS & Learned Perceptual Image Patch Similarity \\

LR &  Logit Reconstructor \\

LSTM & Long Short-Term Memory \\

MAE& Mean Absolute Error\\

MAP & Multispectral Adversarial Patch \\

ME& Material Emissivity\\

MI-FGSM & Momentum Iterative Fast Gradient Sign Method \\

ML & Machine Learning \\

MRL & Meta Reinforcement Learning \\

MultiD-WGAN & Muti-Discriminator Wasserstein GAN \\

N-BaIoT & Network-Based detection of IoT\\

NAS & Neural Architecture Search \\

NAttack & NES adversarial Attack\\

NLM & Non-Local spatial smoothing \\

NLP & Natural Language Processing \\

NSGA-PSO & Non-dominated Sorting Genetic Algorithm \\
  & with Particle Swarm Optimization \\
  
NSS & Normalized Scanpath Saliency (minus)\\ 

OOD& Out Of Domain inputs\\ 

PCA & Principal Component Analysis \\

PCAE & Principal Component Adversarial Example \\ 

PCL& Prototype Conformity Loss\\

PDA &  Progressive Diversified Augmentation \\

PEC &  Polyhedral Envelope Certifier \\

PER &  Polyhedral Envelope Regularization \\

PGD & Projected Gradient Descent \\

PLC &  Pearson’s Linear Coefficient (minus)\\ 

PR& Precision-Recall\\

PS-GAN & Perceptual-Sensitive GAN \\

RE & Reconstruction Error  \\

RMS  & Root Mean Square\\

ROC& Receiver Operating characteristic Curve\\

ROSA & RObust SAliency \\

RP2 & Robust Physical Perturbation \\

RRF & Rectified Reverse Function \\

RSLAD & Robust Soft Label Adversarial Distillation \\

RMSE & Root Mean Squared Error\\

SACNet  & Self-Attention Context Network \\

SAD & Saliency Adversarial Defense\\

SCA & Side Channel Attack \\

SCE & Softmax Cross-Entropy \\

SGA & Simplified Gradient-based Attack \\

SGADV & Similarity-based Gray-box Adversarial Attack \\

SML & Single-directional Metric Learning \\

SNN & Spiking Neural Network\\

SNS & Sensitive Neuron Stabilizing \\

SOTA & State Of The Arts \\

SPSA & Simultaneous Perturbation Stochastic Approximation \\

SRLIM & Surrogate Representation Learning with Isometric Mapping \\

SSAH & Semantic Similarity Attack on High-frequency components\\

SSIM& Structural Similarity\\

STDB & Spike-Timing-Dependent Backpropagation\\
 
STFT &  Short-Time Fourier Transform \\

TP& True Positive\\

TriATNE & Tripartite Adversarial Training for Network Embeddings\\

UAP & Universal Adversarial Perturbation\\

WaveCNet & Wavelet-integrated Convolutional Network \\

WD&Wasserstein Distance \\
\bottomrule
\end{tabular}
\end{table}

\section{Introduction}

Deep neural networks (DNNs) are a crucial component of the artificial intelligence (AI) landscape due to their ability to perform complex tasks, object detection~\cite{DBLP:journals/access/HeLH23, DBLP:conf/dsmlai/KaurS21a}, image classification~\cite{DBLP:journals/apin/XuGLBLZ23, DBLP:conf/icitee/Liu21}, language translation~\cite{DBLP:conf/acl/AngelovaA022, DBLP:journals/taccess/Ananthanarayana21}, and many more~\cite{DBLP:journals/air/ChanBLPC23,DBLP:journals/apin/SunTDB23, DBLP:journals/csur/AbdullahA23}. The availability of advanced hardware, such as GPUs, TPUs, and NPUs, has facilitated the training of DNNs and made them a popular research direction in AI~\cite{DBLP:conf/5gwf/KelkarD21, 10.1145/3487890}. However, despite their strong learning ability, DNNs are susceptible to adversarial attacks, such as classical attack method Projected Gradient Descent (PGD)~\cite{DBLP:conf/iclr/MadryMSTV18}, Square attack~\cite{DBLP:conf/eccv/AndriushchenkoC20} or C\&W~\cite{DBLP:conf/sp/Carlini017}.
These attacks exploit the model's sensitivity to small and carefully crafted perturbations in the input data, causing the DNN to produce false predictions. Adversarial attacks represent a serious challenge to the robustness of DNNs and require proactive attention and action to mitigate the risks they pose.

\begin{figure} 
	\centering
	\subfloat[In digital world~\cite{DBLP:conf/ijcnn/YaoG21}]{ \includegraphics[width=3.2in]{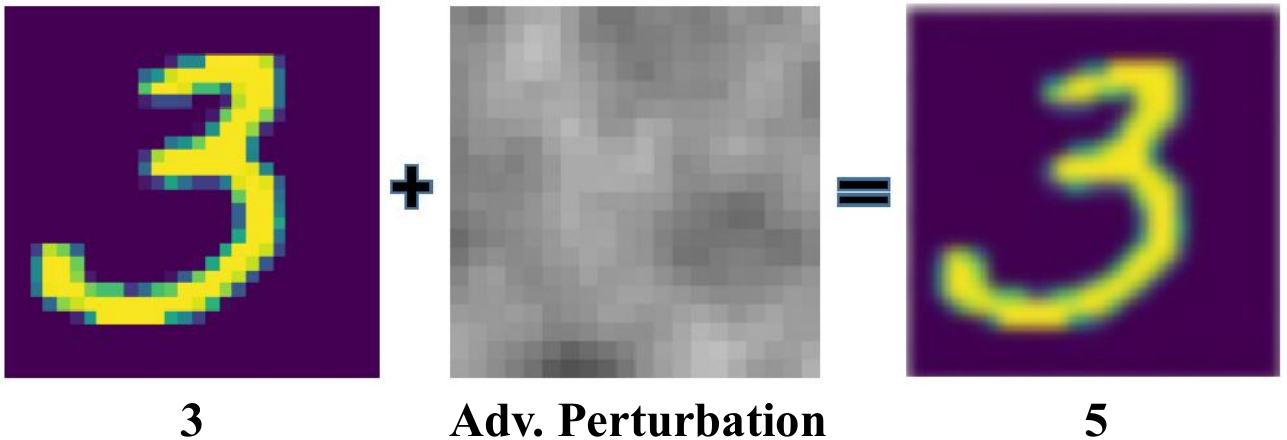} \label{fig:subfig:digital}}
	\quad  
	\subfloat[In physical world~\cite{DBLP:conf/cvpr/DuanMQCYHY21}]{ \includegraphics[width=3.2in]{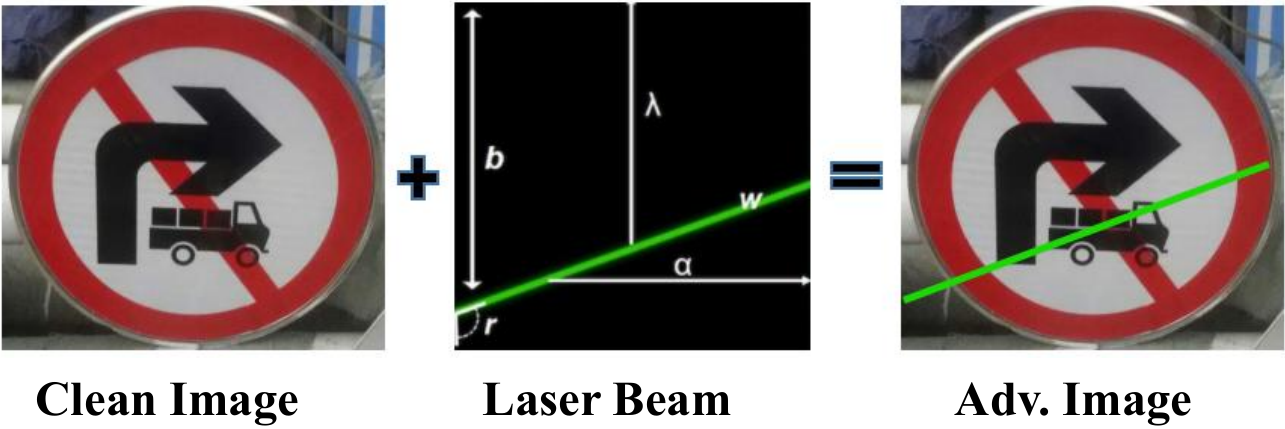} \label{fig:physical}}
    \quad  
    \subfloat[In cyberspace~\cite{9918564}]{ \includegraphics[width=3.2in]{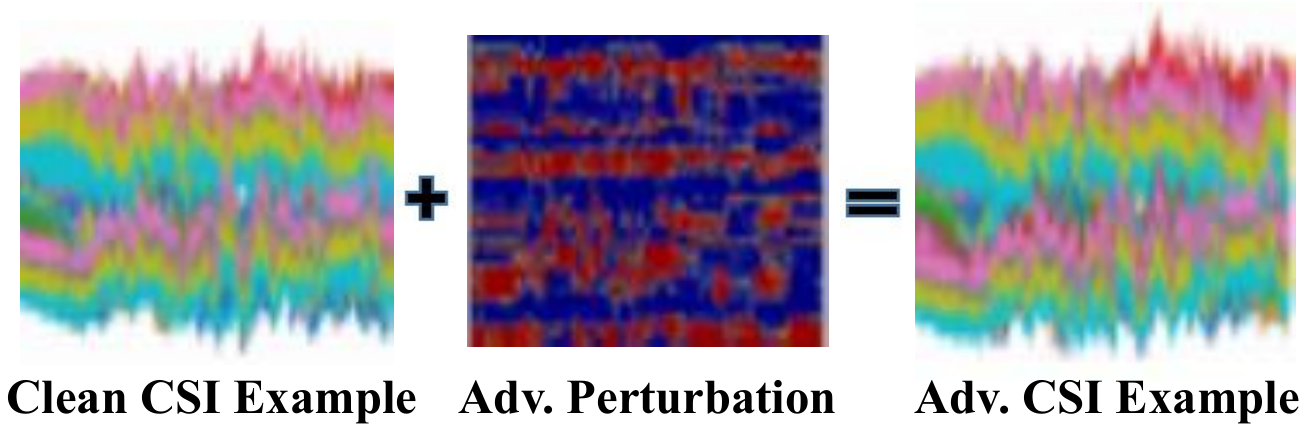} \label{fig:cyberspace}}
	\caption{Examples of adversarial perturbation}
	\label{examples}
\end{figure}

Adversarial attacks in deep learning can have serious consequences, as captured in many recent studies. For example, Fig.~\ref{examples} illustrates various such attacks, including a deliberately devised alteration to an input image resulting in misclassification by a convolutional neural network (CNN) with a 99\% level of certainty~\cite{DBLP:conf/ijcnn/YaoG21}, a traffic sign recognition attack that uses a laser beam to fool self-driving cars~\cite{DBLP:conf/cvpr/DuanMQCYHY21}, and a channel state information (CSI) recognition attack in an Internet-of-Things (IoT) scenario, where CSI examples are adversarially perturbed to mislead DNN models~\cite{9918564}.
To this end, adversarial attacks can pose significant risks and impacts on many important areas with deep learning involved, as articulated in the following.

\subsection{Impacted Areas}
Adversarial attacks can pose significant risks and impacts on network and communications~\cite{9609969,9816059,9734753}, computer vision~\cite{DBLP:journals/csur/MirskyL21, DBLP:conf/ivsp/YinUD21}, autonomous driving~\cite{DBLP:conf/cvpr/DuanMQCYHY21, DBLP:journals/tcps/JiangHZPA21}, finance and e-commerce~\cite{DBLP:conf/kdd/FursovMKKRGBK0B21, 10.1145/3440084.3441213}, human-machine interaction~\cite{2019A, DBLP:conf/chi/ParkAHL21}, and many other areas where deep learning can be involved.

{\color{black}
\subsubsection{Network and Communications}
With the widespread adoption of DNN models in the network and communications domain, the impact of adversarial attacks cannot be overlooked. To facilitate signal detection and fast tracking, DNN-based wireless signal classifiers have been used by wireless signal receivers to classify over-the-air received signals into different modulation schemes and orders. It is shown in \cite{9609969} that these models are vulnerable to channel-aware adversarial attacks, where an adversary can transmit an adversarial perturbation, given a power budget, to mislead the receiver into making erroneous predictions when classifying superposed signals and adversarial perturbations. 
Moreover, adversarial attacks can substantially diminish the performance of DNN models. used for modulation scheme recognition in communication systems~\cite{9259112}. 
A study in~\cite{9893902} shows that adversarial attacks also significantly increase the bit error rate (BER) of DNN-based OFDM detectors that are trained to recover the payload bits directly from received symbols. Furthermore, generative adversarial networks (GANs), previously used for generating adversarial image examples~\cite{DBLP:conf/ijcai/XiaoLZHLS18}, have now been adapted to produce adversarial examples that attack modulation classifiers in wireless communications~\cite{9756577} or conduct wireless signal spoofing~\cite{9144305}. The feasibility of adversarial attacks has also been evaluated in 5G~\cite{9816059} and 6G networks~\cite{9390408}. Moreover, Huang \textit{et. al}~\cite{9531421} showed that adversarially contaminated Wi-Fi signals can mislead DNN-based non-intrusive human activity recognition systems. Xu \textit{et al.}~\cite{9918564} constructed adversarial perturbation by customizing Fast Gradient Sign Method (FGSM)~\cite{DBLP:journals/corr/GoodfellowSS14} and PGD~\cite{DBLP:conf/iclr/MadryMSTV18}, and reduced the performance of Wi-Fi sensing applications, such as  user identification, gesture recognition, and human activity recognition. With tiny perturbation-to-signal ratios of around -18 dB in CSI-based Wi-Fi fingerprinting, adversarial attacks can reach an extraordinary success rate of over 90\%~\cite{10015738}. Adversarial defense schemes for communication networks are receiving increasing attention due to the severe threats posed by transferred adversarial attacks. Defense schemes have been proposed to counteract adversarial examples in radio signals~\cite{9477416}, radio signal classification~\cite{9887932}, and modulation classification~\cite{9734753, 9542973, 9916315}.

Adversarial attacks have also been observed in networking and network management, due to the increasing application of DNN-based network traffic classification for network traffic management, policy enforcement, and intrusion detection systems. 
Universal Adversarial Perturbation (UAP) \cite{DBLP:conf/cvpr/Moosavi-Dezfooli17}, originally developed for adversarial attacks against DNN-based image classifiers, has been evaluated for attacking DNN-based network traffic classification~\cite{9328496}. 
DNN models trained 
using collected TCP/IP traffic data can be fooled by adversarially perturbed network packets sent from the host controlled by an attacker. 
Nowroozi \textit{et al.}~\cite{9747933} show that adversarial attacks developed against image classification models, such as Jacobian-based Saliency Map (JSMA) \cite{DBLP:conf/eurosp/PapernotMJFCS16}, Iterative Fast Gradient Method (I-FGSM) \cite{DBLP:conf/iclr/KurakinGB17}, PGD~\cite{DBLP:conf/iclr/MadryMSTV18}, Limited-memory Broyden Fletcher Goldfarb Shanno (L-BFGS) \cite{DBLP:journals/corr/SzegedyZSBEGF13}, and DeepFool attack~\cite{DBLP:conf/cvpr/Moosavi-Dezfooli16}, can be used to attack CNN models trained on well-known computer network datasets, including the Domain Generating Algorithms (DGA)
dataset, the Network-based Detection of IoT
(N-BaIoT) dataset, and the RIPE Atlas dataset, with attack success rates ranging from 63\% to 100\%. Zhang \textit{et al.}~\cite{9674195} unveil that adversarial attacks can be adapted to the network layer, where adversarially perturbed network traffic can evade a network intrusion detection system (NIDS) with up to a 35.7\% success rate. 
An initial investigation of defenses against a DNN-based NIDS is presented in~\cite{9448103}.
}

\subsubsection{Computer Vision}
{\color{black}
This is an area conventionally susceptible to adversarial attacks, where adversarial attacks involve adding subtle disturbances to input samples that are undetectable to humans.} 
For example, face recognition systems can be deceived by adversarial examples, where slight modifications to a person's face can trick the system into thinking it is someone else~\cite{DBLP:journals/csur/MirskyL21, DBLP:conf/www/TariqJW22,DBLP:journals/tissec/QinPLRB22}.
Object detection systems can be misled by adversarial examples, where a slight alteration to the input image can cause the system to miss detecting an object or misclassify it as a different object~\cite{DBLP:journals/access/IlioudiDWS22, DBLP:conf/kdd/LiuGJ0D21}. 
These disturbances cause a DNN model to output an incorrect classification result with a high degree of certainty, allowing the model to misclassify a sample without over-modifying it. The performance of a DNN model in real-world scenarios cannot be solely evaluated based on its ability to correctly classify benign samples. It is also crucial that the model is robust against negative samples or adversarial examples, which have the potential to trick the model into making incorrect predictions.

An increasingly crucial computer vision application of DNNs is autonomous driving, which is susceptible to adversarial attacks, where specifically-designed images can cause the perception model to ``ignore'' important features of the environment, such as roadblocks and pedestrians. Adversarial samples can also interfere with the recognition systems of autonomous vehicles, causing them to output false information and activate wipers. In one recent incident, an autonomous car failed to identify ``trolleybus'' and ``road signs'' when its camera was hit by a laser beam~\cite{DBLP:conf/cvpr/DuanMQCYHY21, DBLP:conf/icons2/OrthSPD22}. 
Another example is that placing adversarial sample stickers on the road could cause a vehicle in autopilot mode to merge into the opposite lane~\cite{DBLP:journals/iotj/YangLZLT21, DBLP:conf/raid/MaLW0H0L22}.

\subsubsection{Human-Machine Interaction}
Another area where adversarial attack concerns is human-computer interaction, where the increasing use of speech and images as input methods, particularly on mobile devices, has made these technologies widely accepted and popular~\cite{DBLP:journals/access/XuWZC22, DBLP:conf/chi/WadleyKKSWCGHMS22, DBLP:conf/hai/NalepkaCKKPR22}. The precision of speech and image object identification is key to the machine's capability to comprehend and carry out users' requests, making it vulnerable to exploitation by attackers who can modify the data slightly to cause the machine to make incorrect judgments and actions~\cite{2019A, DBLP:conf/ACMdis/Alves-OliveiraL21}.

\subsubsection{Finance and E-Commerce}
Deep learning algorithms have also been applied by financial institutions. As a consequence, the finance industry is at risk from adversarial attacks. For instance, fraud detection systems in finance can be deceived by adversarial examples, where small perturbations are added to transactions to evade detection by the system~\cite{2022Credit, DBLP:conf/icbct/ZhangHZ21}. Algorithmic trading systems can be misled by adversarial examples in the stock market, where minor alterations to the input can result in the system making incorrect predictions and trade decisions~\cite{2022Stealthy, DBLP:conf/icaif/ShearerBBW21}. Moreover, credit scoring systems can be tricked by adversarial examples, where small modifications to the input data can cause the system to provide an incorrect credit score~\cite{2020Managing, DBLP:conf/pris2/ChenJT22}. These examples highlight the importance of developing robust and secure financial systems that are resistant to adversarial attacks.

All these examples demonstrate how adversarial samples, which are barely noticeable to humans but detectable by machines, can deceive deep learning systems. 
Other areas that can be impacted by adversarial attacks are healthcare, security and surveillance, as well as other security systems for making incorrect decisions, natural language processing (NLP), and robotics, 
where adversarial examples can trick medical image analysis systems into misdiagnosing diseases or medical conditions~\cite{DBLP:journals/algorithms/KogaT22, DBLP:conf/niss/BengagBM21}; trick language translation systems, sentiment analysis systems, and other NLP systems into making incorrect predictions~\cite{DBLP:conf/aaai/Swenor22, 9557814}; and deceive robotics systems, leading to incorrect control decisions~\cite{DBLP:journals/access/TasoojiM22, DBLP:conf/atal/Youssef21}.
These attacks are not limited to laboratory settings, and can occur in real-world applications and systems.

\subsection{Attack Scenario}
Adversarial attacks can occur during a model inference stage. Specifically, an attacker can aim to deceive a DNN-based model, e.g., an image classifier, by launching a two-phase attack: Generating an adversarial example from the DNN-based image classifier and feeding it back into the image classifier.

During the first stage, the attacker perturbs the pixel values of a benign example to maximize the loss function value of the image classifier. This forces the image classifier to misclassify the adversarially perturbed example or minimize the loss function value with regards to an incorrect class designed by the attacker. The attacker can employ different strategies to guide the direction of perturbation based on their \textit{a-priori} knowledge of the DNN model under attack. If the neural network architecture, learned parameter values (weights and biases), and the loss function of the DNN model is available (e.g., due to a compromised server or a rogue employee), the attacker can exploit a gradient-based attacking algorithm to calculate the perturbation and produce the adversarial example. 

For instance, the Fast Gradient Sign Method (FGSM)~\cite{DBLP:journals/corr/GoodfellowSS14} generates an adversarial instance, denoted by $\mathbf{x}^{adv}$, by applying the following rule:
\begin{align}
\mathbf{x}^{adv} = \mathbf{x} + \epsilon\times \text{sign} (\nabla_{\mathbf{x}} \mathcal{L}( \mathbf{x}, y)), \nonumber
\end{align}
where $\mathbf{x}$ is the input data, $\epsilon \in \mathbb{R}^+$ is the perturbation magnitude, $y$ indicates the ground-truth class, $\text{sign}(\cdot)$ returns the sign of a real value, and $\nabla_{\mathbf{x}} \mathcal{L}(\mathbf{x}, y)$ indicates the gradient of the loss function $\mathcal{L}(\mathbf{x}, y)$ with regard to the input example~$\mathbf{x}$. 

To produce an effective adversarial example within a perturbation budget, the attacker can progressively perturb the elements of an example (e.g., the pixels of an image), by running more sophisticated adversarial attack algorithms, e.g., Projected Gradient Descent (PGD)~\cite{DBLP:conf/iclr/MadryMSTV18}. The attacker repeats this process until the DNN-based classifier misclassifies the example $\mathbf{x}$ into an attacker-specified target class that differs from the source category of the benign example~$\mathbf{x}$.

The information necessary to undertake an adversarial attack against a DNN is relatively easy to obtain. This is due to the fact that the best-performing DNN-based classifiers generally use well-known architectures, e.g., ResNet models~\cite{8804390}, and commonly employ the Cross-Entropy loss function~\cite{9409715} for classification tasks. Even in the case where the parameters of the DNN-based classifiers are not accessible for the attacker, it is possible for the attacker to learn a good surrogate of the DNN-based classifiers by sending queries to the classifiers and collecting responses~\cite{9252132}.

The attacker can also use other strategies, such as constrained optimization-based or heuristic approaches (see Section~\ref{attack}), to seek out effective adversarial instances. After the attacker confirms the effectiveness of the generated adversarial example in a controlled environment, they can launch an actual adversarial attack by feeding perturbed examples to the DNN model under attack.

\subsection{Notable Attack Incidents}

In the past several years, there have been several notable adversarial example attacks:
\begin{itemize}

 \item 
In 2019, researchers demonstrated that they could cause a machine learning (ML) model to misclassify a fraudulent credit card transaction as legitimate by applying weak perturbations to the transaction data~\cite{2019An, DBLP:conf/acmturc/ChenZLYMW19,DBLP:journals/pvldb/CaoYCZLQ19}. This type of attack could potentially result in significant financial losses for financial institutions and consumers. 
 \item 
In 2020, researchers demonstrated that they could cause a chatbot to generate inappropriate or offensive responses by adding small perturbations to the input text~\cite{DBLP:conf/mc/FeineMM20}. This type of attack could potentially cause damage to a company's reputation or lead to lost customers. 
 \item 
In 2021 and 2022, researchers demonstrated that they could cause an autonomous vehicle to mistake a stop traffic sign for a speed limit sign by putting a small, almost imperceptible sticker on the sign~\cite{DBLP:journals/eswa/BadueGCACFJBPMV21, DBLP:conf/cscs2/ForsterBOS22}. While this type of attack has not yet resulted in any real-world accidents, it has raised concerns about the safety of autonomous vehicles and the potential for malicious actors to cause accidents or other harm using adversarial attacks.
 \item 
In 2022, a group of researchers demonstrated that adversarial examples could be utilized to evade spam filters, allowing malicious emails to bypass detection~\cite{DBLP:journals/access/SalloumGVS22, DBLP:conf/codaspy/QachfarVM22}. They created adversarial examples of spam emails by adding perturbations to the email content, and caused the spam filter to incorrectly classify the email as non-spam. 
 \item 
In 2022, a team of researchers showed that adversarial examples could trick image classification systems in self-driving cars~\cite{DBLP:journals/tcad/SunYRYH22}. The researchers showed that a small perturbation added to a traffic sign could cause the self-driving car to misidentify the sign, potentially leading to dangerous mistakes on the road. 
 \item 
In 2022, researchers demonstrated that adversarial examples could trick a voice assistant, e.g., Amazon Alexa and Google Assistant~\cite{DBLP:conf/cui/SeymourCS22}. They manipulated voice commands to make them sound normal to humans but caused voice assistants to perform actions that were not intended. This can sabotage the security and privacy of users who use voices to control smart home devices. 

\end{itemize}

While these adversarial attacks on ML or DNN models have not yet caused widespread financial or economic loss, they have sparked worries about safety and dependability of these systems, and research into approaches for detecting and defending against these attacks is ongoing.

\subsection{Contributions of This Survey}

As the applications of machine learning and artificial intelligence continue to expand into almost all aspects of human life and society, the robustness and security of machine learning models become increasingly crucial~\cite{2018A, DBLP:conf/icccv/DomyatiM22, DBLP:journals/tecs/MorrisEKIAS22}. 
As a result, adversarial attacks and defenses make up an active and rapidly growing research area. For example, in the Year 2022 alone, over 1,200 research articles were published on adversarial attacks and defenses, documenting many new attack and defense techniques, and incidents. In the Year 2021, over 1,000 research articles were published on these topics\footnote{These search results are based on IEEE Xplore with the keyword ``adversarial'' and ``attack'' as of 7 February 2023.}. 
Most of these new research outcomes have not been covered by any of the latest literature reviews and surveys due to the fast pace of this active research area of adversarial {\color{black}attacks and defenses}. To this end, a timely summary of emerging attacks and new defense techniques is critical to keep the research community and security practitioners well-informed and equipped with the latest knowledge. 
 
In this comprehensive survey, we delve into the cutting-edge advancements in adversarial attack and defense techniques over the last 24 months, including over 180 quality research papers published in IEEE and ACM journals and top conferences, such as CVPR, ICCV, and AAAI, since 2021. With a particular emphasis on DNN-based image classification models, which are the widely used deep learning models and a primary target for {\color{black}adversarial attack and defense} research{\color{black}, our goal} is to offer a comprehensive overview of the latest breakthroughs and pave the way for future advancements in the arena of {\color{black}adversarial attack and defense}. This survey is also expected to serve as a catalyst for the next wave of research and innovation in the rapidly evolving field of {\color{black}adversarial attack and defense}.

These are the key findings of this survey:
\begin{itemize}
\item A comprehensive classification of recent adversarial attack methods as well as the SOTA adversarial defense techniques based on a variety of attack principles, presented in a visually appealing table and tree diagram format.

\item The categorization of the methods into counter-attack detection and robustness enhancement, with a specific focus on regularization-based methods for enhancing robustness, represented through tables and tree diagrams.

\item A rigorous evaluation of the existing works, including an analysis of their strengths and limitations, and recommendations for future research avenues.
\end{itemize}

Some of the noteworthy highlights from the survey include:
\begin{itemize}
\item An exploration of new avenues of attack in the last two years, including search-based attacks, decision-based attacks, drop-based attacks, and beyond the traditional optimization-based and gradient-based attacks.

\item The emergence of physical-world adversarial attacks, particularly in the form of adversarial patches.

\item A hierarchical classification of the latest defense methods, highlighting the challenges of balancing training costs with performance, maintaining clean accuracy, overcoming the effect of gradient masking\footnote{Gradient masking here refers to the phenomenon that the gradient of the model is hidden or obsolete, e.g., towards potential adversaries. On the other hand, it also refers to a category of defense techniques that exploit or aim to achieve the phenomenon of gradient masking~\cite{DBLP:conf/ccs/PapernotMGJCS17}.} (or in other words, a defense method appears to work but is actually ineffective), and ensuring method transferability.
\end{itemize}

\begin{figure*}
    \centering
    \includegraphics[height=5in]{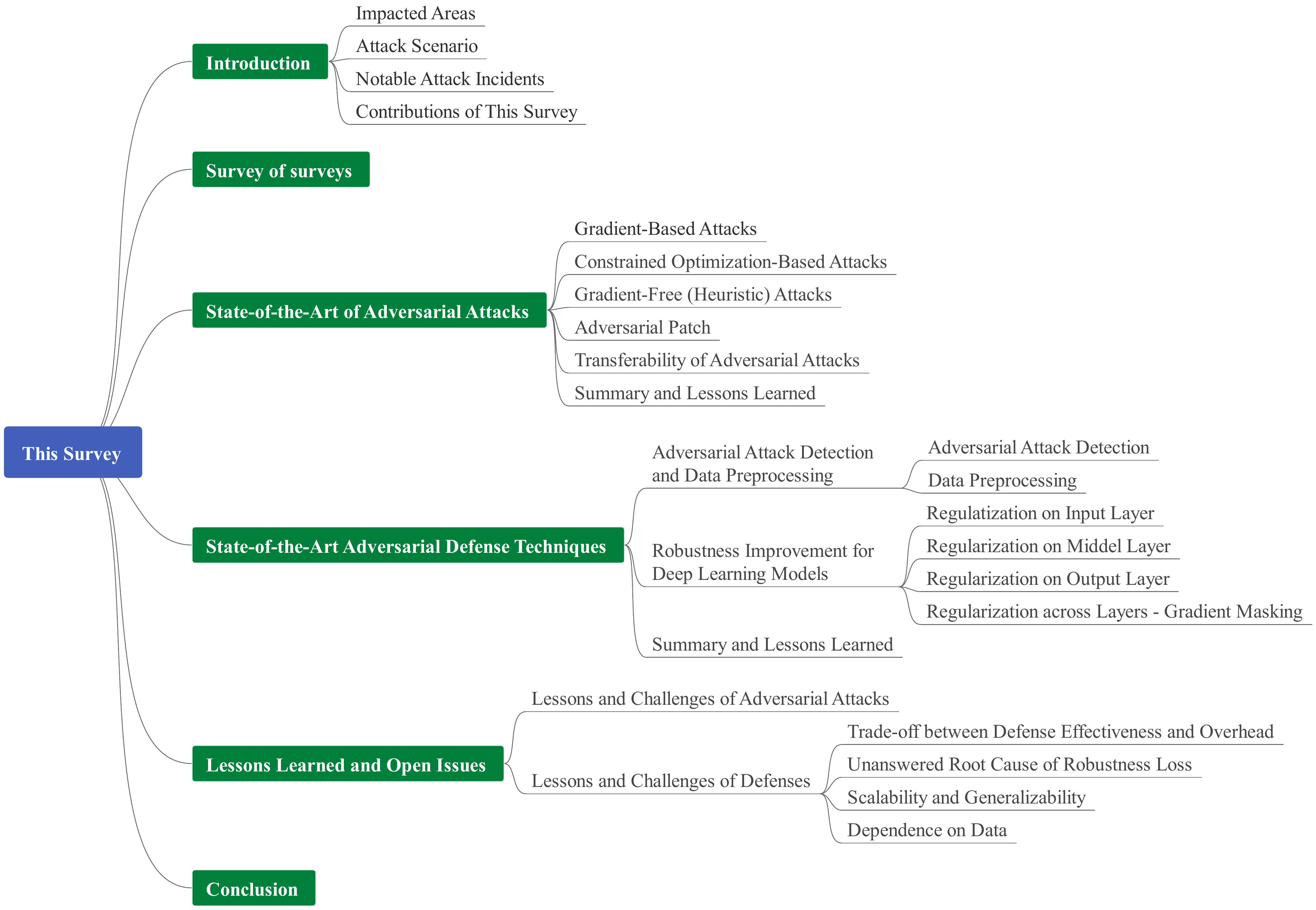}
    \caption{{\color{black}The anatomy of this survey.}} 
    \label{fig: paper structure}
\end{figure*}

As illustrated in Fig.~\ref{fig: paper structure}, this survey is organized as follows. Section \ref{sec: survey of surveys} provides a brief overview of the existing surveys of adversarial attacks and defenses, and clarifies the key differentiating factors of the current survey. 
Section~\ref{attack} categorizes the most recent adversarial attacks, and provides a comprehensive analysis of each of the categories, {\color{black}as well as the transferability of adversarial attacks}. Section~\ref{defense} classifies and analyzes various adversarial defense and detection techniques, and their effectiveness and limitations against the latest adversarial attacks.
Lessons learned and open challenges are delineated in Section~\ref{future}.
At last, Section~\ref{conclusion} summarizes the current state and suggests avenues for future investigations. Table~\ref{tab:notation_def_others} defines the notation used in this survey. 

\begin{table}[!t]
\caption{Notation and definition}
\label{tab:notation_def_others}
\begin{tabularx}{8.75cm}{lX}
\hline
\textbf{Notation} & \textbf{Definition} \\
\hline
$\mathbf{x}$, $y$ & A clean input example and its ground-truth label\\
$y_t$ & the label of the class designated by an attacker.\\
$\mathbf{x}^{adv}$, $t$ & An adversarially perturbed example and the attacker-specified target class \\
$\mathbf{x}^{(i)}$ & An adversarial example generated in $i$-th round of processing\\
$\tilde{\mathbf{x}}$ & The optimal state in the Markov decision process tackled by reinforcement learning\\
$\epsilon \in \mathbb{R}^+$ & The magnitude of perturbation applied to an example\\
$\nabla_{\mathbf{x}} \mathcal{L}( \cdot, \cdot)$ & The gradient of the loss function $\mathcal{L}( \cdot, \cdot)$ against $\mathbf{x}$\\
$\alpha$, $\alpha^{(i)}$ & A constant step size used for iterative adversarial example generation, and a variable step size used in the $i$-th iteration of generation, respectively \\ 
$\parallel \cdot \parallel_p$, $\ell_p$ & p-norm, where $p$ can be 0, 1, 2 or $\infty$.\\
$g_i$ & The gathered gradient in the $i$-th iteration\\
$z_i$ & The logit of example class $i$\\
$f(\cdot)$ & The predict function built from an ML or DNN model\\
$\text{Clip}_{\mathbf{s}}(\cdot)$ & The operation that clamps the elements of the input to within $\mathbf{s}$.\\
$\xi_i^\sigma$ & The random variables drawn i.i.d. from a distribution $P^{\sigma}$ parameterized by the standard deviation $\sigma \in \mathbb{R}^+$\\
$\eta$ & The neighboring hypothesis in IoU \\
$\mathcal{L}^{sce}$ & The softmax cross-entropy (SCE) discrepancy between the one-hot ground truth and the output\\ 
$\mathcal{L}_{T_i}(\cdot)$ & The loss function of a reinforcement learning task $T_i$\\
$\mathit{I}$ & The maximum number of gradient-update iterations \\
$\mathbf{z}^{(l)} $ &
The feature derived from the $l_{th}$ layer\\
$\bigtriangledown$ & The gradient calculation \\
$\odot$ & The Hadamard product \\
$S \subset \mathbb{R}^{n}$ & A subset of n-dimensional real number space.\\
\hline
\end{tabularx}
\end{table}

\section{Survey of Surveys}\label{sec: survey of surveys}
Adversarial attacks and defenses in ML and DNN models are crucial areas of research that have garnered significant attention in recent years. There are several reviews on the topic, each delving into specific aspects of the topic.
The authors of~\cite{DBLP:journals/cybersec/SunTZ18} first present preliminary knowledge concerning adversarial examples, and then contrast theoretical models of adversarial example attacks with actual instances of attacks. They also present existing examples of actual adversarial attacks.
The authors of~\cite{DBLP:journals/access/AkhtarM18} review adversarial attacks and defenses in the computer vision domain, as well as their real-world applications. They analyze the various methods proposed for attacking and defending against adversarial attacks in this domain and explore these methods' effectiveness and limitations. Similarly, the authors of~\cite{REN2020346} discuss the theoretical underpinnings, methods, and applications of adversarial attack techniques. In addition, they present several research initiatives on defensive strategies that span a broad variety of frontiers in the area, followed by a discussion of a number of open issues and challenges. The authors of~\cite{DBLP:journals/wicomm/KongXWHNL21} expand the scope of their review to include adversarial attacks in the context of images, malicious code, and text across various domains. They discuss the various types of adversarial attacks proposed in these contexts and analyze their effectiveness.
The authors of~\cite{app9050909} focus on summarizing the recent studies on adversarial attack and defense techniques in the deep learning area. They study existing defense methods from three perspectives: Data altercation, model modification, and utilization of auxiliary tools. They analyze the benefits and drawbacks of each strategy and discuss the limitations of existing methods.

There are also many investigations focusing on real-world attacks. In~\cite{DBLP:journals/corr/abs-2209-14262} and~\cite{DBLP:journals/corr/abs-2211-01671}, the authors focus on physical adversarial attacks. They classify and summarize current physical adversarial attacks from the perspective of physical world attacks, discussing the benefits and limitations of various approaches. In~\cite{DBLP:conf/kdd/FursovMKKRGBK0B21}, the authors examine adversarial attacks and defenses against transaction records from the viewpoint of NLP.

Other existing surveys are concerned with techniques for enhancing the robustness and resilience of DNN models in the face of adversarial attacks. For example, in~\cite{DBLP:journals/pr/QianHWZ22}, the authors analyze and compare adversarial training methods. They discuss the various approaches proposed for adversarial training and analyze their effectiveness in enhancing the robustness of deep learning models.

These above-mentioned earlier reviews, e.g.,~\cite{DBLP:journals/access/AkhtarM18,DBLP:journals/cybersec/SunTZ18}, focus more on classical attack and defense approaches. 
Conversely, with the advancement of deep learning in the last two years, more and more new risks have emerged. For example, 
Gallagher \textit{et al.}~\cite{DBLP:journals/compsec/GallagherPCPMK22} adapt FGSM as a single value and label flipping attack on financial stock data-based prediction networks, and find that it can result in a significant reduction in profitability and financial losses in a financial trading simulation. It is highlighted that the potential consequences of manipulating stock prices through buying and selling in the public trading market could be significant.
Goldblum \textit{et al.}~\cite{DBLP:conf/icaif/GoldblumSPG21} examine the impact of adversarial attacks on trading robot-based stock price predictions. These systems, known as high-frequency trading (HFT) systems, operate in extremely short time frames, making it difficult to prevent harmful behavior through human intervention. This is particularly concerning as it is well accepted that irregular behavior and security breaches in HFT systems have precipitated major market incidents like ``Flash Crash''~\cite{2017The}. The severity of the damage caused by adversarial attacks in such systems cannot be underestimated.

As summarized in Table~\ref{compare}, this survey aims to bridge the gaps in the current literature by not only focusing on classical methodological analysis but also systematically examining new methods that have emerged in the last two years. Specifically, it reviews the new methods in the context of adversarial attack methods, classifying them in a new light, and documenting effective but under-reported new attack methods, such as decision-based and drop-based methods. It also sheds light on the recent developments in adversarial patching, a powerful physical world attack that has been under-explored in previous surveys. From a defensive perspective, the survey covers both adversarial detection methods and model robustness enhancement methods, categorizing them from a novel perspective starting from the hierarchy of operations, summarizing them in the form of tables and tree diagrams, and suggesting future research directions. The survey provides an in-depth understanding of the SOTA in adversarial attacks and defenses in deep learning.

\begin{table*}
\caption{Comparison of the existing surveys on adversarial attacks, where ``n.a.'' stands for ``not applicable.'' }
\label{compare}
\resizebox{\linewidth}{!}{
\begin{tabular}{|c|c|c|c|c|c|c|c|} \hline
\multirow{2}{*}{\textbf{Reference}} 
& \multirow{2}{*}{\textbf{Year}} 
& \multicolumn{6}{c|}{\textbf{Focus Area}} \\ \cline{3-8}   
  &   & \textbf{Digital-world Attacks} & \textbf{Physical-world Attacks} & \textbf{Defenses} &  \textbf{Detection} & \textbf{Transferability} & \textbf{No. of references since 2021}\\
\hline
~\cite{DBLP:journals/access/AkhtarM18} & 2018 & \cmark &  \cmark & \cmark  & \cmark & \xmark& n.a. \\
\hline
~\cite{DBLP:journals/cybersec/SunTZ18} & 2018 &\cmark   & \cmark & \xmark& \xmark & \xmark &  n.a. \\
\hline
~\cite{app9050909} & 2019& \cmark & \xmark & \cmark & \cmark  & \cmark & n.a. \\
\hline
~\cite{REN2020346} & 2020 & \cmark & \cmark & \cmark & \cmark & \xmark & n.a. \\
\hline
~\cite{DBLP:conf/kdd/FursovMKKRGBK0B21} & 2021 & \cmark&\cmark &\cmark& \xmark &\xmark  &  n.a. \\
\hline
~\cite{DBLP:journals/wicomm/KongXWHNL21} & 2021& \cmark  & \cmark  & \cmark& \cmark & \xmark & 4\\
\hline
~\cite{article}  & 2022 & \cmark & \xmark & \cmark &\cmark  & \xmark  &  1 \\
\hline
~\cite{DBLP:journals/pr/QianHWZ22} & 2022 & \cmark & \xmark &\cmark & \xmark & \xmark & 5 \\
\hline
~\cite{Amirkhani2022ASO}& 2022&\cmark & \cmark & \cmark & \cmark & \xmark & 16 \\
\hline
~\cite{DBLP:journals/corr/abs-2211-01671} & 2022& \xmark & \cmark & \cmark & \cmark& \cmark& 38 \\
\hline
~\cite{DBLP:journals/corr/abs-2209-14262} & 2022& \xmark & \cmark &\xmark  &\xmark& \cmark& 42 \\
\hline
This survey & 2023  & \cmark & \cmark &\cmark & \cmark & \cmark & 183\\
\hline
\end{tabular}}
\end{table*}

\begin{table*}
	\caption{Brief summary of existing surveys on adversarial attacks.}
	\setlength{\tabcolsep}{3pt} 
	\renewcommand\arraystretch{1.5} 
	\begin{tabular}{|p{1.5cm}|p{0.8cm}|p{15.2cm}|}
\hline
\centering{\textbf{Survey}} & \textbf{Year} & \centering{\textbf{Focus Area}} \cr  \hline

This survey & 2023 & 
$\sbullet[.75]$ A comprehensive classification of recent adversarial attack methods and the SOTA adversarial defense
techniques based on various attack principles presented
in a visually appealing table and tree diagram format.

$\sbullet[.75]$   Category of the methods into adversarial attack
detection and robustness enhancement, with a specific
focus on regularization-based methods for enhancing robustness, represented through tables and tree diagrams.

$\sbullet[.75]$  A rigorous evaluation of the existing works, including an
analysis of their strengths and limitations, and recommendations for future research avenues.
\\\hline

~\cite{DBLP:journals/pr/QianHWZ22} & 2022& 
$\sbullet[.75]$  Introduce Robust adversarial training to defend against adversarial samples. 

$\sbullet[.75]$  Connections to traditional machine learning theories are investigated.

$\sbullet[.75]$  Different approaches with adversarial attacks and defense/training algorithms are summarised.

$\sbullet[.75]$  Present analysis, outlook and comments on adversarial training.\\\hline

~\cite{DBLP:journals/corr/abs-2209-14262} & 2022& 
$\sbullet[.75]$  Examine the evolution of physical adversarial attacks in computer vision applications using DNNs, e.g., image recognition, object detection, and semantic segmentation.

$\sbullet[.75]$  The survey classifies the present physical adversarial attacks, focusing on strategies employed to keep adversarial features in physical contexts.
\\\hline


~\cite{DBLP:journals/corr/abs-2211-01671} & 2022& 
$\sbullet[.75]$  Classify physical attacks in terms of attack task, attack form, and attack method. 

$\sbullet[.75]$  Classify adversarial defenses in terms of pre-processing, intra-processing and post-processing of DNN models.

$\sbullet[.75]$  Challenges of this research area are discussed, and present some future research directions.\\\hline

~\cite{article}  & 2022&
$\sbullet[.75]$  Overviews the fundamentals and features of adversarial attacks and evaluates recent adversarial instance-producing mechanisms. 

$\sbullet[.75]$  In-depth discussion of adversarial instance defense mechanisms from three aspects: Models, data, and networks. 

$\sbullet[.75]$  Challenges and outlooks in the field are presented within the context of the present development of adversarial instance generation and defense techniques.\\\hline

~\cite{Amirkhani2022ASO}& 2022& 
$\sbullet[.75]$  A comprehensive survey of the latest advancements in the adversarial robustness of models for object detection is provided. 

$\sbullet[.75]$  Prominent attack and defense approaches are reviewed, their advantages and disadvantages are discussed, 

$\sbullet[.75]$  A review of recent literature on adversarial robustness in the context of autonomous vehicles is conducted.\\\hline

~\cite{DBLP:journals/wicomm/KongXWHNL21} & 2021& 
$\sbullet[.75]$  Illustrate the importance of adversarial attacks, and outline the key ideas, categories, and dangers of adversarial attacks. 

$\sbullet[.75]$  Typical attack and defense techniques for various application areas are reviewed. 

$\sbullet[.75]$  Focus on images, text and malicious codes, present open questions, and conduct a comparison study with other relevant surveys.\\\hline

~\cite{DBLP:conf/kdd/FursovMKKRGBK0B21} & 2021 & 
$\sbullet[.75]$  Adversarial attacks and defenses against transaction record data are investigated. 

$\sbullet[.75]$ Analyses the distinct structure of transaction records compared to typical NLP or time series data and summarizes the characteristics of transaction records and their impact on adversarial attacks.

$\sbullet[.75]$  A scenario for black-box attacks is considered, focusing specifically on adding transaction tokens to the sequence's end. \\\hline

~\cite{REN2020346} & 2020 & 
$\sbullet[.75]$  Present the theoretical basis, algorithms, and practical uses of adversarial attack algorithms.

$\sbullet[.75]$  Considerable research efforts on defensive techniques are described, which cover a wide range of frontiers within the arena. 

$\sbullet[.75]$  Several challenges and open issues are discussed.\\\hline

~\cite{app9050909} & 2019& 
$\sbullet[.75]$  Explain adversarial attack approaches in both the training and testing phases.

$\sbullet[.75]$  Adversarial attack methods are sorted out by their applications in computer vision, NLP, cyber security, and the real world.

$\sbullet[.75]$  Adversarial defense methods are organized into three categories: Data modification, model modification, and the use of auxiliary tools.\\\hline

~\cite{DBLP:journals/cybersec/SunTZ18} & 2018 &
$\sbullet[.75]$  Provide some fundamental information about adversarial examples. 

$\sbullet[.75]$  The theoretical model for adversarial example attacks is contrasted with the real-world model.

$\sbullet[.75]$  Existing practical examples of adversarial attacks are presented.\\\hline

~\cite{DBLP:journals/access/AkhtarM18} & 2018 & 
$\sbullet[.75]$  Works on designing adversarial attacks are reviewed, and the occurrence of the attacks is analyzed.

$\sbullet[.75]$  Defenses against these attacks are proposed. 

$\sbullet[.75]$  Separately review the contribution of evaluating adversarial attacks in real-world scenarios. 

$\sbullet[.75]$  Offer a more comprehensive perspective on this field of research.\\\hline

\end{tabular}
	\label{abb}
\end{table*} 

\section{State-of-the-Art of Adversarial Attacks}
\label{attack}
An adversarial attack is a deliberate intent to mislead an ML or DNN model by introducing subtle, imperceptible interference to an input sample. This might result in the model drawing an incorrect conclusion confidently. Szegedy \textit{et al.}~\cite{DBLP:journals/corr/SzegedyZSBEGF13} were among the first to discover that DNNs are susceptible to slight adversarial perturbations. Ever since considerable efforts have been committed to producing more potent adversarial attacks for evaluating the robustness of DNNs.

From the perspective of the attack environment, adversarial attacks can be categorized into black-box attacks~\cite{DBLP:conf/ccs/PapernotMGJCS17}, white-box attacks~\cite{DBLP:conf/cvpr/CarliniF20}, or gray-box attacks~\cite{DBLP:conf/naacl/XuZJL21}. 
\begin{itemize}
    \item 
    A black-box attack signifies that the attacker has no knowledge of the underlying structure, learnable parameters, or defense strategies of the model under attack. The attacker interacts only with the model via its inputs and outputs~\cite{DBLP:conf/ccs/PapernotMGJCS17}.
    \item
    A white-box attack occurs when the attacker has all prior knowledge of the model under attack, e.g., the loss function and the optimized parameters of the model, and exploits the knowledge to facilitate the attack~\cite{DBLP:conf/cvpr/CarliniF20}.
    \item
     A gray-box attack accounts for the case, where the attacker only possesses partial knowledge of the model under attack in prior~\cite{DBLP:conf/naacl/XuZJL21}.
\end{itemize}

\subsection{Gradient-Based Attacks}
Gradient-based attacks are a common type of attack used against neural network models. These attack methods work by manipulating the input data {\color{black}according to the gradient of the loss function regarding the input} to cause the model's loss function to increase, effectively causing the model to make errors in its predictions. Classic gradient-based attack methods include FGSM~\cite{DBLP:journals/corr/GoodfellowSS14}, Basic Iterative Method (BIM)~\cite{2015Basic}, PGD~\cite{DBLP:conf/iclr/MadryMSTV18}, Jacobian-based Saliency Map Attack (JSMA)~\cite{DBLP:conf/eurosp/PapernotMJFCS16}, DeepFool~\cite{DBLP:conf/cvpr/Moosavi-Dezfooli16} and others. Numerous recent endeavors have been made to design and create new gradient-based attacks. Table~\ref{gradient} categorizes the latest gradient-based attacks from the perspectives of input, attack type, and invisibility metric (or ``inv-metric'' for brevity).

Yu \textit{et al.}~\cite{DBLP:conf/cvpr/YuG021} discover that certain ``robust'' models have hidden features that are unexpectedly susceptible to adversarial attacks. They propose latent feature attack (LAFEAT), a unified $\ell_{\infty}$-norm white-box attack algorithm that uses latent features throughout gradient descent steps for computationally efficient attacks, which can be cast as
\begin{equation}
\label{lfpgd} 
\begin{aligned}
    \max_{h, \lambda, \alpha, \mathcal{L}^{sur}} \mathcal{L}^{sce}(f _{\theta} (\text{PGD}_{\epsilon, x, y}(\mathcal{L}_\lambda ^{lf},\alpha, \mathit{I})),y),\\
  where \quad  \mathcal{L}_\lambda ^{lf}(\mathbf{z})=\mathcal{L}^{sur}\left(\sum_{l=1}^N \lambda^{(l)}h^{(l)}(\mathbf{z}^{(l)}), y\right).
\end{aligned}
\end{equation}
Here, $\mathcal{L}^{sce}$ represents the softmax cross-entropy (CE) loss between the one-hot truth value $y$ and the output. The constant $\mathit{I}$ defines the maximum number of iterations of gradient update iteration. For each layer $l\in \{1,\cdots,N\}$, the value $\lambda^{(l)} \in [0, 1]$ is assigned to the gradient of the layer, and the sum of all values is equal to 1. $\mathbf{z}^{(l)} = f^{(l)} \circ \dots \circ f^{(1)}(\mathbf{z})$ indicates the feature obtained from the $l_{th}$ layer. The mapping $h^{(l)}$ maps the features from $f^{(l)}$ to the logits for the $l$-th layer. For ease of exposition, Yu \textit{et al.}~\cite{DBLP:conf/cvpr/YuG021} define 
\begin{equation}
\label{pgd2} 
  \text{PGD}_{\epsilon,\mathbf{x},\mathbf{y}}(\mathcal{L},\alpha ,i )=\hat{\mathbf{x}}_{i},
\end{equation}
where $\hat{\mathbf{x}}_{i}$ is obtained by running the PGD algorithm~\cite{DBLP:conf/iclr/MadryMSTV18}. PGD identifies an adversarial instance by iteratively updating: 
\begin{equation}
\label{pgd} 
  \hat{\mathbf{x}}_{i+1}
= \mathcal{P}_{\epsilon , \mathbf{x} }(\hat{\mathbf{x}}_{i} + \alpha_i\, sign(\bigtriangledown_{\hat{\mathbf{x}}_{i}}{\mathcal{L}}^{sce}(f_\theta({\hat{\mathbf{x}}_{i}}),\mathbf{y}) )).
\end{equation}
Here, $\mathcal{I} \in \mathbb{R}^{C\times H\times W}$ limits the image data to a valid range, the function $\mathcal{P}_{\epsilon , \mathbf{x} } : \mathbb{R}^{C\times H\times W}\longrightarrow \mathcal{I} $
clips its input into the $\epsilon$-ball neighbor that is denoted as $\mathcal{I}$. The term $\bigtriangledown_{\hat{\mathbf{x}}_{i}}{\mathcal{L}}^{sce}(f_\theta({\hat{\mathbf{x}}_{i}}),\mathbf{y}) $
calculates the loss' gradient regarding the input $\hat{\mathbf{x}}_{i}$. 
$\alpha_i$ indicates the step size. At last, $sign(\cdot)$ is a sign function and returns -1, 0 or 1 for each element of the gradient. 

The objective of LAFEAT is to determine the optimal combination of logit mappings $h = (h^{(1)},\cdots ,h^{(N)})$, their respective weights $\lambda = (\lambda^{(1)},\cdots ,\lambda^{(N)})$, the size of the steps in the schedule $\alpha$, and the choice of the surrogate loss $\mathcal{L}^{sur}$ to use. Nonetheless, the efficient utilization of latent features as novel attack vectors has not yet been completely comprehended.

Che \textit{et al.}~\cite{DBLP:journals/tip/CheBZLLTGC21} propose an Iterative Partially White-box subspace attack (IPW). This technique establishes the cost function in the key hidden space, in which the receptive field is at its peak. The cost function penalizes the part of the feature activations that corresponds to both salient and guiding regions, rather than penalizing each pixel across the comprehensive dense output space. Moreover, they present an Iterative Black-box attack (IBA). This approach employs non-redundant variations from original models as initial hints to gauge the gradient of a target black-box model. The estimation is done through a zeroth-order iterative optimization process that computes the directional derivatives along the initial directions that are not redundant.

\begin{figure*}
    \centering
    \includegraphics[height=4in]{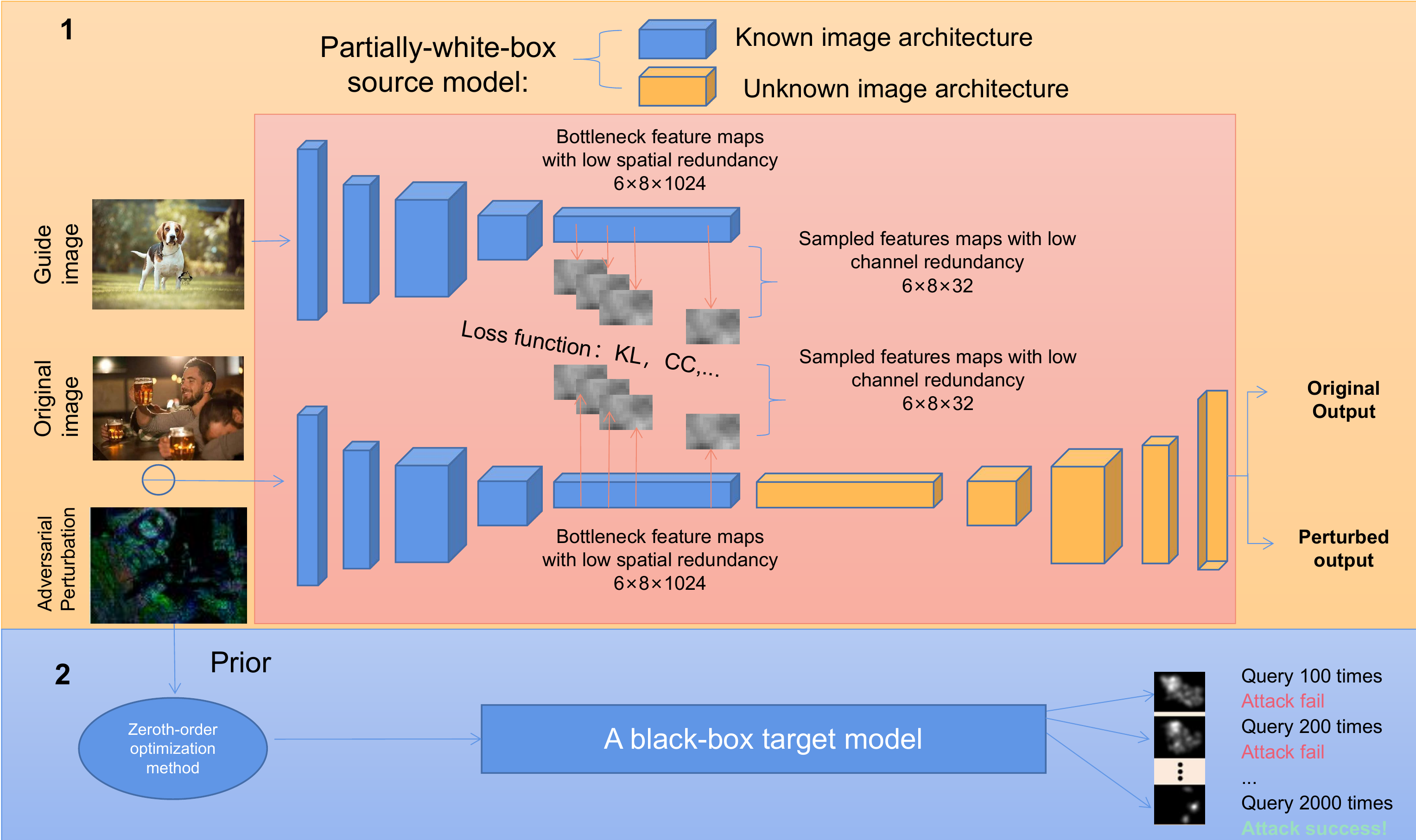}
    \caption{Sketch of the IPW\&IBA: The first step illustrates the concept of a subspace assault by creating a non-redundant prior perturbation from a partial white-box source model. To deceive a target model that is unknown to them, they merge an \textit{a-priori} optimizer with a zero-order optimizer in Step 2
    .~\cite{DBLP:journals/tip/CheBZLLTGC21}} 
    \label{ipw}
\end{figure*}

Fig.~\ref{ipw} offers the overview of the IPW\&IPA framework developed in~\cite{DBLP:journals/tip/CheBZLLTGC21}. The first step illustrates the concept of a subspace assault by producing a non-repetitive initial perturbation from a partial white-box source model. To deceive a target model that is unknown to them, they merge an \textit{a-priori} optimizer with a zero-order optimizer in Step 2. The method balances the attack capability and perturbation redundancy, overcoming the issues of costly attack cost and imperceptibility. Although the proposed non-repetitive initial cues enhance the black-box attacks, it remains challenging to fulfill the demands of certain time-sensitive applications, e.g., adversarial training necessitating a large quantity of adversarial instances.

Dynamically Sampled Nonlocal Gradient Descent (DSNGD)~\cite{DBLP:conf/ijcnn/SchwinnNRZETB21} computes the gradient direction for an adversarial attack by calculating the weighted mean of previous gradients from an optimization record.
The gradient computation in DSNGD can be written as
\begin{equation}
\label{DSNGD} 
\begin{aligned}
    \bigtriangledown_x \mathcal{L} (f_\theta (x_t),y) & := \sum_{i=1}^{t} w_i \cdot \bigtriangledown_x \mathcal{L} (f_\theta (\hat{x}_i),y),\\
\hat{x}_i & = \text{Clip}_{[0,1]}(x_i + \xi_i^\sigma).
\end{aligned}
\end{equation}
Here, $\mathcal{L}$ indicates the loss function, such as CE, of a neural network $f_\theta$. $\text{Clip}_{[0,1]}(\cdot)$ clamps the input to the range between 0 and 1; 
$\hat{x}_i$ denotes a noisy sample in the optimization process; $w_i$ refers to the gradient weight associated to $\hat{x}_i$; the random variables $\xi_i^\sigma$ are taken from the i.i.d. distribution $P^{\sigma}$ parametrized by the standard deviation $\sigma \in \mathbb{R}^+$. The variable $t$ stands for the iteration number during the current attack. This improves the efficiency of the algorithm by reducing the computational burden caused by sampling operations, eliminating the need for manually tuning additional hyperparameters, and providing a more precise estimation of the overall upward direction. However, its performance on larger datasets, e.g., ImageNet, is yet to be determined.

Chen \textit{et al.}~\cite{DBLP:journals/pami/ChenHSYH22} propose an Attack on Attention (AoA) method that depends on the semantic characteristics common among multiple DNNs to enhance the transferability of adversarial attacks. As opposed to prior techniques that concentrate on attacking the output, such as the One-Pixel attack developed in~\cite{DBLP:journals/tec/SuVS19}, AoA aims to modify the attention heat map and achieves exceptional results in black-box attacks. This method produces adversarial instances that can deceive numerous DNNs using zero queries and leads to a substantial improvement in transferability if the standard CE loss is substituted with an attention loss. The AoA attack can be seamlessly integrated with other transferability-enhancement methods to attain cutting-edge performance.

Graph structures are common in the physical world, and DNNs are commonly employed to tackle graph network problems, including node classification~\cite{DBLP:conf/sigir/Wen0L21} and link prediction~\cite{DBLP:conf/jist/Ugai21}. Iterative Gradient Attack (IGA)~\cite{DBLP:journals/tcss/ChenLSL20} is a new approach for link prediction that leverages gradient information from a trained Graph Autoencoder (GAE) model. IGA offers effective results 
under both complete and incomplete graph information,
and it can be integrated with various tasks. IGA also has good transferability across various realistic diagrams. Unfortunately, its computational complexity can grow significantly when the size of the graph increases.

\begin{figure}
    \centering
    \includegraphics[height=1.9in]{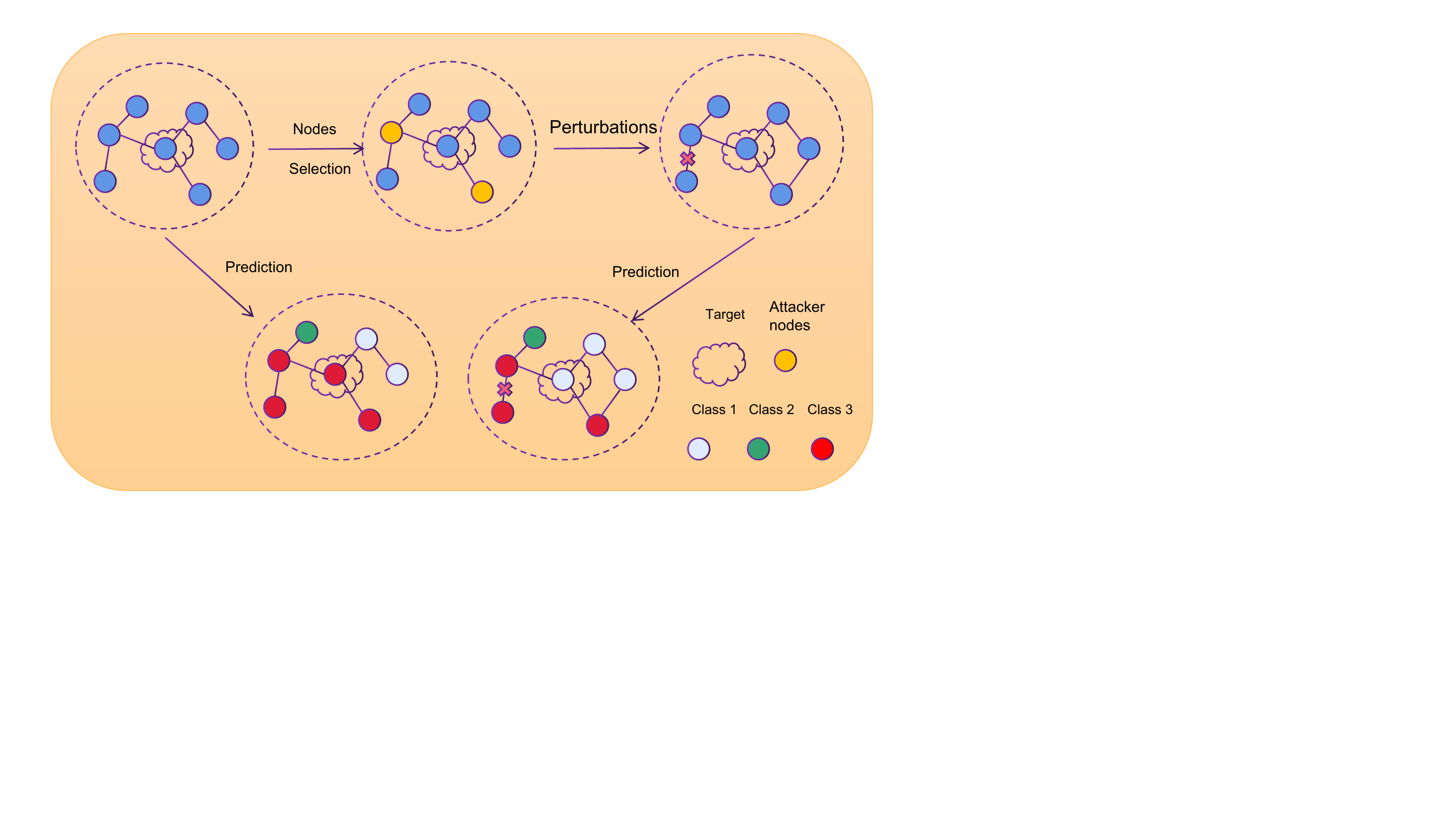}
    \caption{
Iterative gradient-based attack through subgraph
expansion. Here, $k=2$.
The dotted circle represents the neighborhood range ($k$) around the target node, i.e., a subgraph containing the target node. Flipping edges within a subgraph lead to misclassification of the target node.}
    \label{graph attack}
\end{figure}

Considering node classification tasks, Li et al.~\cite{DBLP:journals/tkde/LiXCXHZ23} propose a Simplified Gradient-based Attack (SGA), 
which addresses the difficulty in attacking large-scale graphs by leveraging a subgraph that comprises $k$-hop neighbors of attacked. An input graph is perturbed by sequentially flipping edges whose magnitude of gradients is the biggest in this subgraph. 
The SGA method overcomes the issue of gradient fading in gradient-based attack techniques by using a smaller subgraph centered around the node under attack and by incorporating a scaling factor. As depicted in Fig.~\ref{graph attack}, SGA has a significant advantage over other cutting-edge attack techniques, {\color{black}e.g., the GradArgmax method developed in~\cite{DBLP:conf/icml/DaiLTHWZS18} and the Nettack method developed in~\cite{7876373},} in terms of time and memory efficiency, meanwhile still achieving considerable attack results. 

{\color{black} On the other hand, a novel attack scenario is known as a node injection attack, in which attackers can inject a set of malicious nodes into a graph to circumvent the original graph's topology and misclassify victim  nodes~\cite{DBLP:journals/datamine/WangLSLYZ20}.
SGA is generally inapplicable to the node injection attack because the injected malicious node is a singleton node and is not initially linked to any nodes, where  a $k$-hop subgraph cannot be extracted,}
and its high computational cost is also a drawback. Currently, SGA is limited to node classification tasks and targeted attack scenarios. Ongoing research is expected to expand the method and make it more adaptable to various graph analysis tasks.

{\color{black} Gradient-based attackers collect gradients of node features and graph structures and produce perturbations based on them using pre-trained Graphic Neural Network (GNN) classifiers, referred to as proxy models. However, the majority of existing work~\cite{DBLP:journals/datamine/WangLSLYZ20, DBLP:journals/tcss/ChenLSL20}. concentrate on using gradients to produce perturbations rather than looking at how to get more dependable gradients from different models. The gradient-based perturbations manufactured by the attacker are affected by the proxy model's embedding layer mapping. The perturbations generated are model-specific, and lose their generalization to another model. To solve this problem, Liu \textit{et al.}~\cite{9667076} propose Surrogate Representation Learning with Isometric Mapping (SRLIM) to enable the model to learn topologies. Keeping the similarity of nodes from the input layer to the embedding layer, SRLIM passes topological knowledge to the embedding layer, thus improving the effectiveness of the adversarial attacks produced by gradient-based attackers in non-target poison gray-box attacks. However, as the complexity of the graph structure rises, the computational complexity will also increase exponentially.}

In contrast to the prevalent adversarial attack methods that generate only one adversarial instance for an input, Wang \textit{et al.}~\cite{DBLP:journals/pami/WangLLL22} introduce an attack method that produces a range of adversarial examples for a given input. This is achieved through the use of Hamiltonian Monte Carlo with Accumulated Momentum (HMCAM). They also present a novel generative technique, namely Contrastive Adversarial Training (CAT), which employs Hamiltonian Monte Carlo (HMC) to simulate the creation of adversarial examples and achieves an equilibrium distribution of adversarial instances within just a few rounds by performing some moderate changes of the conventional Contrastive Divergence~\cite{DBLP:journals/neco/Hinton02}. As a result, CAT strikes a balance between attack accuracy and efficiency in adversarial training. The experimental outcomes demonstrate that CAT attains a higher attack success rate (ASR) than black-box models, and is on par with other white-box models.

\begin{figure}
    \centering
    \includegraphics[width=3.5in]{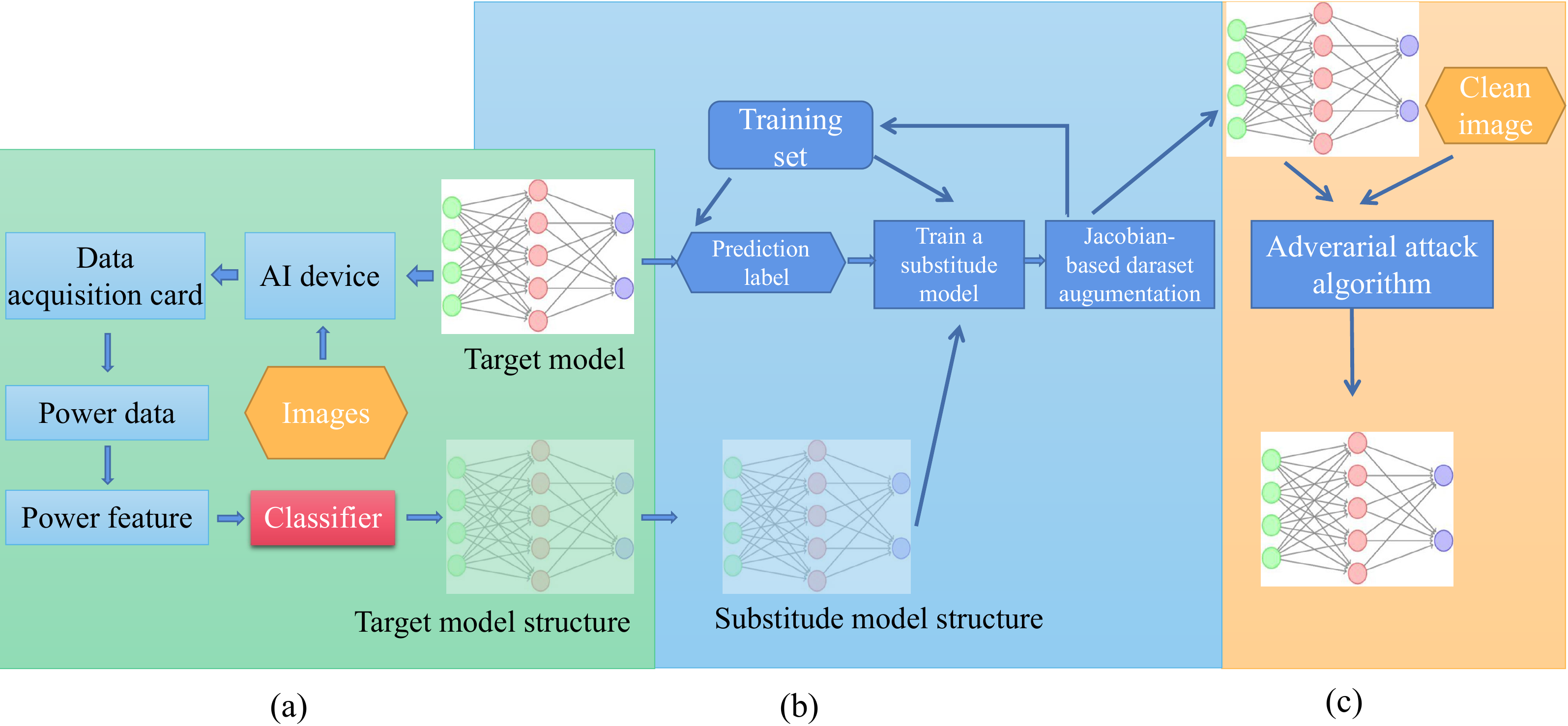}
    \caption{Working flow chart of the SCA-based attack. (a) Obtain the network structure of the target model. (b) Training an alternative model. (c) Creating an example of hostility.} 
    \label{sca}
\end{figure}

{\color{black} Xiang \textit{et al.}~\cite{9149635} come up with a straightforward and efficient gray-box attack strategy based on the side-channel attack (SCA) policy. The SCA-based attack is illustrated in Fig.~\ref{sca}. First, the target model's fundamental network structure is derived using an SCA-based attack. The alternative model is then trained using the derived network structure. Adversarial samples are produced using the trained alternative parameters in order to mislead the target model. The trained gray-box replacement model's decision boundary is nearer to the target model because gray-box attacks use abundant internal information, as opposed to black-box attacks. It is thus more realistic than a white-box attack and more efficient than a black-box attack. However, there might be more possible architectures in real-world applications. The algorithm must run every step of adversarial and side-channel attacks against every candidate architecture, which can take significant time and resources.}

{\color{black}In the face of errors caused by the discreteness of graph structures and subsequent rough gradients, for the vulnerability of GNNs to the semantic space and parameter random initialization resulting in an unstable representation of GNNs, Liu \textit{et al.}~\cite{10.1145/3511808.3557238}  proposed two modules to solve the problem, namely the semantic invariance module and the momentum gradient integration module, and integrated the above modules to propose an attack model named Attacking by Shrinking Errors (AtkSE). This method solves the gradient fluctuation in semantic graph enhancement and the instability of the proxy model to some extent, increases the attack intensity of the attacker, and ensures the transferability of the gray-box attack. But at the same time, the trade-off between computational efficiency and error reduction is also worth further study.}

All of these methods aim to find ways to generate adversarial examples that can fool the DNNs by exploiting their gradients. However, they have different approaches and techniques to {\color{black}leverage the gradients}. They also have different strengths and limitations. For instance, some methods take less computational time and memory, such as SGA~\cite{DBLP:journals/tkde/LiXCXHZ23}, CAT~\cite{DBLP:journals/pami/WangLLL22}, and LAFEAT~\cite{DBLP:conf/cvpr/YuG021}, while others are better in terms of transferability and attack performance, such as IGA~\cite{DBLP:journals/tcss/ChenLSL20}, AoA~\cite{DBLP:journals/pami/ChenHSYH22}, SRLIM~\cite{9667076}, and AtkSE~\cite{10.1145/3511808.3557238}.

{\small
\setlength{\tabcolsep}{2pt}
\setlength\LTleft{-0.1in}
\setlength\LTright{-1in plus 1 fill}

\begin{table*}
\caption{Gradient-based Adversarial Attack Methods}
	\label{gradient}
	\setlength{\tabcolsep}{3pt} 
	\renewcommand\arraystretch{1.5} 
	\begin{tabular}{|p{1.2cm}|p{3.4cm}|p{1cm}|p{0.9cm}|p{1.2cm}|p{4.5cm}|p{4cm}|} 
    		\hline
    \centering{\textbf{Attack}} & \centering{\textbf{Short Description}} & \centering{\textbf{Input}} & \centering{\textbf{Attack type}} & \centering{\textbf{Invisibility Metric}} & \centering{\textbf{Advantage}} &\centering{\textbf{Disadvantage}} \cr
    \hline

    IGA \cite{DBLP:journals/tcss/ChenLSL20} & A new attack method for link prediction that utilizes gradient information from a trained GAE model. & Discrete & White-box  & cross-entropy 
    & IGA achieves effective attack results regardless of whether the global graph information is complete or not. Strong transferability on different realistic diagrams.
    & The algorithmic complexity dramatically increases when the size of graphs grows larger.\\
     \hline
    
    SGA \cite{DBLP:journals/tkde/LiXCXHZ23} & A framework that reduces the scale of the graph to a smaller subgraph centered around the target node, resolving the challenge of performing attacks on large graphs.& Discrete & Black-box & DAC 
    & Significantly enhance time and memory efficiency and attach considerable attack strength; Strong transferability among different commonly used graph neural networks. 
    & Not available for node injection attack due to the costly computation of DAC; Often used for node classification or targeted attacks.\\
     \hline
    
    CAT \cite{DBLP:journals/pami/WangLLL22} & A generative scheme named Contrastive Adversarial Training is designed, inspired by HMCAM, which aims to produce a series of adversarial examples in a single run. & Conti-nuous & White-box, Black-box & $\ell_{\infty}$ 
    & Its ASR is superior to other black-box models, equivalent to other white-box models, strikes a compromise between efficiency and accuracy in AT, and demonstrates enhanced transferability. &In the ImageNet, the ResNet-50 model trained with the PGD algorithm takes one-third of its time, but the best robust accuracy is inferior.\\
     \hline

    IPW\&IBA \cite{DBLP:journals/tip/CheBZLLTGC21} & Create a cost function that penalizes a subset of feature activations. To determine the real gradient of the black-box target model, compute the directional derivatives along the non-redundant previous directions using an iterative zeroth-order optimization procedure.& Conti-nuous & White-box, Black-box & KL, PLC, NSS, $\ell_1$ 
    & The best trade-off between attack capabilities and redundancy compared to other white-box attacks, making a good trade-off between black-box attack capabilities and query costs. 
    & While speeding up black-box attacks, it currently faces challenges in meeting the demands of time-sensitive applications, such as AT, which requires a substantial amount of adversarial examples.\\
     \hline
     
    DSNGD \cite{DBLP:conf/ijcnn/SchwinnNRZETB21} & Compute the weighted mean of previous gradients from the optimization history to determine the gradient direction of an adversarial attack. & Conti-nuous & White-box, Black-box & $\ell_{\infty}$ 
    & The sampling operation's computing overhead is reduced, resulting in greater efficiency. Less vulnerable to noise and local optimization, and more accurately approximate the global upward direction.
    & The performance of the method on larger datasets, e.g., ImageNet, has not been determined.\\
     \hline
     
    AoA \cite{DBLP:journals/pami/ChenHSYH22}& Based on the semantic properties shared by DNN. Unlike other methods that focus on attack output, AoA changes the attention heat map and the loss function. & Conti-nuous & White-box, Black-box & RMSE  & 
    Beat many DNNs with zero queries. Increased transferability when using traditional cross-entropy loss instead of attention loss. Easy to combine with other transferability enhancement technologies to achieve SOTA performance.
    & Even though the generated examples are distinct from others, they can still be captured by AT.\\
     \hline
     
    LAFEAT \cite{DBLP:conf/cvpr/YuG021}& LAFEAT algorithm takes advantage of latent features in its gradient descent steps. & Conti-nuous & White-box & $\ell_{\infty}$ & Seek to harness latent features in a generalized framework. Computationally efficient. & It remains unclear how latent features can be leveraged as viable attack vectors. \\
    \hline
    
   SCA-based \cite{9149635} & A gray box attack method using SCA to predict model structure using pre-trained classifiers. & Conti-nuous & Gray-box & $\ell_p$ & The decision boundary of a trained gray-box alternative model is nearer to the target model. More effective than a black-box attack, and more practical compared to a white-box attack. & For complex architectures, the algorithm can be time-consuming and resource-intensive.\\
     \hline
    SRLIM \cite{9667076} & An approach that uses SRLIM to preserve the topology in proxy embedding and thereby improve the performance of a gradient-based attacker in a non-target poison gray-box scenario for adversarial attacks. &  Discrete & Gray-box & $\ell_0$ & SRLIM enables the proxy model to learn topologies through isometric mapping, thereby improving the reliability of gradients utilized in the attack models and the transferability. & As the complexity of the graph model structure rises, the computational complexity grows exponentially. \\
     \hline
     
     AtkSE \cite{10.1145/3511808.3557238} & An attack model that integrates semantic invariance modules and momentum gradient ensemble modules to reduce errors within the structural gradients & Discrete & Gray-box & $\ell_0$ & The gradient fluctuation in semantic graph enhancement and the instability of proxy models are addressed. It improves the attack intensity of the attacker and ensures the transferability of the gray-box attack. & The trade-off between computational efficiency and error reduction is also worth further study. \\
     \hline
     \end{tabular}
\end{table*}
}

\subsection{Constrained Optimization-Based Attacks}
Attacks based on constrained optimization involve creating adversarial examples by tackling a constrained optimization problem. This method seeks the minimum perturbation, subject to $\ell_0$, $\ell_2$, or $\ell_{\infty}$-norm constraints that cause the neural network model to make an incorrect classification. As such, adversaries can generate an adversarial example $\mathbf{x}^{adv}$ for an untargeted attack (i.e., misclassifying the adversarial example to any different class from the correct one) by following:
\begin{equation}\label{atta} 
\begin{aligned}
    \max_{\mathbf{x}^{adv}} &~ \mathcal{L}(f(\mathbf{x}^{adv}),y), \\
  \text{s.t.}&  \left \| \mathbf{x}^{adv}-\mathbf{x} \right \|_{\infty } \le \epsilon,
\end{aligned}
\end{equation}
where the objective is to find an adversarial counterpart $\mathbf{x}^{adv}$ and the constraint $\epsilon$ specifies the invisibility requirement of adversarial perturbation. $\mathcal{L}(\cdot,\cdot)$ is the loss function of the target model $f$, and $\mathbf{x}$ is a clean example. In the case of a targeted attack (i.e., misclassifying the adversarial example to an incorrect class designated by the attacker), the objective function of the optimization problem is:
\begin{equation}\label{targeted-atta} 
\begin{aligned}
    \min_{\mathbf{x}^{adv}} \mathcal{L}(f(\mathbf{x}^{adv}),y_t), \\
\end{aligned}
\end{equation}
where $y_t$ is the label of the class designated by the attacker.

Szegedy \textit{et al.}~\cite{DBLP:journals/corr/SzegedyZSBEGF13} first propose an L-BFGS algorithm to transform a difficult optimization problem of finding a perceptually-minimal input perturbation into a box-constrained formulation. Many other popular methods, such as C\&W~\cite{DBLP:conf/sp/Carlini017}, AdvGAN~\cite{DBLP:conf/ijcai/XiaoLZHLS18}, and UAP~\cite{DBLP:conf/cvpr/Moosavi-Dezfooli17}, are also achieved by carrying out constrained optimizations. Table~\ref{table2} categorizes the latest constrained optimization-based attacks from the perspectives of input, attack type, and inv-metric.

Many recent works utilize constrained optimization techniques to generate adversaries. For instance, Chen \textit{et al.}~\cite{DBLP:conf/ijcnn/ChenCW21} present adGAN, a method for producing adversarial attacks that greatly degrade the performance of reinforcement learning systems. The fundamental concept of adGAN is to manipulate the current state in reinforcement learning, misleading the agent into thinking it is in a correct state, thereby causing it to make subpar decisions in each step and leading to a decrease in overall rewards. AdGAN has demonstrated its ability to transfer and adapt well to different situations. 
The loss function of the adversarial attack is written as:
\begin{equation}
\label{adgan} 
\begin{aligned}
    \tilde{\mathbf{x}}  = \max_{\mathbf{x}} & ~\mathcal{L}_{T_i}(f_\varphi ),\\
   \text{s.t. }& \parallel \tilde{\mathbf{x}} - \mathbf{x}^{adv} \parallel_2 \le \epsilon. 
\end{aligned}
\end{equation}
Here, $\mathcal{L}_{T_i}(f_\varphi)$ represents the loss function of the reinforcement learning task $T_i$, $\tilde{\mathbf{x}}$ symbolizes the optimal state in the Markov decision process tackled by the reinforcement learning, $\mathbf{x}^{adv}$ is the altered state, and $\epsilon$ indicates the perturbation magnitude. The aim of $T_i$ is to learn a function $f$, which is parameterized by $\varphi$ to maximize the expected overall discounted reward.

Adversarial Transformation-enhanced Transfer Attack (ATTA)~\cite{DBLP:conf/cvpr/WuSLK21} is a technique that uses adversarial learning to train a CNN as an adversarial transformation network. This network is capable of capturing the most destructive deformations and transforming them into adversarial noises. The adversarial samples designed to withstand distortions induced by the adversarial transformation network are more robust and transferable. ATTA's performance may be enhanced by combining it with other {\color{black}transfer-based attacks, e.g., momentum iterative fast gradient sign Method (MI-FGSM) developed in~\cite{DBLP:conf/cvpr/DongLPS0HL18} and Query-Efficient Black-Box Adversarial attack developed in~\cite{9609659}.} However, overly simplistic or complex structures can negatively impact the attack's performance, as the former lacks sufficient representation power, and the latter causes the adversarial transformation network to be excessively adaptive to the backbone attack algorithm.

\begin{figure*}
    \centering
    \includegraphics[height=2.5in]{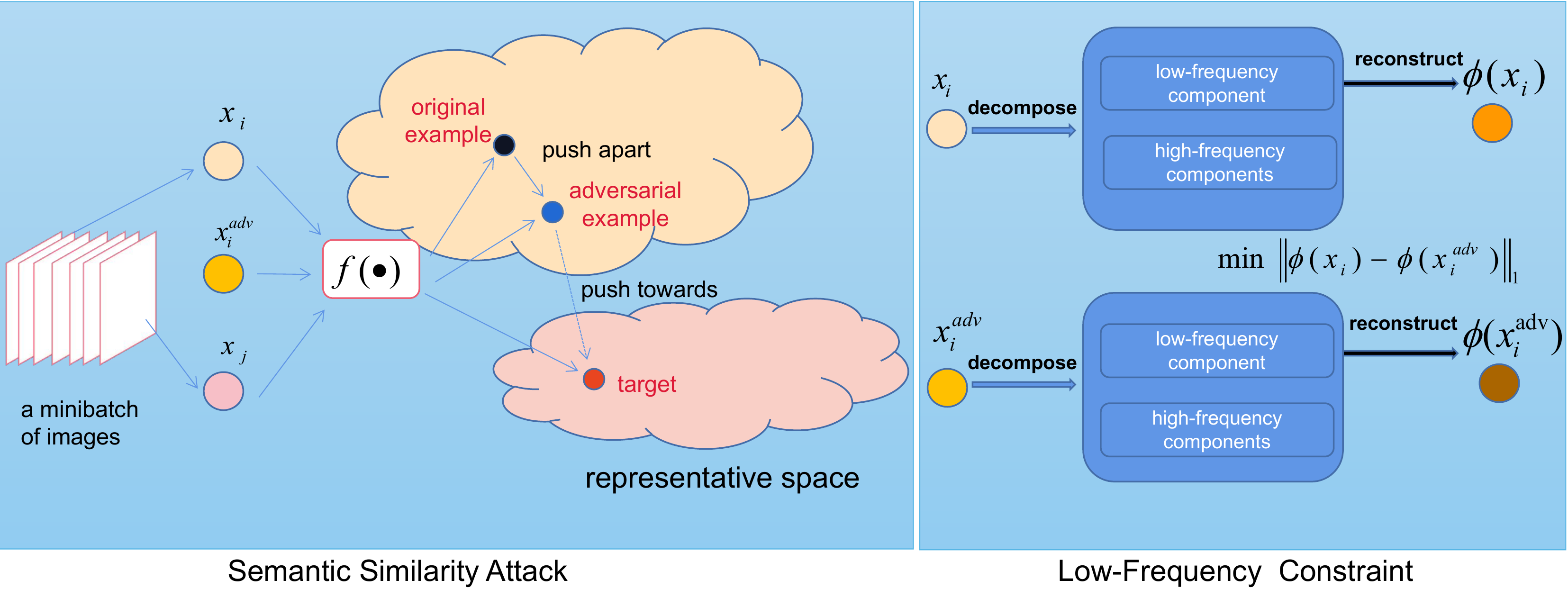}
    \caption{Overview of the SSAH method. The left subfigure illustrates semantic similarity attacks, while the right subfigure demonstrates the low-frequency constraint. $f(\cdot)$ represents the function mapping from the image to the representation space, and $\phi(\cdot)$ denotes the shallow neural network that divides the image into different frequency components and then reconstructs it by using low-frequency components.} 
    \label{ssah}
\end{figure*}

Luo \textit{et al.}~\cite{DBLP:conf/cvpr/LuoL0WXS22} propose an adversarial attack method called SSAH (semantic similarity attack on high-frequency components) based on frequency space constraints, which restricts the adversarial noise to the high-frequency components of the picture, so that the human eye perceives the noise with relatively low similarity. The framework of SSAH is displayed in Fig. ~\ref{ssah}. The attack strategy involves increasing the semantic resemblance between the adversarial sample and a randomly selected sample, while simultaneously decreasing the feature similarity between the adversarial sample and the original image. SSAH steps out of the original framework based on $\ell_p-$norm constraints, and provides a new idea of adversarial noise generation and constraints in frequency space. However, although SSAH has a great decrease in the recognition, the attack success rate has not significantly improved. Therefore, for invisible attacks, the trade-off of their invisibility and attack success rate is still a problem that needs to be solved.

Bonnet \textit{et al.}~\cite{DBLP:journals/tifs/BonnetFB22} propose a method specifically for quantizing adversarial perturbations. The quantization is implemented as a customizable post-processing approach that may be employed over any white-box attacks aimed at any model, with less additional distortion and fewer cycles required for the attack operation. This strategy, however, needs to access gradients available in the white-box design and does not ensure transferability to other DNNs.

Because the white-box attack needs access to predictions and labels, it is impractical for a realistic learning system. Hence, researchers have focused on black-box attacks. Chang \textit{et al.}~\cite{9720115} present a generalized adversarial attack framework (GF-Attack). This black-box attack system can execute adversarial attacks on different kinds of graph embedding models (GEMs) without access to labels or model predictions. The objective is to improve the robustness of GEMs. Although GF-Attack has a lower computational efficiency than the Random method~\cite{2021Hypergraph} and the Degree method~\cite{DBLP:journals/tsipn/ChenTTLKH19}, it can execute an adversarial attack on a range of GEM types with high transferability, flexibility, and extensibility without altering the target embedding model.

Non-Dominated Sorting Genetic Algorithm with Particle Swarm Optimization (NSGA-PSO)~\cite{DBLP:conf/ijcnn/FengFXWHX21} is an optimization-based method for generating digital watermarking adversarial perturbations. It has higher ASRs than existing black-box attack approaches, good transferability among multiple network models, and greater resistance to image modification countermeasures. However, testing findings demonstrate that its performance is worse on the CIFAR-10 dataset, compared to the ImageNet dataset.

\begin{figure}
    \centering
    \includegraphics[height=2in]{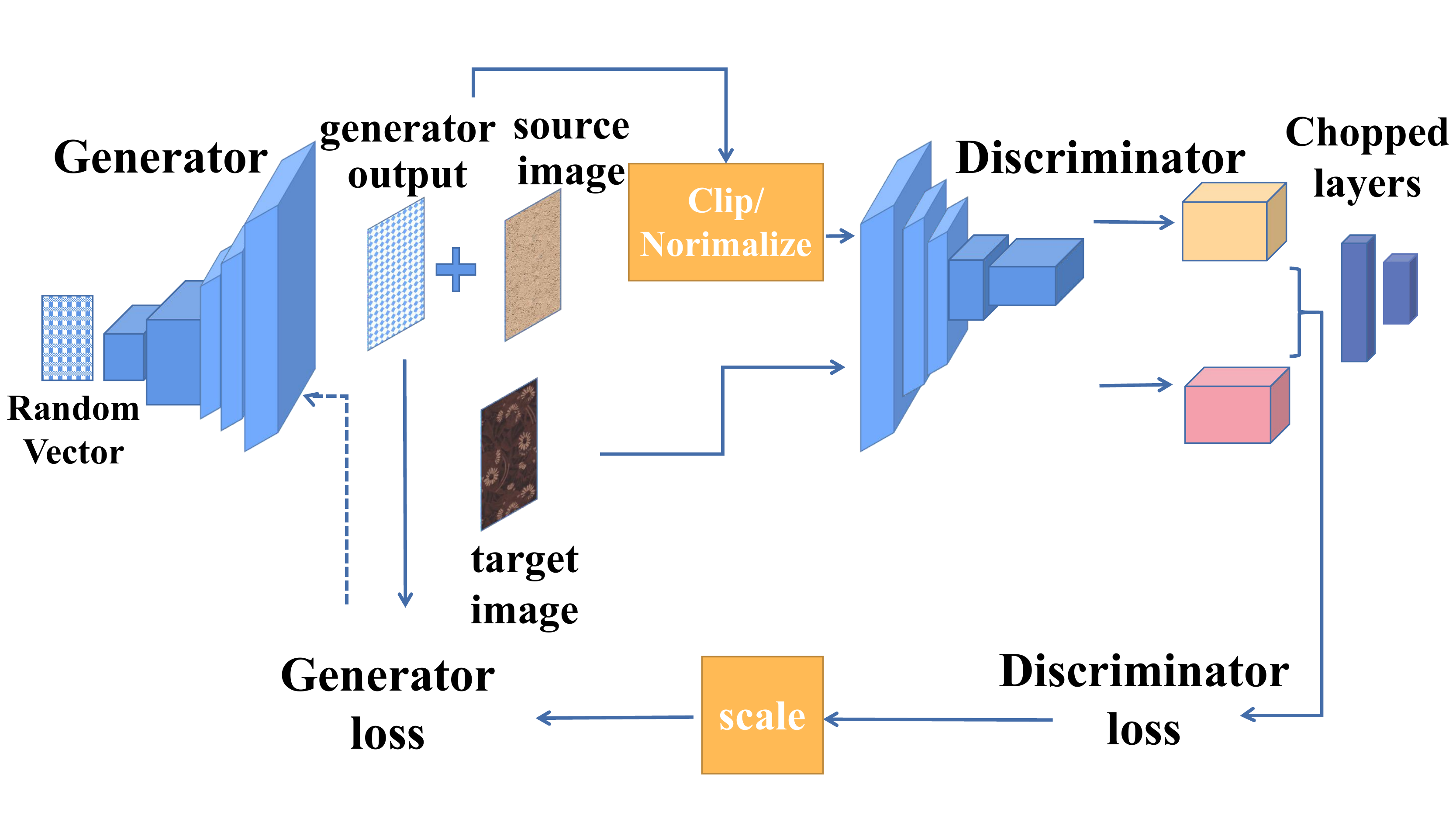}
    \caption{{\color{black} Overview of the Language Model Agnostic Attack. The output of the generator combined with the source image and the target image are both inputs to the discriminator, and then the discriminator backpropagation updates the generator. The resulting generator output overlaid onto the source image to generate the title.}} 
    \label{caption}
\end{figure}

{\color{black}Despite the fact that attacks on visual models (such as CNNs) have been well studied, the adversarial vulnerability of neural image captioning has not been thoroughly investigated, because of the unique difficulties of the ``multi-model" problem in subtitles. Aafaq \textit{et al.}~\cite{9970367} suggest that images be altered in line with internal representations of visual models applied in captioning frames to deceive encoder-decoder-based image captioning frameworks. A GAN-based method is suggested, which can alter the representation of the internal layers necessary for the image in order to produce adversarial images. The diagram of the language model agnostic adversarial attack on image caption is demonstrated in Fig.~\ref{caption}. The attack begins by sampling a random vector from a uniform distribution via a generator. The generator output is combined with the source image. The scaled outcome is fed into the discriminator to obtain the desired depth representation. Similarly, the target image is fed into the discriminator. The discriminator back propagates the gradient to update the input image. In scale and again after deducting the original image, the disturbance is separated, and the gradient of the generator has been updated. In order to produce significant image features for the target (incorrect) class and suppression features for the source (correct) class, generators are trained to compute perturbations. The output of the generator is then attached to the source image to fabricate a title close to the title of the target image. This enables an attacker to successfully control the image's title without needing to be familiar with the caption model. But the method also has the problem that the disturbance is imperceptible and difficult to control.}

All of these methods, e.g.,~\cite{DBLP:conf/ijcnn/ChenCW21}--~\cite{9970367}, generate adversarial examples to attack ML or DNN models by using constrained optimization, such as bi-level optimization, quantization, or particle swarm optimization (PSO). They have been shown to have good transferability spanning many network architectures, and solid resistance to example transformation defensive strategies. Most of these methods, such as  GF-Attack~\cite{9720115} and NSGA-PSO~\cite{DBLP:conf/ijcnn/FengFXWHX21}, concentrate on improving the robustness of graph embedding models (GEMs) and digital watermarking adversarial perturbations and have good extension and flexibility. Moreover, they do not depend on access to the predictions and labels, making them more suitable for real-world scenarios. 
However, these methods could perform poorly on different datasets.

{\small
\setlength{\tabcolsep}{2pt}
\setlength\LTleft{-0.1in}
\setlength\LTright{-0.1in}
\begin{table*}
\caption{Constrained optimization-based adversarial attacks}
	\label{table2}
	\setlength{\tabcolsep}{3pt} 
	\renewcommand\arraystretch{1.5} 
	\begin{tabular}{|p{1.22cm}|p{4.6cm}|p{1cm}|p{0.9cm}|p{1.2cm}|p{4.4cm}|p{3cm}|} 
    \hline
    \centering{\textbf{Attack}} & \centering{\textbf{Short Description}} & \centering{\textbf{Input}} & \centering{\textbf{Attack type}} & \centering{\textbf{Invisibility Metric}} & \centering{\textbf{Strength}} & \centering{\textbf{Weakness}}\cr
    \hline
    GF-Attack \cite{9720115} & A framework for adversarial attacks that may be launched against various GEM types.& Discrete &  Black-box & $\ell_2$
    & Good transferability on various kinds of GEMs, flexibility and extensibility, not change the target embedding mode. 
    &  The calculation efficiency is lower than the Random method and the Degree method.\\
     \hline
    adGAN \cite{DBLP:conf/ijcnn/ChenCW21}& A framework that undermines the current state in reinforcement learning by enticing the agent to make sub-optimal choices in each step, leading to a reduction in overall rewards. &  Discrete &  White-box &  $\ell_2$  & Model agnostic, good generalization capacity, could converge quickly in all environments, quite stable performance. & The calculation cost may be higher. \\\hline
    
    SSAH \cite{DBLP:conf/cvpr/LuoL0WXS22} & By attacking the semantic similarity of the images, a wide range of settings is applied. & Conti-nuous& White-box  & LF & More transferable across different architectures and datasets, significantly imperceptible. & No significant increase in aggressiveness.
  \\ 
     \hline
    
    NSGA-PSO \cite{DBLP:conf/ijcnn/FengFXWHX21} & A method for producing adversarial perturbations for digital watermarking by using an optimization algorithm. & Conti-nuous & Black-box & $\ell_2$ 
    & Satisfactory transferability across different networks, fewer queries but higher success rates.
    & Lower ASRs on CIFAR-10 than on ImageNet.\\
    \hline
     
    ATTA \cite{DBLP:conf/cvpr/WuSLK21} & A CNN is trained as the adversarial transformation network through adversarial learning, allowing it to capture the most damaging deformations in response to adversarial noise.& Conti-nuous & White-box, Black-box & $\ell_{\infty}$ 
    & The adversarial samples created are more robust and transferable; Combine well with other transfer-based attacks to boost effectiveness.
    & Too simplistic or complex structures reduce attack performance. \\
     \hline

    Adversarial Quantization \cite{DBLP:journals/tifs/BonnetFB22}& A method specifically designed to quantify against perturbation, with the aim of minimizing quantization errors after quantization while maintaining the image's adversarial nature. & Conti-nuous &  White-box & $\ell_2$ & Propose post-processes that can be utilized for any white box attack. Require fewer iterations than the conventional attack process and add little additional distortion. & Its transferability to other DNNs cannot be guaranteed; Only works on white-box attacks.\\
    \hline
      
    Language Model Agnostic Attack \cite{9970367} &  An algorithm for generating adversarial instances on a multi-model image captioning frame. The algorithm does not require language module information and controls the predicted title by attacking the visual encoder of the title frame.& Conti-nuous & Gray-box & Meteor Score & The perturbation is calculated through a single forward pass of the deployment model, unlike the typical iterative approach, which incurs higher time consumption. The output of the language model can be controlled with no need for any information about the model.& The improvement in attack performance comes at the cost of more perceptibility of disturbances in the image.\\
    \hline
     \end{tabular}
\end{table*}
}

\subsection{Gradient-Free (Heuristic) Attacks}
Heuristic attacks are a sort of attack that does not depend on gradients and can include techniques such as search-based, decision-based, and drop-based methods. These methods can have their own advantages and disadvantages, and a summary of the latest gradient-free attacks from several perspectives can be found in Table~\ref{table3}.

Feature-Wise Convex Polytope attack (FeaCP)~\cite{DBLP:journals/tifs/LiQHQLD21} places emphasis on limiting the placement of generated samples. It aims to find adversarial samples close to the decision boundary and correct existing areas of vulnerability in neural networks. Rather than solely focusing on the capacity for an attack, FeaCP places a greater emphasis on controlling the generation process for the purpose of defending the model. FeaCP considers the significance of adversarial instances in relation to the target model during the creation process, and provides a clear insight into the location of adversarial instances through the adversarial direction. FeaCP can be applied to other fields, such as sentiment analysis~\cite{DBLP:journals/access/AriasNGTV22}. 
FeaCP creates potential adversarial examples with confined variables, as given by
\begin{equation}
\label{feacp} 
\begin{aligned}
    \mathbf{x}^{adv} = \mathbf{\lambda}_s \odot \mathbf{x} + \sum_{i=1}^{M}\mathbf{\lambda}_g^i\odot \mathbf{x} _g^i 
\end{aligned}
\end{equation}
Here, $\mathbf{x}$ stands for the benign example; $\mathbf{x} _g^i$ denotes the $i$-th guidance example in a collection of randomly selected guidance samples of $M$ different classes; $\mathbf{\lambda}_s$ and $\mathbf{\lambda}_g^i$ represent the tensors of coefficients that has the same shape as that of $\mathbf{x}$; 
$\mathbf{\lambda} = \left \{ \mathbf{\lambda}_s, \mathbf{\lambda}_g^1,\cdots, \mathbf{\lambda}_g^M \right \} $ and satisfies the following condition:
\begin{equation}
\label{eqn:convex}
\begin{aligned}
   \mathbf{\lambda}_s^q +\sum_{i=1}^{M}\mathbf{\lambda}_g^{iq}=1,
\end{aligned}
\end{equation}
where $q$ indicates the $q$-th element of a tensor. Eqn.~\ref{eqn:convex} ensures each feature of the composite sample is a convex combination of the bootstrap sample and the relevant features in the source samples to provide sufficient flexibility for perturbation of each feature in finding the blind spots of the DNNs. 

Adversarial Laser Beam (AdvLB) is a novel attack method introduced by Duan \textit{et al.}~\cite{DBLP:conf/cvpr/DuanMQCYHY21}, which uses a greedy search and laser beams as a malicious perturbation. This method has high flexibility, allowing it to attack any object, even from long distances actively. AdvLB also has high temporal stability because of its physical attack mechanism. On the other hand, its deployment is simple, making it less secretive than other methods, such as AdvCam~\cite{DBLP:conf/cvpr/DuanM00QY20}.

Hash adversary generation (HAG)~\cite{DBLP:journals/tcyb/YangLDT20} is a technique for creating adversarial examples for a search in the Hamming space that solves a widely perceived ``gradient vanishing'' issue\footnote{``Gradient vanishing'' refers to the phenomenon that during the backpropagation of a deep neural network, the gradients of the network can become very small as they propagate through the layers of the network. When the gradients become too small and effectively become zero, this essentially prevents the lower layers of the network from learning any useful features~\cite{9526915}.} by introducing a smoother activation function. The objective of HAG is to generate subtly altered samples that bear no semantic connection to the original queries and whose closest neighbors come from a chosen hashing model. Even though the perceivability is still low, HAG can successfully attack target hash models. The learned perturbation is highly portable across settings and is more pronounced for the same architecture at varying hash bit lengths.

An Intersection over Union (IoU) attack~\cite{DBLP:conf/cvpr/JiaS0Y21} is a black-box, decision-based technique for visual object tracking. It creates disturbances using the calculated IoU scores from current and prior video frames. The attack decreases the accuracy of temporally consistent bounding boxes by lowering the IoU scores. Denote the original example (i.e., the original image in the video frame) as $\mathbf{x}$, the heavy noise example as $\mathbf{X}$, and the intermediate example on the $i$-th iteration as $\mathbf{x}^{(i)}$. The IoU attack labels $\eta$ as a nearby assumption based on $\mathbf{x}^{(i)}$ and advances $\mathbf{x}^{(i)} + \eta$ towards the highly noisy example $\mathbf{X}$ by following the update rule below:
\begin{equation}
\begin{aligned}
   \mathbf{x}^{(i+1)} = (\mathbf{x}^{(i)} + \eta) + \alpha \cdot \psi(\mathbf{X}, \mathbf{x}^{(i)} + \eta),
\end{aligned}
\end{equation} 
where $\alpha$ represents the stride towards $\mathbf{X}$ and $\alpha \cdot \psi(\mathbf{X}, \mathbf{x}^{(i)} + \eta)$ represents the disturbance in the direction of greater noise, i.e., in the direction of the normal to the noise level contour.

To circumvent the constraints of model-dependent approaches, such as the C\&W constraint undergone by a well-trained classifier in~\cite{DBLP:conf/sp/Carlini017}, Zhang \textit{et al.}~\cite{DBLP:journals/tip/ZhangTLWT20} present the Principal Component Adversarial Example (PCAE) method. PCAE produces adversarial samples without a specific target in mind. It is based on the idea of the adversarial zone where data points offer a possible danger to all classifiers. As an untargeted adversarial sample generation approach, PCAE utilizes a data manifold that does not depend on classification models. As a consequence, it is immune to overfitting and the restrictions of inadequate labeled data.

Different from all previous attacks, AdvDrop 
  is a novel adversarial attack proposed by Duan \textit{et al.}~\cite{DBLP:conf/iccv/DuanCNYQH21}, which creates adversarial examples by removing certain features from benign images. This makes the resultant images unnoticeable to humans but essential for DNNs to misclassify them. AdvDrop is more resistant to existing defensive mechanisms, e.g., AT~\cite{DBLP:conf/icml/AthalyeC018} and feature squeezing~\cite{DBLP:conf/ndss/Xu0Q18}, and paves the way for a new approach to assessing the robustness of DNNs. Focusing on the frequency domain, it deletes high-frequency information more often than low-frequency information.

{\color{black}The majority of adversarial attack strategies rely on label data, but face recognition (FR) authentication systems don't keep track of the label data for the target user. A similarity-based gray-box adversarial attack (SGADV) is put forth by Wang.\textit{et al.}~\cite{9667076} to address the shortcomings of current adversarial attacks on FR authentication systems. To implement adversarial attacks based on benchmark labels, a conditional binary cross-entropy (C-BCE) objective function is also designed as a baseline against FR-based authentication. Additionally, the experimental findings demonstrate that the pre-trained model is not secure in practice even if the database for face template storage is unharmed, demonstrating the importance of this research for raising the privacy threat to users. SGADV achieves effective attacks and a satisfactory time cost, but it is less efficient than FGSM~\cite{DBLP:journals/corr/GoodfellowSS14}, PGD~\cite{DBLP:conf/iclr/MadryMSTV18} and DeepFool~\cite{DBLP:conf/cvpr/Moosavi-Dezfooli16}, and no studies have been conducted on transferability.}

The above-mentioned attack methods, i.e., FeaCP~\cite{DBLP:journals/tifs/LiQHQLD21}, AdvLB~\cite{DBLP:conf/cvpr/DuanMQCYHY21}, HAG~\cite{DBLP:journals/tcyb/YangLDT20}, PCAE~\cite{DBLP:journals/tip/ZhangTLWT20}, IoU~\cite{DBLP:conf/cvpr/JiaS0Y21}, AdvDrop~\cite{DBLP:conf/iccv/DuanCNYQH21} and SGADV~\cite{9667076} represent the SOTA gradient-free attacks for DNNs.
\begin{itemize}
    \item
    FeaCP~\cite{DBLP:journals/tifs/LiQHQLD21} defends neural networks by limiting generated samples, finding close adversarial samples, and controlling the generation process while providing insight into their location.
    \item 
    AdvLB~\cite{DBLP:conf/cvpr/DuanMQCYHY21} is a new attack method that uses a search method and laser beams to attack objects from a distance. It is easy to use but not as secretive as other methods.
    \item
    HAG~\cite{DBLP:journals/tcyb/YangLDT20} generates adversarial examples in the Hamming space with no semantic connection to the original queries. Its perturbation is portable across settings.
    \item
    The IoU attack~\cite{DBLP:conf/cvpr/JiaS0Y21} is a method that uses past and current frames to reduce the accuracy of object tracking by generating perturbations.
    \item
    PCAE~\cite{DBLP:journals/tip/ZhangTLWT20} is an approach for crafting adversarial examples that do not require a target and do not have the issues of overfitting or lack of data.
    \item
    AdvDrop~\cite{DBLP:conf/iccv/DuanCNYQH21} generates adversarial examples by modifying image frequencies, making it challenging to defend against and a useful tool for testing DNNs' robustness. 
    \item
    {\color{black} SGADV~\cite{9667076} utilizes different similarity scores to generate optimized adversarial samples, effectively breaking FR-based authentication in both white-box and gray-box attacks.}
\end{itemize}

All of these methods have their own strengths and limitations. For example, AdvLB is highly temporally stable but less secretive. PCAE is target-free, but it does not rely on any classifier, so it is hard to evaluate its performance. AdvDrop is more robust to current defense methods, but it is relatively simple and focuses on the frequency domain.
{\small
\setlength{\tabcolsep}{2pt}
\setlength\LTleft{-0.1in}
\setlength\LTright{-0.1in}
\begin{table*}
	\caption{Gradient-free adversarial attacks}
	\label{table3}
	\setlength{\tabcolsep}{3pt} 
	\renewcommand\arraystretch{1.5} 
	\begin{tabular}{|p{1.1cm}|p{2.8cm}|p{1.2cm}|p{0.8cm}|p{0.9cm}|p{1.2cm}|p{4cm}|p{4cm}|}
    		\hline
    \centering{\textbf{Attack}} & \centering{\textbf{Short Description}} & \centering{\textbf{Optimizer}} & \centering{\textbf{Input}} & \centering{\textbf{Attack type}} & \centering{\textbf{Invisibility Metric}} & \centering{\textbf{Strength}} & \centering{\textbf{Weakness}} \cr \hline
      
    FeaCP \cite{DBLP:journals/tifs/LiQHQLD21} & A method to seek adversarial examples near the decision boundary. & Search-based & Conti-nuous & White-box  & $\ell_{\infty}$ 
    & Provide explainable hints on the locations of adversarial examples. Be generalizable in both computer vision and sentiment analysis fields.
    & Computationally expensive\\
     \hline
     
    AdvLB \cite{DBLP:conf/cvpr/DuanMQCYHY21} & Manipulate laser beam’s physical parameters for an adversarial attack.& Greedy Search based& Conti-nuous & Black-box  & $\ell_p$ 
    &Direct use of laser beams as a perturbation. High flexibility to actively attack any object, even at long distances, higher temporal stability, easy to deploy. 
    & It is less secretive than some samples generated by other methods, such as AdvCam. \\
    \hline

    HAG \cite{DBLP:journals/tcyb/YangLDT20} &  Craft adversarial examples for Hamming space search.& Search-based & Conti-nuous & Black-box & Hamming Distance  &  Successfully attack target hash models with low perceivability. High transferability at various settings, more transferable at different hash bit lengths for the same architecture. & Merging the perturbations from distinct hash digits to attack the model with a similar design did not result in a noticeable enhancement of performance.\\ 
     \hline
          
    AdvDrop \cite{DBLP:conf/iccv/DuanCNYQH21} &Craft adversarial examples by dropping existing information of images.& Drop-based & Conti-nuous & White-box, Black-box   & LPIPS  
    & AdvDrop is a completely different paradigm from previous attacks and is more robust to current defense methods. Computation cost and perceptual quality balance. 
    & By focusing on the frequency domain, a relatively simple strategy for eliminating information is used, which tends to lose high-frequency information.\\
     \hline
     
    PCAE \cite{DBLP:journals/tip/ZhangTLWT20} & Generate adversarial examples via principal component analysis.& DM based & Conti-nuous & White-box & $\ell_2$  & Not rely on any classifier, there will never be an overfit, and the impact of insufficient labeled data is limited. Competitive transferability. & 
    PCAE has poor performance due to its high complexity in the adversarial region. Compared with neural networks, kernel PCA is limited in its ability to approximate such complex manifolds. \\
     \hline
     
    IoU \cite{DBLP:conf/cvpr/JiaS0Y21} & Generates perturbations in sequence based on the predicted IoU scores from current and past frames. & Decision-based & Conti-nuous & Black-box & Cosine Distance & The IoU score is gradually reduced by using the smallest amount of noise, which in turn reduces the accuracy of the target task. 
    Be generalizable among different structures. & Performs less effectively than a white-box attack method CSA when applied to track, as it lacks access to the trackers' network architecture.\\
     \hline

   SGADV \cite{9667076} & SGADV uses different similarity scores to generate optimized adversarial examples, i.e. adversarial attacks based on similarity, effectively break FR-based authentication in both white-box and gray-box Settings. & Similarity-based & Conti-nuous & Gray-box, White-box & Cosine Distance, SSIM, LPIPS & High attack success and acceptable time cost. & The attack efficiency of SGADV is lower than that of FGSM, DeepFool and PGD, and no research has been carried out on the transferability.\\
     \hline
\end{tabular}
	\label{tab: gradient-free}
\end{table*}
}

\subsection{Adversarial Patch}
Although adversarial sample attacks, such as PGD~\cite{DBLP:conf/iclr/MadryMSTV18} and CAT~\cite{DBLP:journals/pami/WangLLL22}, can achieve a high ASR and undetectable perturbation effect, its generalization ability is generally poor to be used in the physical world, and a specific perturbation must be generated for each attack. As a result, adversarial patch attacks~\cite{DBLP:conf/cvpr/DuanMQCYHY21, DBLP:conf/eccv/ChengLCTCLZ22} come into view as a variant of the adversarial sample attack, in contrast to adversarial sample attacks where an attacker always aims to minimize the level of perturbation to avoid detection. In adversarial patch attacks, the attacker never again confines themselves to imperceptible changes. The attack generates an image-independent patch, which can be set anyplace in the image to attack a DNN-based image classifier and cause it to output a specified target class. Table~\ref{table4} collates the latest adversarial patch methods.


{\color{black}
The advantage of adversarial patching over adversarial sample attacks is that adversarial patching can be more targeted and effective in deceiving a deep learning model by creating a small, localized patch that can be placed at a specific location of an image. By contrast, adversarial sample attacks typically add noise or distortion to an entire image. Moreover, adversarial patching can potentially attack deep learning models in real-world scenarios where digital attacks are not possible, such as in physical security systems. This makes it a powerful tool for attackers to bypass ML-based security systems in many practical application scenarios. However, adversarial patching requires more effort and knowledge to create and raises important questions regarding the responsible development and deployment of ML systems.
}

\begin{figure}
    \centering
    \includegraphics[height=1.4in]{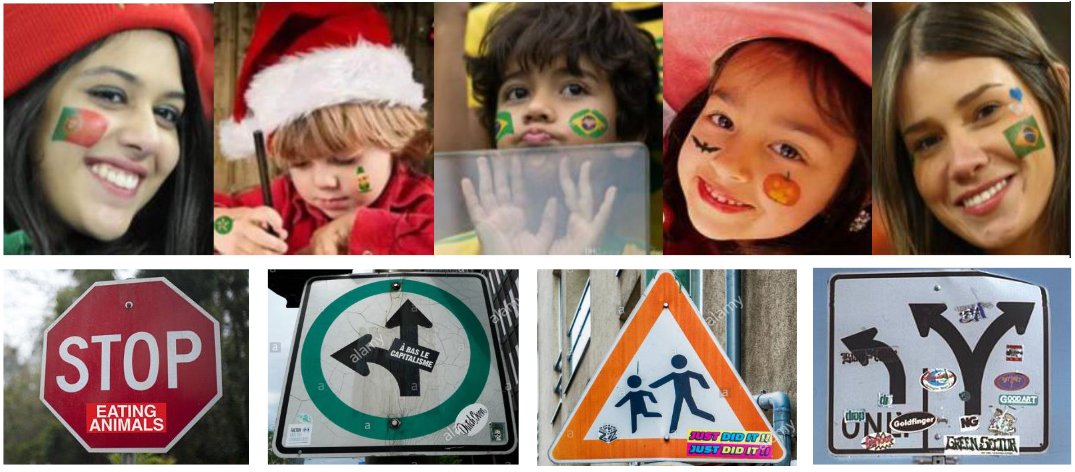}
    \caption{\color{black} Examples of stickers for faces and traffic signs~\cite{DBLP:journals/pami/WeiGY23}.}
    \label{sticker}
\end{figure}

FaceAdv~\cite{DBLP:journals/tifs/ShenYZXLH21} is a physical adversarial attack technique, which uses malicious stickers to fool face recognition applications. FaceAdv comprises a malicious sticker generator and a converter. The generator makes a variety of differently-shaped stickers (some examples of stickers are shown in Fig.~\ref{sticker}). At the same time, the latter applies the stickers digitally on human faces and provides the generator with attack results to enhance its efficacy. Despite changes in environmental conditions, FaceAdv can dramatically increase the success rate of avoidance and simulated attacks, showing robustness. 
However, the ``sticker'' attacks require a high degree of hardness to avoid early detection by humans. They also require the adversaries to physically access the target object for pasting the stickers, which may not always be possible.

Robust Physical Perturbation (RP2)~\cite{DBLP:conf/cvpr/EykholtEF0RXPKS18} is a generic attack algorithm that produces perturbations robust to varying angles and distances under different physical conditions. The perturbations are visible but inconspicuous and only perturb objects (e.g., road signs) without disturbing the object's environment. The algorithm utilizes a mask to transform the estimated perturbations into a graffiti-like form after sampling from a range of simulated physical dynamics. An attacker may then print out the resultant perturbations and apply them to the road sign under attack, resulting in a high rate of misclassification of the target by the road sign classifier, which might lead to catastrophic consequences.

Perceptual sensitivity is a crucial aspect of visual recognition systems. The more natural-looking the generated adversarial blocks, the more likely the attacks are successful. 

\begin{figure}
    \centering
    \includegraphics[height=2.2in]{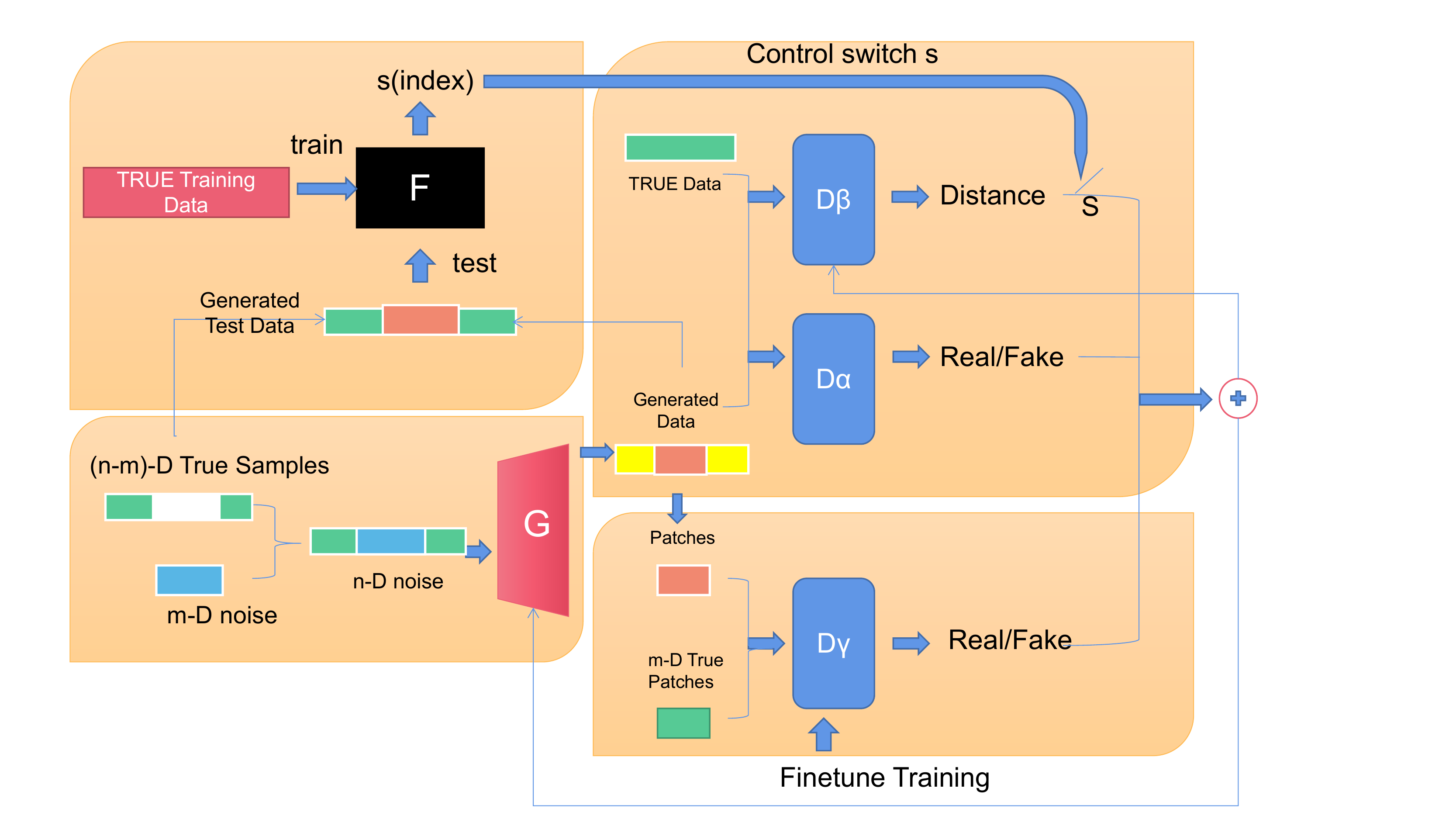}
    \caption{The framework of the MultiD-WGAN method. It is composed of a generator $G$, several discriminators, and a target classifier $F$.}
    \label{fig:mutidwgan}
\end{figure}

Perceptual-Sensitive Generative Adversarial Network (PS-GAN)~\cite{DBLP:conf/aaai/LiuLFMZXT19} uses perceptual sensitivity to improve the visual plausibility and attack capability of adversarial patches. It adopts a visual attention mechanism to capture the sensitivity of the spatial distribution and guide the localization of the adversarial patches for a stable attack effect. PS-GAN can also generate adversarial patches on-the-fly without the need to access the target model at the time of inference. This makes it a powerful tool for attackers looking to bypass visual recognition systems in real-world scenarios. 
Similarly, Wang \textit{et al.}~\cite{DBLP:conf/iscas/0250CWGHZW021} propose a data-driven, Muti-Discriminator Wasserstein GAN (MultiD-WGAN) algorithm based on GANs to craft adversarial patches that focus on the perceived sensitivity of the attacked neural network model, as shown in Fig.~\ref{fig:mutidwgan}. The algorithm enhances both the aggressiveness and authenticity of adversarial patches by utilizing multiple discriminators. The research demonstrates, theoretically and experimentally, a positive correlation between attack strength and attack capability.

Wang \textit{et al.}~\cite{DBLP:journals/tip/WangLBL22} devise a bias-based framework to produce generic adversarial patches that exploit attentional bias and perceptual bias to improve attack capabilities and increase the generality of adversarial patches. The framework uses style similarity~\cite{DBLP:conf/iccv/RutaMF0JFGC21} to extract a patch that comes before texture from a hard example with high model uncertainty and accounts for perceptual bias. An attentional bias is utilized by obscuring the same attentional patterns shared by models, which are identical for the same image across multiple models. This allows the created adversarial patch to be more transferable between models.

To tackle the hindrance of feeble disguise of adversarial patches and lengthy computational time, Bai \textit{et al.}~\cite{DBLP:journals/iotj/BaiLZ22} advance a procedure to bring forth inconspicuous adversarial patches exploiting singular images. The technique initially ascertains patch areas depending on the perceptual sensitivity of the target model, and then fabricates adversarial patches in a coarse-to-fine system, which utilizes multiscale producers and judges. The patches are urged to coordinate with the background image through adversarial training. At the same time, it still maintain a powerful attack aptitude. Experimentally, the proposed method has demonstrated formidable attack capability in white-box settings and good transferability for black-box situations, making it difficult to identify. 

To increase adversarial stealthiness and camouflage flexibility while maintaining adversarial strength, AdvCam~\cite{DBLP:conf/cvpr/DuanM00QY20} uses a style migration approach to achieve stealthiness and an adversarial attack technique to strengthen adversarial capabilities. The attacker specifies the target image, the target attack area, and the intended target style. AdvCam transforms significant adversarial perturbations into adjusted styles. The latter is then disguised in the target object or the background outside the target. Experiments conducted in both the digital- and physical-world scenarios show that AdvCam's faked adversarial samples are highly concealable and yet still effective in spoofing the latest DNN-based image classifiers.

The existing studies on adversarial patches, such as those developed in \cite{DBLP:conf/eccv/LiuWLCZY20, DBLP:conf/iccv/YuCXLWBM21}, have only looked at the robustness of single-spectrum (RGB or Thermal) models and have not evaluated multispectral models. They have only analyzed digital space perturbations and have not considered vulnerabilities in the physical world. To address these limitations, Kim \textit{et al.}~\cite{DBLP:conf/icassp/KimLR22} introduce a new framework for generating multispectral adversarial patches (MAP) using material emissivity (ME) loss optimization and cross-spectral mapping (CSM). The experiments show that the generated MAP can successfully attack multispectral personnel detectors in both physical and digital spaces, highlighting the need for further research in this area. Tarchoun \textit{et al.}~\cite{DBLP:conf/cw/TarchounAKM21}  investigate the influence of viewpoint on the efficacy of adversarial patches. In order to replicate the effect of perspective alterations in multi-perspective settings, they combine known adversarial patches with perspective geometric transformations. Experiments demonstrate that perspective substantially affects the efficacy of adversarial patches, which can sometimes drop substantially. This finding encourages academics to investigate the influence of viewpoint on adversarial attacks and reveals new options for adversarial defenses.

The above-mentioned methods represent the current SOTA in adversarial patch generation, with a focus on image recognition systems. These methods include FaceAdv, which crafts adversarial stickers to deceive facial recognition systems~\cite{DBLP:journals/tifs/ShenYZXLH21}; RP2, which produces perturbations that are robust to varying distances and angles under different physical conditions~\cite{DBLP:conf/cvpr/EykholtEF0RXPKS18}; PS-GAN, which uses perceptual sensitivity to improve the visual plausibility and attack capability of adversarial patches~\cite{DBLP:conf/aaai/LiuLFMZXT19}; MultiD-WGAN, which enhances both the aggressiveness and authenticity of adversarial patches by utilizing multiple discriminators~\cite{DBLP:conf/iscas/0250CWGHZW021}; AdvCam, which increases adversarial stealthiness and camouflage flexibility while maintaining adversarial strength by using a style migration approach and an adversarial attack technique~\cite{DBLP:conf/cvpr/DuanM00QY20}; and MAP, which generates multi-spectral adversarial patches to attack multi-spectral personnel detectors in both physical and digital spaces~\cite{DBLP:conf/icassp/KimLR22}.

Some of the methods, i.e., FaceAdv~\cite{DBLP:journals/tifs/ShenYZXLH21}, RP2~\cite{DBLP:conf/cvpr/EykholtEF0RXPKS18}, and PS-GAN~\cite{DBLP:conf/aaai/LiuLFMZXT19}, have demonstrated considerable strengths. For example, they are powerful tools for attackers looking to bypass ML-based security systems in the real world. They improve the visual plausibility and attack capability of adversarial patches. They can increase adversarial stealthiness and camouflage flexibility, while maintaining adversarial strength. They consider the impact of perspective in adversarial attacks.

On the other hand, they also have some weaknesses. Some methods require a high degree of hardness to avoid early detection by humans. Some methods require the attacker to physically access the target object to paste stickers, which may not always be possible. Moreover, they have not been thoroughly tested in multispectral models.
There is still much work to be done in terms of understanding the limitations and weaknesses of these methods, and developing effective countermeasures to protect against them.

\subsection{Transferability of Adversarial Attacks}
\label{transferability}
Transferability accounts for the ability of adversarial attacks to be applied to different models and datasets. Ideally, {\color{black}from the perspective of attackers,} an adversarial sample generated to deceive a specific model can also deceive other models. This is significant because it means that an attacker does not need to generate a new adversarial sample for each model or dataset it wants to attack, which can be time-consuming and computationally expensive. Suppose an adversarial sample is highly transferable. In this case, it is more likely to be successful in the real world, where the attackers might not know the specific model or dataset they are trying to attack. This makes the attack more powerful and can increase the ASR.

Goodfellow \textit{et al.}~\cite{DBLP:journals/corr/GoodfellowSS14} believe that a linear model is sufficient to produce adversarial instances in high-dimensional space, rather than relying on highly nonlinear features of DNNs. They explain that the reason for cross-model generalization is that adversarial examples are highly consistent with the weight vector of the model, and different models that carry out the same task learn similar functions.
Su \textit{et al.}~\cite{DBLP:conf/eccv/SuZCYCG18} evaluate eighteen DNN-based image classification models and concluded that untargeted attacks obtain higher transferability than targeted attacks. The transferability of adversarial samples was sometimes symmetric. They also discover that most adversarial samples from one model could only migrate among similar models. In addition, the transferability of the Visual Geometry Group (VGG) model~\cite{DBLP:journals/corr/SimonyanZ14a} performs far better than that of other models, making it a solid starting point for enhancing the black-box transfer-based attack.

According to~\cite{DBLP:journals/tip/ZhangTLWT20}, the transferability of adversarial cases is mostly due to the junction of adversarial area divisions and distinct classifier borders. The authors propose PCAE, a data-generated approach that can generate more transferrable adversarial instances than certain model-dependent methods. They also demonstrate that the target-free strategy might discover more transferrable adversarial scenarios and that target-free adversarial instances have greater transferability when model and/or dataset similarity is high.

Many other researchers have made efforts from various directions to strengthen the transferability of adversarial cases. AoA~\cite{DBLP:journals/pami/ChenHSYH22} focuses on attention heat maps, for which diverse DNNs provide comparable results, making AoA highly transferable. CAT~\cite{DBLP:journals/pami/WangLLL22} provides adversarial instances on ensemble models as opposed to a single model and has shown its efficacy in the black-box situation for improving transferability.

{\color{black}
Wang \textit{et al.}~\cite{9667076} study the impact of proxy representation learning on the transferability of adversarial attacks in gray-box graphs. The authors put forth that the proxy models need to maintain the consistency of node topology in the embedding layer and input layer, and use the SRLIM to maintain the topology of nodes mapped from a non-input space to a Euclidean embedding space. The proposed method realizes the improvement of the generalization and transferability of adversarial attacks.}

FaceAdv~\cite{DBLP:journals/tifs/ShenYZXLH21} use the collection of face recognition systems to train the sticker generator and update the loss function. ATTA~\cite{DBLP:conf/cvpr/WuSLK21} enhances the transferability of generated adversarial samples by adversarial transformations, which is a network of adversarial transformations that automates the distortion adjustment procedure. IGA~\cite{DBLP:journals/tcss/ChenLSL20} is a transferrable attack for unknown link prediction approaches. {\color{black}In IGA,} since the perturbations caused by GAE are universal and the attack is transferrable, the adversarial graph may still be successful in a variety of link prediction models. This is due to the fact that GAE can extract critical information from graphs in pursuit of link prediction.

As discussed above, several studies have been undertaken on the transferability of adversarial instances, or the ability of an adversarial attack to deceive different models or datasets. Studies have shown that the transferability of adversarial examples depends primarily on the closeness of the models or datasets under attack, and that untargeted attacks tend to have more transferability than targeted ones. Some studies have suggested approaches to enhancing the transferability of adversarial instances, including the use of ensemble models, attention heat maps, and adversarial transformations.

However, there is still room for improvement in the transferability of adversarial examples, particularly in creating more effective and efficient transferable attacks and in better understanding the underlying causes of transferability. Moreover, current methodologies tend to focus on image classification models. There is a need for more studies on other types of models, such as NLP. For example, Wallace \textit{et al.}~\cite{DBLP:conf/emnlp/WallaceSS20} build an imitation model like the victim model to study the transferability of a black-box machine translation system by using gradient-based attacks.

On the other hand, it is important to develop new approaches to defending against transferable adversarial attacks. One strategy is to use transferable adversarial examples to enhance the robustness of DNN models through {~\color{black}adversarial learning}, as suggested by~\cite{9157681}. The limitation of this strategy is that it requires a large number of transferable adversarial examples for training, which can be time-consuming and can adversely affect the prediction accuracy of the DNN model on natural examples due to an increased ratio of adversarial examples in the training dataset. 
Another strategy is to assemble various defenses into an ensemble solution to compensate for the lack of diversity in a single defense mechanism. For example, Deep Fusion Defense~\cite{9762031} employs three or five DNN models trained with different perturbation magnitudes to achieve superior performance in defending against transferable adversarial examples. However, the ensemble strategy can worsen the time and computational cost of the defense. Therefore, developing few-shot (i.e., using fewer training examples) solutions for defending against transferable adversarial examples is essential. 

Recently, Zhou \textit{et al.}~\cite{9612007} indicate that introducing randomness into neural network models can hinder the transferability of adversarial attacks. They also reveal that the transferability of adversarial attacks is closely related to the spread of DNN models distributed in the version space and the severity of adversarial attacks.
As a result, the robustness of a DNN model can be enhanced using any subset of the DNN models, or by adding a mild Gaussian noise to the weight of the pre-trained model. In addition, the robustness of adversarial ensemble training also has great potential for improvement combined with randomization techniques. 

Nowroozi \textit{et al.}~\cite{9747933} propose two current defense mechanisms to prevent the transferability of adversarial attacks. The first approach is to fine-tune the classifier with the most powerful attacks (MPAs) each time a shift occurs against a given adversarial attack. Another strategy relies on using the long short-term memory (LSTM)~\cite{DBLP:journals/neco/HochreiterS97} architecture, instead of CNN, as the target network, allowing the attacker to have less attack information than a previous system that did not know the target network architecture.
Moreover, the Luring Effect~\cite{DBLP:conf/ijcnn/BernhardMD21} is a new way for boosting the robustness of DNN models against black-box transfer attacks. The key concept is situated in conventional network security methods based on deception, which does not need a labeled dataset but needs access to the target model's predictions. Some other defense methods, such as Robust Soft Label Adversarial Distillation (RSLAD)~\cite{DBLP:conf/iccv/ZiZMJ21} and Dual-Domain based Defense (D2Defend)~\cite{DBLP:conf/ijcnn/YanLDBX21}, demonstrate their effectiveness in defending against transfer-based black-box attacks.

\subsection{Summary and Lessons Learned}
Adversarial attacks can be launched in several different ways, including gradient-based, optimization-based, and search-based methods. Gradient-based attack schemes are known for their high ASRs and good transferability. They still have limitations, such as high computational and time costs, as well as the issue of ``gradient saturation'', which reduces their effectiveness~\cite{DBLP:conf/ispacs/EndoT19}. Moreover, gradient-based methods are relatively easy to defend against~\cite{DBLP:conf/codaspy/LiuKK21}. Many existing defenses, such as obfuscated gradients~\cite{DBLP:conf/cvpr/HeRF19}, can effectively block most gradient-based attacks.

Constrained optimization-based attack methods have good transferability but are also known for their high computational and time costs, making them difficult to use in time-sensitive applications~\cite{DBLP:conf/icml/AthalyeC018}. Search-based attack methods are highly transferable and can be extended to other domains beyond image classification~\cite{DBLP:journals/apin/XuGLBLZ23}. However, for more complex data sets, searching for the optimal adversarial sample needs more iterations and high computational costs, and it is difficult to find the appropriate search start point. Currently, search-based methods are mainly applied to the optimization of other adversarial sample generation algorithms~\cite{DBLP:journals/csur/CaiXXWLP21}.

Adversarial attacks can be performed not only on images but also on other types of media, such as audio~\cite{9552529, 9797884,9599559}, text~\cite{9557814,9627642}, and wireless signals~\cite{9893902,9887932,9609969}. CNNs normalize all inputs to continuous signals, regardless of their semantic meanings. The major difference between images and other types of media is in their dimensions, which require adapted convolutional kernels for feature extraction. In this sense, the same adversarial attacks or their variations are largely applicable to inputs with media other than images.

Adversarial examples affect classification tasks and threaten other deep-learning tasks, such as regression. For instance, a neural network that solves the power allocation problem for a massive multi-input-multi-output (MIMO) system can be misled by FGSM~\cite{DBLP:journals/corr/GoodfellowSS14}, PGD~\cite{DBLP:conf/iclr/MadryMSTV18}, or UAP~\cite{DBLP:conf/cvpr/Moosavi-Dezfooli17} attacks, which were originally developed for image classification problems~\cite{9696006,9570812}. Adversarial attacks, such as FGSM, Iterative FGSM (I-FGSM)~\cite{DBLP:conf/iclr/KurakinGB17}, and PGD, have also shown effectiveness in linear regression tasks~\cite{9814883}. To this end, research progress made on adversarial attacks and defenses, e.g., regarding the image classification tasks, can be highly beneficial to other types of deep learning tasks.

Future investigation is expected to focus on reducing attack costs, improving transferability across different datasets and models, and extending to more deep learning domains. Additionally, it is important to strike a balance between perturbation visibility and attack success in order to develop effective adversarial attack methods.
 
{
\small
\begin{table*}
	\caption{Adversarial patching methods}
	\label{table4}
	\setlength{\tabcolsep}{3pt} 
	\renewcommand\arraystretch{1.5}
	\begin{tabular}{|p{2.3cm}|p{7.5cm}|p{7.5cm}|}
\hline
\centering{\textbf{Strategy}} & \centering{\textbf{Brief description}} & \centering{\textbf{Performance}}\cr \hline

    FaceAdv~\cite{DBLP:journals/tifs/ShenYZXLH21} & A physical attack that creates adversarial stickers to trick face recognition systems, made up of a sticker generator and a converter.
    & Keep robustness in dodging and impersonating attacks. It does not affect the performance of the face detector. Good transferability in both the digital and physical worlds.
    \\\hline
    
RP2~\cite{DBLP:conf/cvpr/EykholtEF0RXPKS18}	&
Samples from a distribution that simulates physical dynamics to project calculated perturbations into a graffiti-like shape.	& Perturbations that are robust against widely varying distances and angles can be generated under different physical conditions.\\\hline

PS-GAN~\cite{DBLP:conf/aaai/LiuLFMZXT19} & Perceptual sensitivity is used to improve the visual rationality and aggression of adversarial patches. A visual attention mechanism is employed to capture the sensitivity of spatial distribution.	& Guides the attack positioning of the adversarial patch for stable attack effects. Taking it a step further, PS-GAN can generate adversarial patches instantly.\\\hline

MultiD-WGAN~\cite{DBLP:conf/iscas/0250CWGHZW021} &Based on the idea of generating adversarial patches by GANs, the data-driven MultiD-WGAN is proposed, which can simultaneously enhance the offensive power and authenticity of adversarial patches by multiple discriminators. &Simultaneous enhancement of the aggressiveness and authenticity of adversarial patches through multiple discriminators.\\\hline

Bias-based Framework~\cite{DBLP:journals/tip/WangLBL22}	& Take advantage of perception bias and attention bias to improve attack capabilities.	& The resulting adversarial tiles allow for greater transferability between different models.\\\hline

Singular image based~\cite{DBLP:journals/iotj/BaiLZ22} 
& Determine patch location according to the perceived sensitivity of the victim model, encouraging patch alignment with background images through AT. &	It has strong attack ability in a white-box environment and good transferability for black-box environments, which is more difficult to detect and can also be applied to the physical world. \\\hline
AdvCam~\cite{DBLP:conf/cvpr/DuanM00QY20} & The style shift approach was used to achieve concealment, and adversarial strength was achieved using the technique of adversarial attack.	& AdvCam's forged adversarial samples in the digital and physical worlds are highly stealthy and still valid when it comes to spoofing the latest DNN image classifiers \\\hline
MAP~\cite{DBLP:conf/icassp/KimLR22} & MAP was optimized using cross-spectral mapping (CSM) and ME losses.	 & Successfully attack multispectral personnel detectors in both physical and digital spaces.\\\hline
Multi-perspective Environments~\cite{DBLP:conf/cw/TarchounAKM21} & Consider the effects of viewing angle changes in multi-perspective environments by integrating adversarial patches with perspective geometry transformations. & Viewing angles have a strong impact on the effectiveness of adversarial patches. In some scenarios, adversarial patches lose most of their effectiveness, opening up new opportunities for adversarial defense.\\\hline
\end{tabular}
	\label{tab:adv patch}
\end{table*} 
}

\begin{figure*}
    \centering
    \includegraphics[height=5.5in]{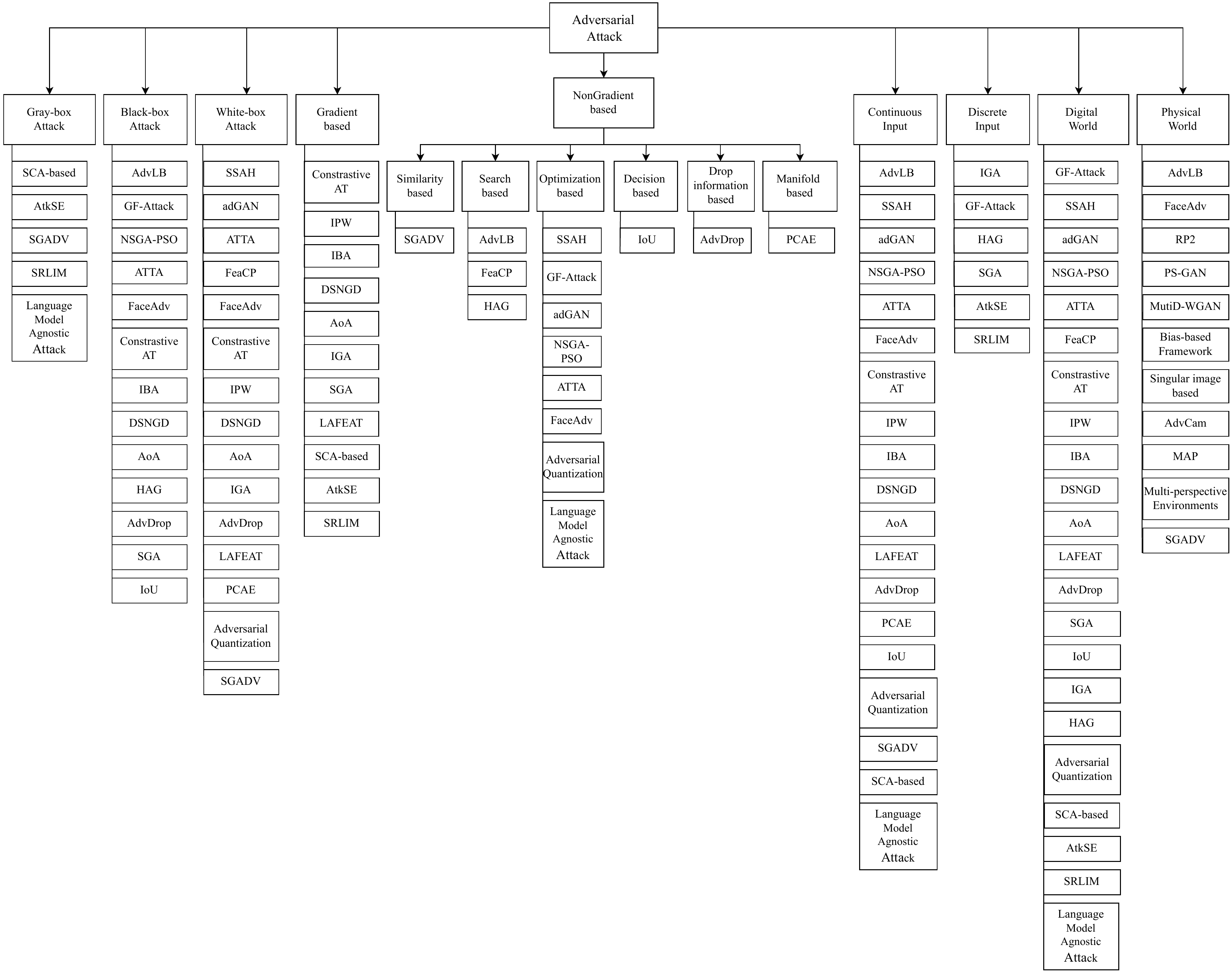}
    \caption{Anatomy of the recent breakthroughs in adversarial attacks since 2021. We divide adversarial attack methods into four categories, namely, black-box \& white-box attack, attack-based gradient or not, input characteristics and application areas. }
    \label{fig:attack}
\end{figure*}

\section{State-of-the-Art Adversarial Defense Techniques}
\label{defense}
To counteract adversarial attacks, various adversarial defense techniques have been devised. These techniques are designed to counteract specific attack techniques and range from specific defenses to general defense strategies. Deep learning models need to have the ability to counteract such attacks to maintain their accuracy and effectiveness. 

\subsection{Overview}
Typical adversarial defense techniques that have been developed include: 
\begin{itemize}
    \item 
{\textit{Adversarial Learning:} } Adversarial learning is a type of deep learning technique that involves training a model to improve its robustness against adversarial examples. One of its key techniques is adversarial training, which involves adding adversarial samples to the training process to improve the robustness of a DNN model. By continuously learning the features of adversarial samples, the model can better defend against attacks that involve adding subtle disturbances to input samples. This can improve the accuracy and effectiveness of the model in many real-world scenarios, where it may be exposed to adversarial samples designed to ``trick" it into making incorrect predictions. 
    \item
{\textit{Monitoring:} }
This is a strategy for identifying adversarial samples, which are input samples modified to cause a deep-learning model to make incorrect predictions. To monitor for adversarial samples, special models can be set up at key points in the system to identify these samples and provide early warning of potential adversarial attacks. This allows the system to take proactive measures to defend against attacks and maintain the integrity of the model's predictions. 
    \item
\textit{{Model Robustness Design:} }
This involves using specific filtering structures in the model to enhance its resilience against adversarial noise. Adversarial noise refers to the subtle disturbances that are attached to input samples to incite the model to produce false predictions. By designing the model to be more resistant to adversarial noise, it can better defend against these types of attacks and maintain the accuracy of its predictions. 
    \item
{\textit{Adversarial Perturbation Structure Destruction:} }
This involves using various strategies to attenuate the effect of adversarial noise and prevent attacks on the deep learning models. The strategies may include the use of filtering algorithms, noise structure destruction algorithms, and noise coverage algorithms in data stream processing. The goal is to achieve more resilience to adversarial noise, which is the subtle disturbance added to input samples to cause the model to make incorrect predictions. 
\end{itemize}

These four aspects are important to build robust, secure, and resilient deep learning systems, particularly in fields where the integrity and accuracy of the model's predictions are critical, e.g., in network security and finance. 

With the constant emergence of new and increasingly destructive adversarial attack methods, many research efforts have been devoted to exploring corresponding defenses. Current adversarial defense strategies can be categorized into two prevalent strategies: One strategy is based on detection and data preprocessing, and the other strategy improves adversarial robustness.

From a DNN model perspective, adversarial learning can be interpreted as gradient masking, which can refer to a class of techniques that hide model gradients from adversaries and prevent the adversaries from obtaining the correct gradients of the models, such as Graph Adversarial Training (GraphAT)~\cite{DBLP:journals/tkde/FengHTC21} and Robust CNN Training~\cite{DBLP:journals/tmm/AminiG20}, as will be delineated in Section~\ref{gradient masking}. More generally, gradient masking can refer to the outcome or effect of techniques designed to take other approaches (e.g., defensive distillation~\cite{DBLP:conf/iccv/ZiZMJ21}) to defend against adversarial attacks and resulting in obscured gradients of the network models under attack. 

\subsection{Adversarial Attack Detection and Data Preprocessing}

This type of defense method primarily detects adversarial attacks through technical means and pre-detected adversarial samples, or preprocesses the input data and destroys some key structures that constitute the adversarial samples.

\subsubsection{Adversarial Attack Detection}
As an adversarial defense method, adversarial sample detection has also attracted much attention from researchers. Given a sample, the goal is to directly detect whether it presents a threat.
In essence, the detector is trained on both the raw and adversarial sample datasets to identify adversarial samples by measuring the differences between them caused by the adversarial perturbation.
A prominent detection method, H\&G~\cite{DBLP:conf/iclr/HendrycksG17}, utilizes techniques, e.g., principal component analysis (PCA), softmax, and the reconstruction of adversarial images. These methods exploit the differences between original and perturbed images, but can be easily bypassed by attacks that target at them.

An algorithm called RObust SAliency (ROSA), presented by Li \textit{et al.}~\cite{DBLP:journals/tcyb/LiLY20a}, is an innovative technique for enhancing the robustness of FCN-based salient object recognition models against adversarial attacks. It works by adding universal noise to the input image, then using a two-part system to predict the saliency map of the image: A piecewise-masked component that disrupts adversarial noise patterns while preserving boundaries, and a context-aware refinement component that adjusts the saliency mapping by using contrast modeling. ROSA enhances  network robustness to attacks and performs comparably or better on natural images than current methods, which only focus on non-target attacks. The defensive performance against target attacks has yet to be explored.

\begin{figure}
    \centering
    \includegraphics[height=1.8in]{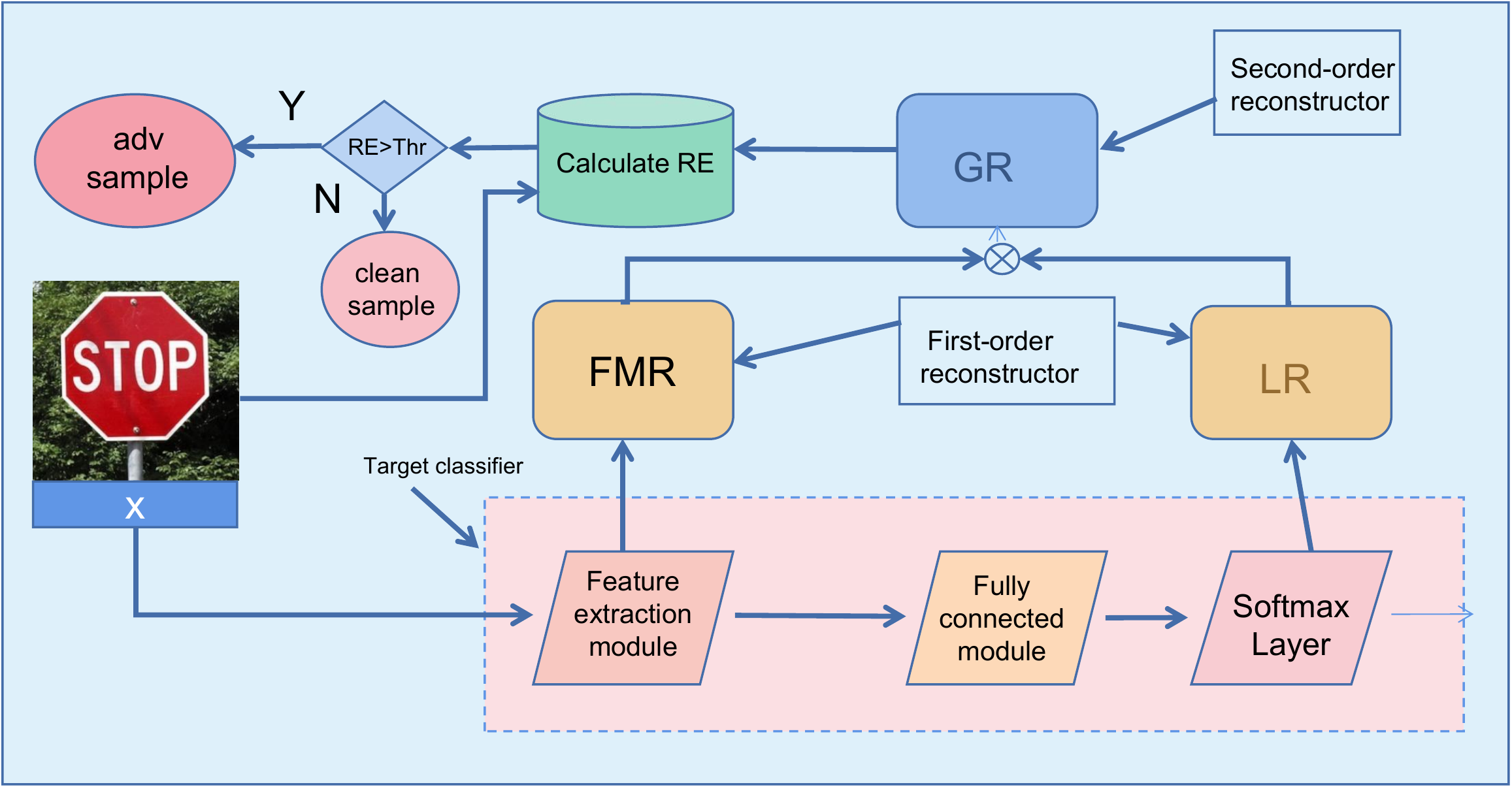}
    \caption{The CMAG workflow during the deployment phase. The CMAG is made up of FMR, LR, and GR, which are three reconstructors. When the reconstruction error exceeds the stated threshold, the sample is considered hostile.} 
    \label{cmag}
\end{figure}

Cascade model-aware generative (CMAG)~\cite{9449325} is an adversarial sample detection technique that consists of two first-order reconstructors, including a Feature-Map Reconstructor (FMR) and Logit Reconstructor (LR), and a second-order Global Reconstructor (GR). Rebuilding the logit and feature mapping, produces an interpretable representation of the final convolution layer. If the reconstruction error (RE) of a sample relative to GR exceeds the predetermined threshold, the sample is classified as adversarial. The process of a CMAG during deployment is shown in Fig.~\ref{cmag}. CMAG provides a new means to detect the presence of adversarial samples, which can accurately detect high-quality adversarial samples compared to existing generative model-based detection methods, e.g., Fence FGAN \cite{DBLP:conf/ictai/NgoWKPAL19} and UADD-GAN \cite{DBLP:journals/access/XieYWCL21}. The drawback of CMAG is that it utilizes a simplistic autoencoder as the generative model, and may not yield satisfactory results for complex datasets.

Zhang \textit{et al.}~\cite{9825039} state that adversarial attacks primarily attain their objectives by altering pixel values and that such attacks often insert perturbations in regions with high textures. In response, they presented a step-based deep learning network known as ADNet. ADNet is a DNN model for adversarial example detection by using steganalysis and attention mechanisms. It features an attention module, an adversarial attack attention module (AAAM), which pays additional attention to vulnerable parts throughout the process of feature learning, hence increasing the model's accuracy. To reduce the misclassification of regular samples in the detection phase, a special adversarial loss function has been designed to fine-tune the model, resulting in impressive outcomes. As an end-to-end model, {\color{black}ADNet does not rely on the extraction of high-quality features,} hence reducing the cost of human participation. However, it encounters the problem that the detection rate for adversarial samples is better than the classification accuracy for clean samples.

Freitas \textit{et al.}~\cite{9378303} indicate that adversarial vulnerability is a consequence of the excessive sensitivity of a model to good generalization features in the data. Since the model not only learns robust features but also information about non-robust features during training, models can be vulnerable despite maximizing accuracy. In light of this, the concept of adversarial vulnerability is extended to combine with prior human knowledge, and a new approach named UnMask is proposed, which is a framework for detection and protection against adversaries that relies on strong feature alignment. UnMask quantitatively evaluates the resemblance between the extracted and expected features, selects an adversarial perturbation to detect with a given similarity threshold, and protects the model by predicting the correct class that best fits the extracted features. The method highlights the advantage that even if an attacker can manipulate the predicted class labels by slightly changing the pixel values, simultaneously manipulating all the individual features that make up the image together is a more challenging task. Currently, UnMask only focuses on non-target attacks, and the defensive performance against target attacks has not been validated.

\begin{figure}
    \centering
    \includegraphics[height=1.3in]{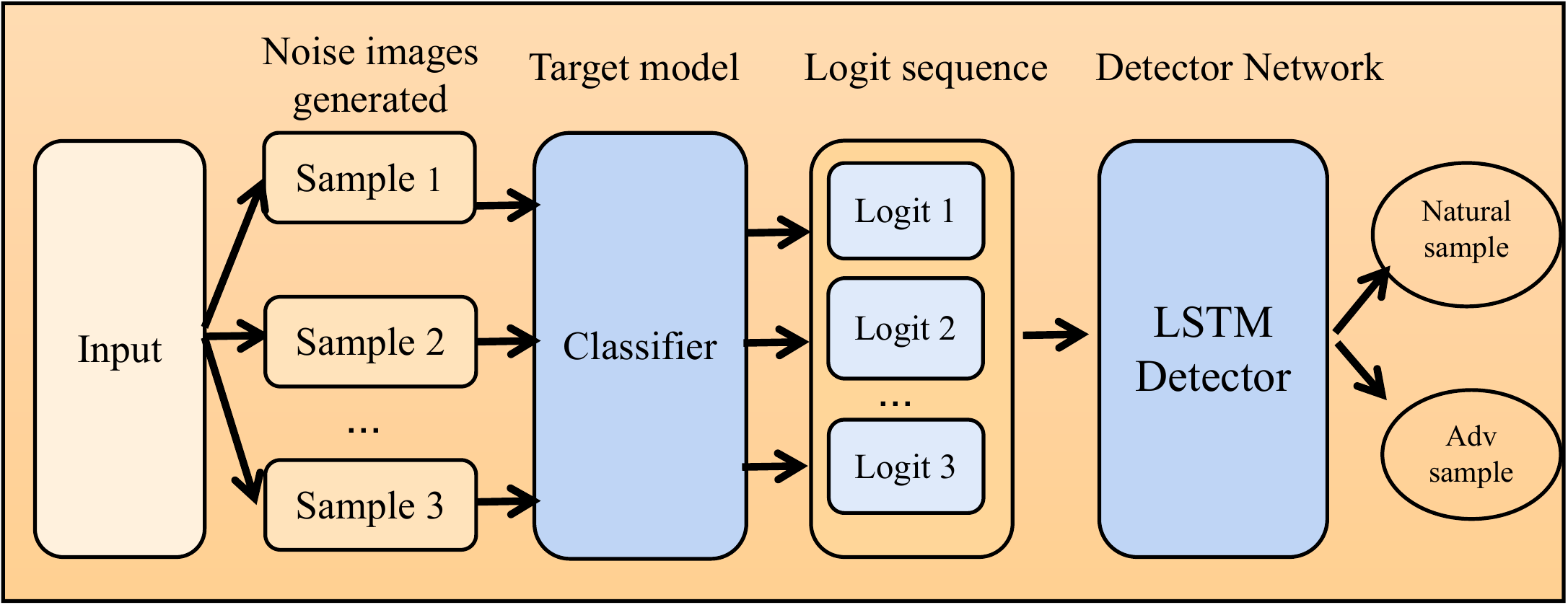}
    \caption{The overview of the architecture of the logit-based adversarial sample detection method. The model analyzes the original and adversarial examples that differ not only in the feature space but also in the semantic space, and then trains an LSTM network to learn the differences of the logit distribution in the semantic space.} 
    \label{logits}
\end{figure}

Although most adversarial defense detection methods have offered satisfactory results, some problems are yet to be addressed, including excessive reliance on target models, difficulty in resisting transfer attacks, relatively weak generalization capabilities, and so on. Model-independent methods to detect adversarial inputs are developed by Wang \textit{et al.}~\cite{9506292}. The  architecture is reviewed in Fig.~\ref{logits}. As the primary architecture of the detector, an LSTM network is trained to capture variations in the logit sequence distribution. They examine the original and adversarial cases, which vary not only in the feature space but also in the semantic space, and then train an LSTM network to discover any discrepancy in the logit distribution in the semantic space. They offer a logit-based adversarial sample detection strategy that is very flexible, simple to implement in all pre-trained models, and has robust detection resistance against both black-box and white-box attacks.

To detect arbitrary adversarial attacks without access to reference spectrographs and adversarial perturbations, Esmaeilpour\textit{ et al.}~\cite{9052913} propose a regularized logistic regression model to distinguish the eigenvalues of malicious spectral graphs from legitimate spectral graphs. They reveal that the manifolds of the adversarial samples are distant from the natural and noisy instances that are slightly disturbed by Gaussian noises. They use the eigenvalues of the legal examples and adversarial examples to train a logistic regression to find the decision boundary between them. This detector's main obstacle is its sensitivity to intra-class sample similarity, particularly in the multi-classification problem of black-box attacks. 

Existing methods, e.g., \cite{DBLP:conf/cacml/LiuWD22,DBLP:journals/access/Wang22e}, focus on the visual field, and cannot detect adversarial examples in the radio signal field, which is an important domain due to the ubiquitous networks. In response to the adversarial attacks in the realm of radio signals, Xu \textit{et al.}~\cite{9477416} describe a novel adversarial sample identification method by means of the integration of many features. They also provide a framework for creating adversarial samples, collecting local intrinsic dimension (LID) characteristics and constellation diagram (CD) characteristics, and recognizing adversarial samples. The framework produces the values of each layer for both normal and adversarial examples of the model. Then it computes the LID eigenvalues of the instance by estimating the maximum probability of a defined range of neighborhoods. The CD eigenvalues are computed simultaneously using the range characteristics and density features of the CD distribution. A logistic regression classifier is trained using several feature fusion values to identify adversarial samples. Experiments demonstrate that the suggested approach can reliably identify hostile radio signals. However, the performance degrades slightly when the perturbation is less than 10\%. The reason is that the perturbations are very small, and the features are inconspicuous between the normal and adversarial examples.

All these methods, such as~\cite{DBLP:journals/tcyb/LiLY20a} and~\cite{9477416}, represent the latest defense techniques that detect and defend against adversarial attacks. 
Adversarial sample detection is a typical defense method that aims to detect whether a sample is undergoing an adversarial attack directly.
Logistic regression and deep learning are often used to classify adversarial and non-adversarial samples.
On the one hand, adversarial sample detection methods can effectively detect and defend against adversarial attacks, in particular, black-box and white-box attacks. They are versatile and widely applicable to pre-trained models. 
On the other hand, adversarial sample detection methods may require large amounts of training data and high-capacity models with high computational overhead. They may have poor generalization capabilities and may be vulnerable to transfer attacks. Moreover, some of the methods are sensitive to intra-class sample similarity, especially in the multi-classification problem of black-box attacks.

{\small
\setlength{\tabcolsep}{2pt}
\begin{table*}
\caption{Adversarial attack detection methods}
	\label{table5}
	\setlength{\tabcolsep}{3pt} 
	\renewcommand\arraystretch{1.5} 
	\begin{tabular}{|p{1.5cm}|p{3.7cm}|p{1.5cm}|p{1.5cm}|p{3.5cm}|p{3cm}|p{1.8cm}|} 
    		\hline
		\centering{\textbf{Defense}} & \centering{\textbf{Brief description}} & \centering{\textbf{Similarity Measure}} & \centering{\textbf{Evaluation}} & \centering{\textbf{Strength}} & \centering{\textbf{Weakness}} & \centering{\textbf{Impacted Area}} \cr \hline
		ROSA \cite{DBLP:journals/tcyb/LiLY20a} & Shuffling pixels of an image and introducing some new universal noise to disrupt the adversarial perturbation then learns to predict the saliency mapping of the input image. & Energy function of low-level similarity & Precision, Recall, MAE, PR curves, $F_{\beta}$-measure& Significantly improve the backbone network robustness to adversarial attacks; show comparable or better performance on natural images. & Only focus on non-target attacks.& 
		computer vision \\ \hline
		
       CMAG \cite{9449325} & Detecting generative adversarial examples that are aware of the cascade model and demonstrate to humans what the model perceives by reconstructing the logit and feature maps of the final convolution layer. & SSIM & Detection Accuracy&  Detect high-quality adversarial examples effectively and be more interpretable, providing a new perspective on the existence of the adversarial example.&The generative model is a simple autoencoder, whose performance is not suitable for a complex dataset. & 
        computer vision \\ \hline
        
       ADNet \cite{9825039} & A steganalysis-based deep learning model in which an attention module is incorporated to focus more on susceptible areas during feature learning. & $\ell_2$ distance& Detection Accuracy& End-to-end, without manually extracting features.& Detection rates on normal images are lower than those on adversarial images created by C\&W. & 
        computer vision \\ \hline

       Logit-based adversarial detection \cite{9506292} & Trained an LSTM network to learn variations in logit distribution in semantic space and suggested a logit-based adversarial example identification technique. &  logit & ROC, AUC & Detect the logit sequence differences between the original and adversarial examples, model-agnostic with strong generalizability.& Contrasting tests with the other methods are missing. & 
        computer vision \\ \hline

       UnMask \cite{9378303} & Based on robust feature alignment, detect adversarial perturbations by selecting a similarity threshold and safeguard the model by predicting an accurate class. 
       & Jaccard Similarity
       & ROC, TP, FP& 
       It is significantly more difficult for an attacker to modify all of the different characteristics that make up the image at the same time. & Only untargeted attacks, gray-box. &
        computer vision \\ \hline

        Regularized logistic regression model \cite{9052913} & A  regularized logistic regression model for discriminating eigenvalues of malicious spectrograms from benign ones. & Chordal Distance & AUC & Be able to detect any adversarial attack when no reference spectrogram or adversarial perturbation is available. & Significant sensitivity to intra-class sample similarities, especially for black-box multi-classification.&     environmental sounds\\ \hline    

       LID\&CD \cite{9477416} & A detector based on multi-feature fusion, including generating, extracting LID\&CD features and detecting adversary.
       & LID & Detection Accuracy& Innovatively detect adversarial examples in the radio signal domain instead of the vision domain.& When the perturbation is less than 10\%, the performance is slightly decreased.
       & radio signals field\\ \hline
    \end{tabular}
\end{table*}
}

\subsubsection{Data Preprocessing}
For defense methods for preprocessing input samples, classical research methods include PixelDefend~\cite{DBLP:conf/iclr/SongKNEK18}, feature compression~\cite{DBLP:conf/cvpr/LiuLLXLWW19} and random transformations~\cite{DBLP:conf/sp/LingJZWWLW19}, etc. Table~\ref{table6} classifies the recent data preprocessing methods from the perspectives of attacked objects, input types, and invisible metrics.

\begin{figure}
    \centering
    \includegraphics[height=1.7in]{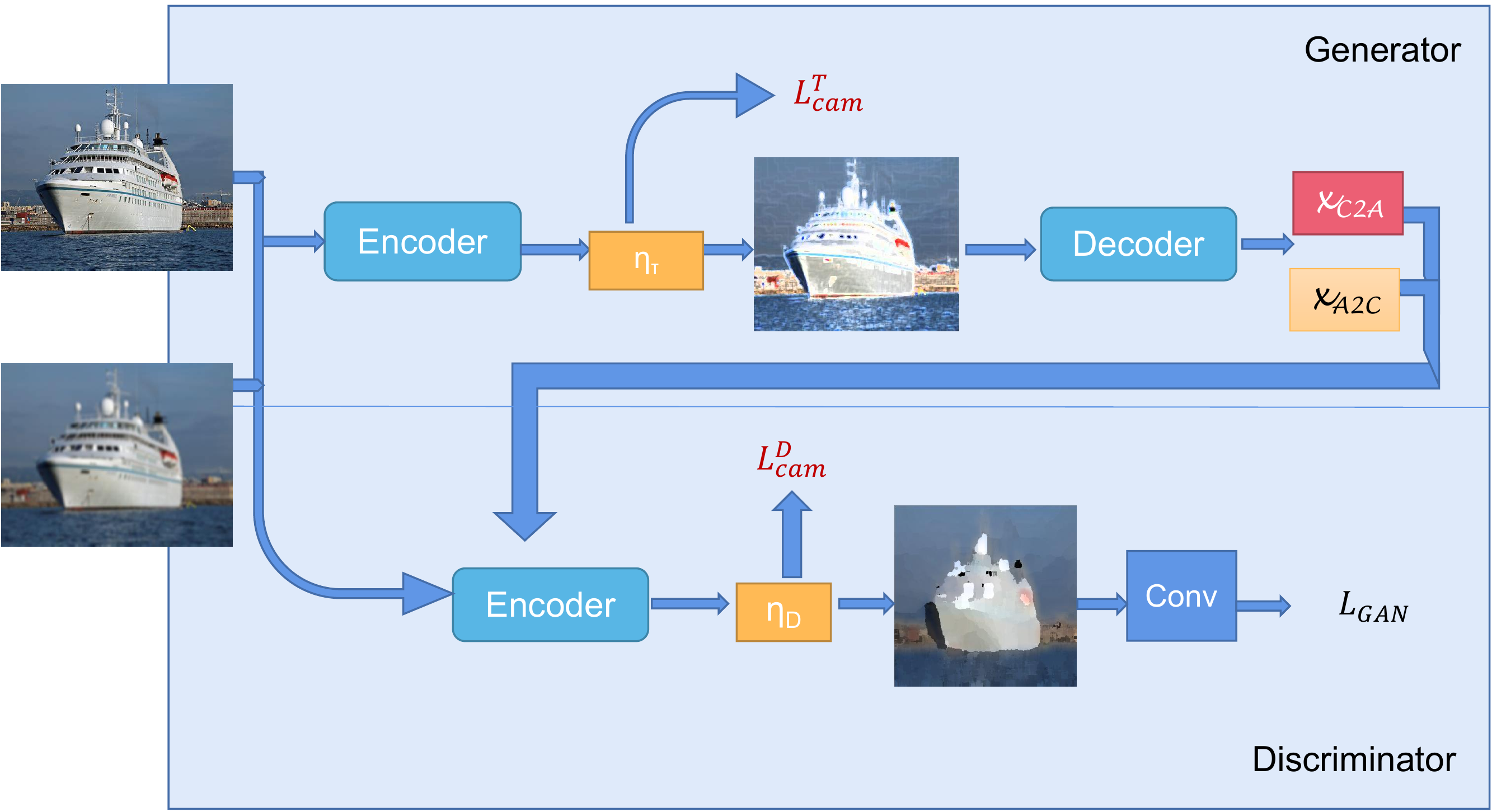}
    \caption{Overview of the Generator and Discriminator in CAP-GAN. To accomplish adequate purification under cycle-consistent learning, the framework trains the purification model to increase robustness and employs the standard GAN training approach to purify adversarial inputs by considering pixel-level and feature-level consistency.} 
    \label{capgan}
\end{figure}

Kang \textit{et al}.~\cite{DBLP:conf/ijcnn/KangTCK21} propose a purification model named CAP-GAN (Cycle-consistent attentional purification GAN), aimed at decreasing the impact of adversarial perturbations by transforming the input. The design of CAP-GAN is plotted in Fig.~\ref{capgan}. The framework trains the purification model to enhance the pre-trained model's robustness in image classification. It uses the standard GAN training process to clean up adversarial inputs by balancing pixel-level and feature-level consistency via cycle-consistent learning for effective purification. On the CIFAR-10 dataset, CAP-GAN surpasses other preprocessing-based defenses, such as the JPG compression method developed in~\cite{DBLP:journals/corr/DziugaiteGR16}, in both black-box and white-box configurations.

To address the issue that preprocessing-based defenses are typically sensitive to the error amplification effect, Zhou \textit{et al.}~\cite{DBLP:conf/iccv/Zhou0P0WYL21} suggest a self-supervised adversarial training mechanism in the class activation feature space to eliminate adversarial noise. Given adversarial assaults in the realm of radio signals, they first produce adversarial instances by substantially disrupting natural examples' class activation features. Then they train a denoising model, also known as the class activation feature-based denoiser (CAFD), by minimizing the disparities between the adversarial and normal examples in the class activation feature space. Consequently, antagonistic noise may be minimized. In comparison to prior approaches, such as the adversarial perturbation elimination GAN (APE-GAN) method developed in~\cite{DBLP:conf/icassp/JinSZDZ19}, the method developed in~\cite{DBLP:conf/iccv/Zhou0P0WYL21} considerably improves adversarial robustness, particularly against unexpected adversarial and adaptive attacks. However, for white-box attacks, the protection capability of the defense model is compromised as the defense model is completely visible to the attacker. In this sense, the defense against white-box adaptive attacks needs to be strengthened.

With the aim of addressing both adversarial examples deliberately created to cause harm and inputs that fall outside of the expected distribution, Wei \textit{et al.}~\cite{DBLP:journals/tdsc/WeiL21} develop XEnsemble, a diversity ensemble verification technique. To limit the harm caused by malicious or incorrect data inputs, XEnsemble, an input-output model verification ensemble protection technique, may automatically examine every input to the prediction model. To protect DNN prediction models from adversarial examples and out-of-distribution inputs, XEnsemble uses a variety of data cleaning strategies, including rotation, color-depth reduction, local spatial smoothing and non-local spatial smoothing (NLM), to generate diverse input denoising verifiers, implements an ensemble learning approach to protect the DNN model from deception, and offers a set of algorithms for combining the input and output verifications. XEnsemble performs well in recognizing out-of-distribution inputs and protecting against adversarial samples. In order to make XEnsemble more resistant to internal attacks on the defensive system, the team have planned to randomize the input denoising integration layer and the output model validation layer, and also to generalize the XEnsemble technique to additional media types, including text, video, and audio.

By taking inspiration from the robustness domain~\cite{DBLP:conf/nips/BastaniILVNC16, DBLP:conf/icml/EtmannLMS19,DBLP:conf/nips/JordanLD19}, Zhu \textit{et al.}~\cite{DBLP:conf/ijcnn/ZhuWZ21} examine adversarial training from the standpoint of data-to-decision boundary distance. They introduce Saliency Adversarial Defense (SAD). This batch normalization strategy can achieve adversarial robustness without adversarial training by processing inputs through their saliency map and changing the Batch Normalization (BN) statistics. Compared to adversarial training, SDA is efficient in protecting against various forms of white-box and black-box attacks. It is generally anticipated that fine-tuning the processed data and adjusting the saliency map intensity based on the sample could lead to further improvement in performance.

When reducing adversarial noise, many existing model-agnostic defenses lose key image content, resulting in low classification accuracy on benign images. In this regard, Mustafa \textit{et al.}~\cite{DBLP:journals/tip/MustafaKHSS20} put forth an image restoration approach by using super-resolution, which projects off-the-manifold adversarial instances into the native image manifold. The method is a model-independent defense mechanism that enhances an image by selectively adding high-frequency elements meanwhile canceling out any unwanted noise introduced by the attacker. Not only does the approach defend against attacks, but enhances image quality while keeping model performance constant on clean images as well. Even when the attack model and attack type are unknown, the method performs better than white-box settings in terms of robustness.

Despite its importance in CNN prediction, the accurate recovery of input image structures has been generally overlooked in existing adversarial defensive systems. Yan \textit{et al}.~\cite{DBLP:conf/ijcnn/YanLDBX21} develop the Dual-Domain based Defense (D2Defend), which recovers both low and high-frequency picture structures in the spatial and transform domains, while eliminating adversary distortions. Unlike previous input-transformation approaches, such as a feature distillation method developed in~\cite{DBLP:conf/cvpr/LiuLLXLWW19}, D2Defend uses bilateral and short-time Fourier transform (STFT) filtering to divide the input image into edge and texture feature layers. D2Defend is simple to develop and model-independent. It has been demonstrated to outperform existing adversarial defense approaches, particularly in high-attack scenarios. D2Defend's loss in clean accuracy is likewise judged acceptable and more stable than other defense methods.

These data processing-based methods~\cite{DBLP:conf/ijcnn/KangTCK21}--\cite{DBLP:conf/ijcnn/YanLDBX21} can be broadly categorized as model-specific or model-agnostic defenses~\cite{DBLP:journals/corr/abs-2207-08089}, and input-transformation or input-verification based defenses~\cite{DBLP:conf/ipccc/QinY21}. One commonality among these methods is that they aim to strengthen DNNs' resistance to adversarial attacks without significantly degrading their performance on non-attacked (clean) inputs. Some of the strengths of the methods discussed include their ability to defend against both white-box and black-box attacks, their model-agnosticism, and their ability to recover image structures of the input. Additionally, some of the methods are able to improve image quality and maintain model performance on clean images, and some of the methods are easy to deploy. On the other hand, some of the methods are found to be weak in defending against white-box adaptive attacks, and others are found to cause gradient masking based on characteristics. Also, some of the methods proposed require fine-tuning the processed data and adjusting the significance map intensity related to the sample to improve the performance further.

\subsubsection{Summary}
Adversary detection approaches offer a strategy to defend against adversarial attacks in recent years. These methods attempt to detect adversarial samples in the input data and reject them, rather than modifying the original models and inputs. The advantages of adversarial detection include the ability to be used in combination with other defense methods and the ability to analyze whether the inputs contain an adversarial sample when the results of the baseline and robust classifiers do not agree.

On the other hand, adversarial detection methods have limitations. Some methods, such as LID\&CD~\cite{9477416}, ADNet~\cite{9825039}, and UnMask~\cite{9378303}, extract feature dimensions to detect adversarial samples. Other methods, such as CMAG~\cite{9449325}, detect adversarial examples based on sample reconstruction comparisons. However, LID\&CD~\cite{9477416} has insignificant detection performance when the perturbation is small. ADNet~\cite{9825039} has a better detection success rate for adversarial samples than for clean samples~\cite{9052913}, and higher sensitivity for intra-class samples. They may also be bypassed by attackers who understand the detection mechanism. 
In the future, a combination of detection and defense is expected to be a promising direction to pursue.

{\small
\setlength{\tabcolsep}{2pt}
\begin{table*}
\caption{Data preprocessing methods for adversarial example detection and defenses.}
	\label{table6}
	\setlength{\tabcolsep}{3pt} 
	\renewcommand\arraystretch{1.5} 
	\begin{tabular}{|p{1.23cm}|p{1.3cm}|p{3.9cm}|p{0.8cm}|p{1.2cm}|p{4cm}|p{4cm}|} 
    \hline
    \centering{\textbf{Defense}}  & \centering{\textbf{Attack type}} & \centering{\textbf{Brief description}} & \centering{\textbf{Input}} & \centering{\textbf{Invisibility Metric}} & \centering{\textbf{Strength}} & \centering{\textbf{Weakness}}  \cr \hline

   XEnsemble \cite{DBLP:journals/tdsc/WeiL21}  &White-box, OOD & Improve DNN robustness against OOD inputs and adversarial examples by using diversity ensemble verification.& Image &  $\ell_p$ & Automatically validate any input to the predictive model, be attack-agnostic. Superior in robustness and defensiveness, with a high defense rate of adversarial samples and a high detection rate of OOD inputs. & Need to include randomization at the input denoising integration layer and the output model validation layer to increase resistance to internal attacks, as well as expand to new media. \\ \hline
    
   SAD \cite{DBLP:conf/ijcnn/ZhuWZ21}   &White-box, Black-box &Achieve adversarial robustness via analyzing inputs using saliency maps and updating BN statistics without AT.& Image & $\ell_2$ &  Widen the average distance between the processed data and the updated decision boundary, significantly smooth the landscape, and be more effective than AT. Reduce the training time significantly, not rely on gradient masking. & Need to fine-tune the processed data and adjust the significance map intensity related to the sample to further improve the performance. \\ \hline
    
   Image super-resolution \cite{DBLP:journals/tip/MustafaKHSS20}  & White-box, Black-box, Gray-box &  A super-resolution-based image restoration technique that projects off-the-manifold adversarial samples into the benign image manifold. & Image &  $\ell_1$ & 
   No need for training or tuning many hyper-parameters. Do not cause gradient masking. Perform well for both black-box and white-box attacks. Support unknown attacks. & Its robustness against white-box attacks is weaker than that in the gray-box settings.\\ \hline
    
    CAP-GAN \cite{DBLP:conf/ijcnn/KangTCK21}   & White-box, Black-box & Use the pixel-level and feature-level consistency for GAN's cycle-consistent learning to achieve adequate purification.& Image  & KL & The introduction of feature-level items fairly enhance model robustness. In both black-box and white-box conditions, CAP-GAN beats alternative preprocessing-based defenses on the CIFAR-10 dataset. & Adversarial interference is mitigated at the cost of removing important information from clean images, making DNN models less accurate for clean samples.\\ \hline
    
    CAFD \cite{DBLP:conf/iccv/Zhou0P0WYL21}  & White-box, Unseen Attack & Devise a self-supervised AT technique in the class activation feature space to eliminate adversarial noise. & Image & CAFA & Compared to previous SOTA methods, the confrontation robustness is significantly enhanced, especially against unknown adversarial and adaptive attacks. & For white-box attacks, the defense model is completely leaked to the adversarial, and the protection capability of the defense model is destroyed. The defense against white-box adaptive attacks needs to be strengthened. \\ \hline
    
  D2Defend \cite{DBLP:conf/ijcnn/YanLDBX21}   & White-box & Maintain the essential high-frequency image structure and filter out adversarial perturbations. & Image & SSIM & It is independent of DNN models and deploy-friendly. Good transferability among different commonly used networks and adversarial attack methods. The clean accuracy degradation is acceptable. More stable performance. & The defense effect under C\&W attack is not optimal. \\ \hline
     \end{tabular}
\end{table*}
}

\subsection{Robustness Enhancement for Deep Learning Models}
Current methods for defending against adversarial attacks focus primarily on improving model robustness. This goal is accomplished by incorporating regularizers into the model's loss function to make it more smooth. In other words, the gradient is Lipschitz continuous~\cite{DBLP:journals/jota/MarchiT22}. The goal is to make the model less sensitive to irrelevant variations in the input and off-manifold perturbations through effective regularization. The recent studies improving adversarial robustness can be broadly categorized into four main layers that regularization can be deployed: The input layer, middle layer, output layer, as well as across the layers.

\subsubsection{Regularization on Input Layer}
\label{input layer}
The essence of input layer regularization is to strengthen the generalization ability of neural networks through data enhancement. This can process input images, such as projection to non-adversarial manifold~\cite{DBLP:journals/ijcv/AnirudhTKB20} and image conversion~\cite{DBLP:journals/nca/XiaoZYWZW22}. It can also train models by adding noise or using pseudo-labels in a semi-supervised learning mechanism~\cite{DBLP:journals/corr/SmilkovTKVW17}. Tables~\ref{table7} and~\ref{table7-2} summarize the latest breakthroughs in the regularization of the input layer, divided into several aspects, including strategy, input, attack type, and inv-metric.

\subsubsubsection{Noise Perturbation}
Noise Perturbation means processing input samples by injecting some mask or noise. Adversarial Margin Maximization Networks (AMM) are provided by Yan \textit{et al.}~\cite{DBLP:journals/pami/YanGZ21} as a learning-based regularization, which substitutes an adversarial perturbation for the geometric margin. By carefully crafting aggregation and shrinkage algorithms, AMM directly improves the classification margin. For many DNN designs, AMM greatly enhances test set precision while maintaining training set accuracy, demonstrating increased generalization power. Nevertheless, the computational cost of AMM increases due to the usage of repeated updates to estimate the classification margin and the calculation of high-order gradients during optimization.

By integrating random differentiable picture transformations when training DeepFake models~\cite{DBLP:journals/csur/MirskyL21}, Yang \textit{et al.}~\cite{9533868} present a transformation-aware adversarial face generation strategy to increase the defense capability against GAN-based DeepFake variants in the black-box situation. The technique can persistently yield more distortions in simulated face images, making it simpler to identify generated counterfeit images and videos, which are independent of models and data. The process follows the same steps under all settings. One potential downside is that the process of model training can be time-consuming.

Yu \textit{et al.}~\cite{DBLP:journals/tip/YuLLYZ21} propose Progressive Diversified Augmentation (PDA), which increases the resilience of DNNs by gradually infusing different adversarial sounds in the training phase. This improves the system's overall resilience against adversarial examples and common corruption. Not only does PDA employ gradient information to make adversarial noises with negligible additional time cost, but uses a progressive schedule to vary the magnitudes of inputs during the training process as well. However, attaining robustness concurrently with PDA for white-box adversarial attacks restricted to numerous locations may be hard due to the incompatibility of different adversarial robustness.

To enhance the robustness of Meta Reinforcement Learning (MRL), Chen \textit{et al.}~\cite{DBLP:conf/ijcnn/ChenCW21} propose adversarial Meta Reinforcement Learning (adMRL) to generate adversarial attack instances by using an adversarial GAN (adGAN) and leverage the generated examples to enhance the Meta Reinforcement Learning (MRL) algorithm's robustness. AdGAN and MRL can obtain good results by optimizing a minimax objective function during training. Building on model-agnostic meta-learning, the agents can learn the initial parameters with better generalization ability. Thus, when facing an unknown new task, the agents can learn to counteract these ``bad'' samples. However, the experimental attack methods only test on FSGM and random noise attacks, which may not be sufficient to fully demonstrate the robustness of the method.

Deep spiking neural networks (SNNs) modeled after the brain have gained popularity owing to their ability to reduce the power consumption of deep learning applications. In their study, Kundu \textit{et al.}~\cite{2021HIRE} present a spike-timing-dependent back-propagation (STDB)-based SNN training method to better leverage the inherent robustness of SNNs. Instead of continually displaying the same image, this approach makes use of the temporal phases of SNN training to input several noisy copies of the same image, hence decreasing the requirement for intermediate gradient storage and eliminating superfluous training time. The enhanced robustness of this method is not a result of gradient masking, and it demonstrates outstanding performance under black-box and white-box attacks, with a negligible loss in clean accuracy.

\subsubsubsection{Adversarial Learning}
{\color{black}Adversarial learning aims to improve network security by improving the robustness of machine learning algorithms. As the main branch of adversarial learning, adversarial training,} which involves training DNN models with adversarial examples, is another strategy for strengthening DNN robustness. Adversarial examples are produced with known adversarial attack algorithms, e.g., those described in Section~\ref{attack}. Adversarial training can also be viewed as a special case of data augmentation that differs from traditional methods. Rather than introducing randomly transformed examples to improve model generality, {\color{black}adversarial learning} introduces adversarially perturbed data to strengthen the model's robustness.

\begin{figure}
    \centering
    \includegraphics[height=2.7in]{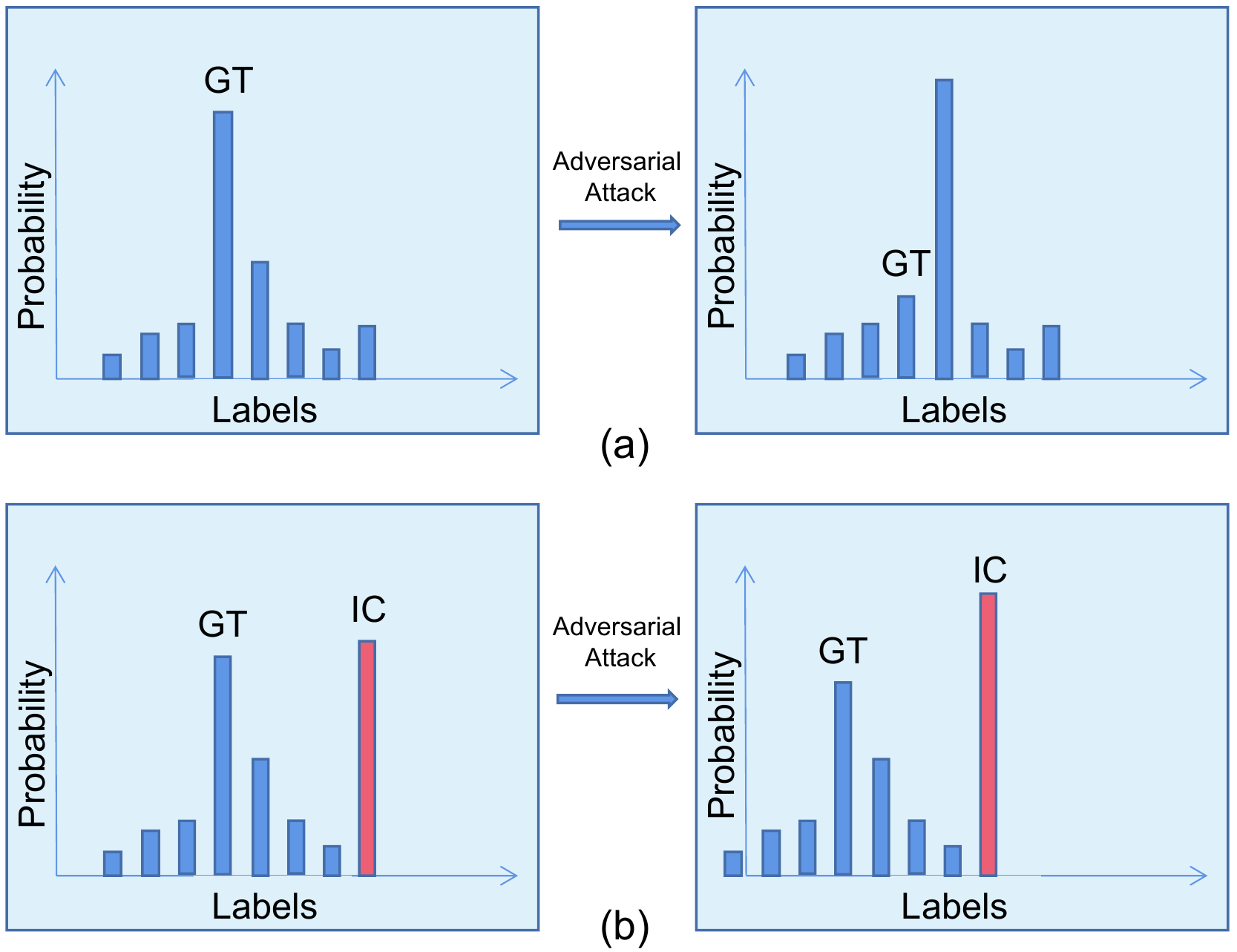}
    \caption{An illustration of the alternation of the predicted distribution before and after an adversarial attack. (a) Traditional DNNs; and (b) The proposed induction method. Here, ``GT'' stands for ``Ground Truth.''} 
    \label{icat}
\end{figure}

According to~\cite{DBLP:conf/ijcnn/XuWP21}, the principal consequence of adversarial attacks is the modification of the prediction distribution. On the basis of this, they suggest Induced Class Adversarial Training (ICAT), a simple but successful strategy that incorporates an extra-induced class to defend against adversarial examples. Fig.~\ref{icat} illustrates the alteration of predicted distributions before and after an adversarial attack. The method demonstrated better defense against white-box attacks compared to black-box attacks.

\begin{figure}
    \centering
    \includegraphics[height=1.5in]{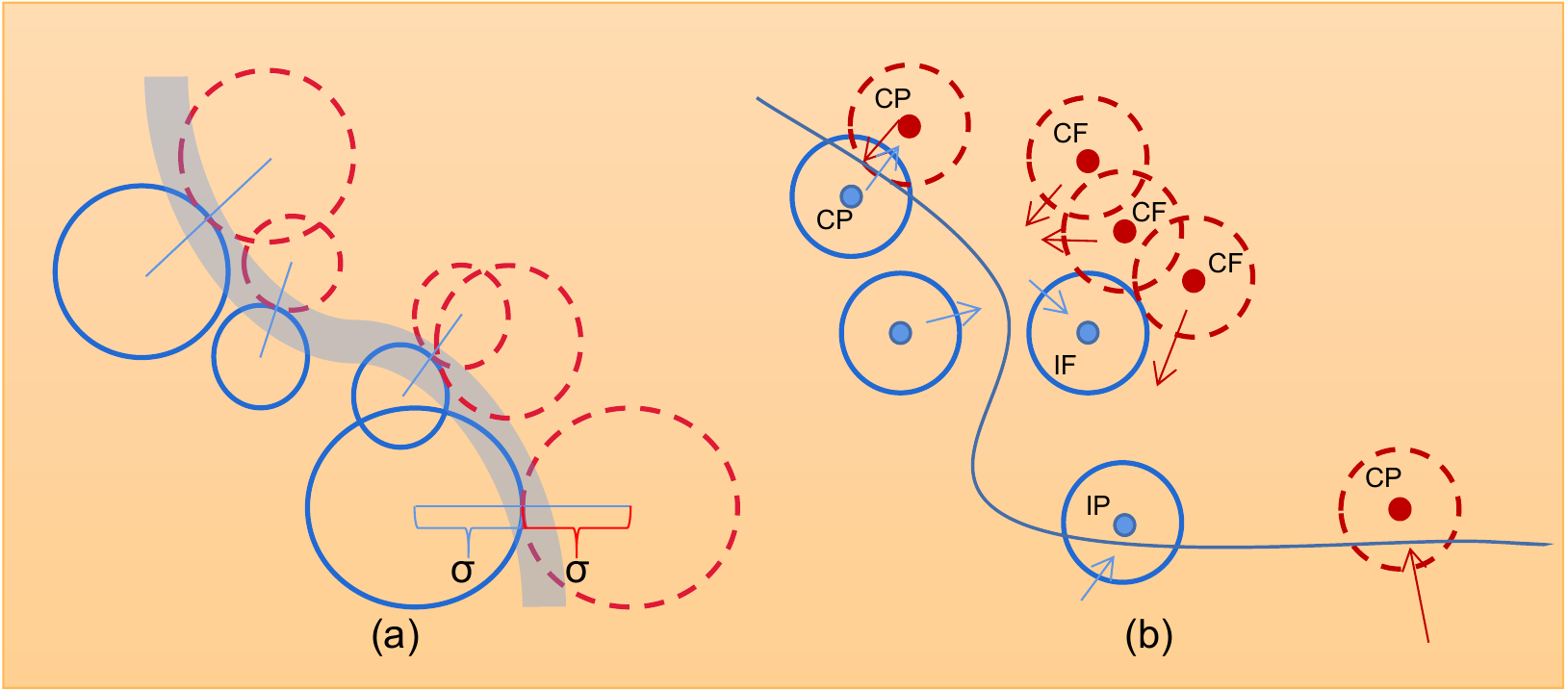}
    \caption{The min-max game's and ART's concepts: Different classifications of data, denoted by color and circle style, compete for a larger influence area in the middle of the circles. (a) The gray area where the influence regions overlap should be the location of the optimum robust decision border. (b) The current decision boundary is shown by the solid line.} 
    \label{art}
\end{figure}

Yao \textit{et al.}~\cite{DBLP:conf/ijcnn/YaoHHP21} target at classification and proposed Adaptive Retraining (ART) for neural networks, which implicitly improves a model's capacity in maximizing the minimal distance from data instances of all classes to the decision border. ART additionally builds a feedback loop and steers the data-generating process with categorization results for data augmentation. Fig.~\ref{art}  shows the concept of the min-max game and ART. ART has a negligible negative impact on classification accuracy, reducing the computational resources for data augmentation and time increment for model retraining, making it suitable for the online optimization of neural networks. However, experiments have only been conducted on small datasets, lacking the verification of defensive performance on large-scale datasets under strong attacks.

Gong \textit{et al.}~\cite{DBLP:conf/cvpr/GongRY021} present MaxUp, a simple yet effective method for enhancing generalization and minimizing overfitting. The objective of the technique is to construct a collection of enhanced data with randomly generated perturbations or modifications, and then to minimize the greatest loss across the improved data. To improve generalization performance, the method implicitly incorporates a smoothness or robustness regularization against random disturbances. While an additional forward pass is all that is required, MaxUp still has non-negligible additional time costs that can be lowered by employing low-resolution picture acceleration during selection.

Tripartite Adversarial Training for Network Embeddings (TriATNE) is an adversarial learning system designed by Liu \textit{et al.}~\cite{2021TriATNE}, which learns stable and durable node embeddings with three players. A basic producer collects features for node pairings, a dynamic seller selects negative samples, and a biased consumer perturbs the objective function, all at the same time measuring the performance of node embeddings. Producing and selling are in competition for consumers. TriATNE is founded on the idea that a resilient approach must be able to endure interruptions and assaults. The TriATNE framework outperforms baselines on link prediction across all datasets and performs well on homogeneous networks, despite limitations in heterogeneous network learning.

Poursaeed \textit{et al.}~\cite{DBLP:conf/iccv/PoursaeedJYBL21} propose Generative Adversarial Training (GAT) to boost the model's generalization ability to test sets and out-of-domain data, and its resilience against unanticipated adversarial attacks. GAT utilizes generative models with a disentangled latent space to produce a variety of low-, mid-, and high-level adjustments as opposed to modifying a single image feature. In addition to improving the model's performance on clean images and out-of-domain data, adversarial training with these cases makes it more robust to unforeseen attacks. The method is applicable to several applications, including classification, semantic segmentation, and object identification.


Hong  \textit{et al.}~\cite{DBLP:conf/cvpr/HongCK21} claim that properly blending the content and style of two input images can result in more numerous and robust samples, which enhances model generalization during training. Based on this concept, they present StyleMix, a new data augmentation approach that provides a variety of training samples via convex combinations of content and style attributes. They further expand this technology to StyleCutMix, which enables sub-image level modification through CutMix's cut-and-paste methodology~\cite{DBLP:conf/iccv/YunHCOYC19}. They also devise a method for automatically determining the degree of style mixing based on the class distance between two images. Experiments show that their strategies increase classifier robustness against adversarial attacks more than other recent mixup methods and improve model training generalization.

\begin{figure}
    \centering
    \includegraphics[height=1.9in]{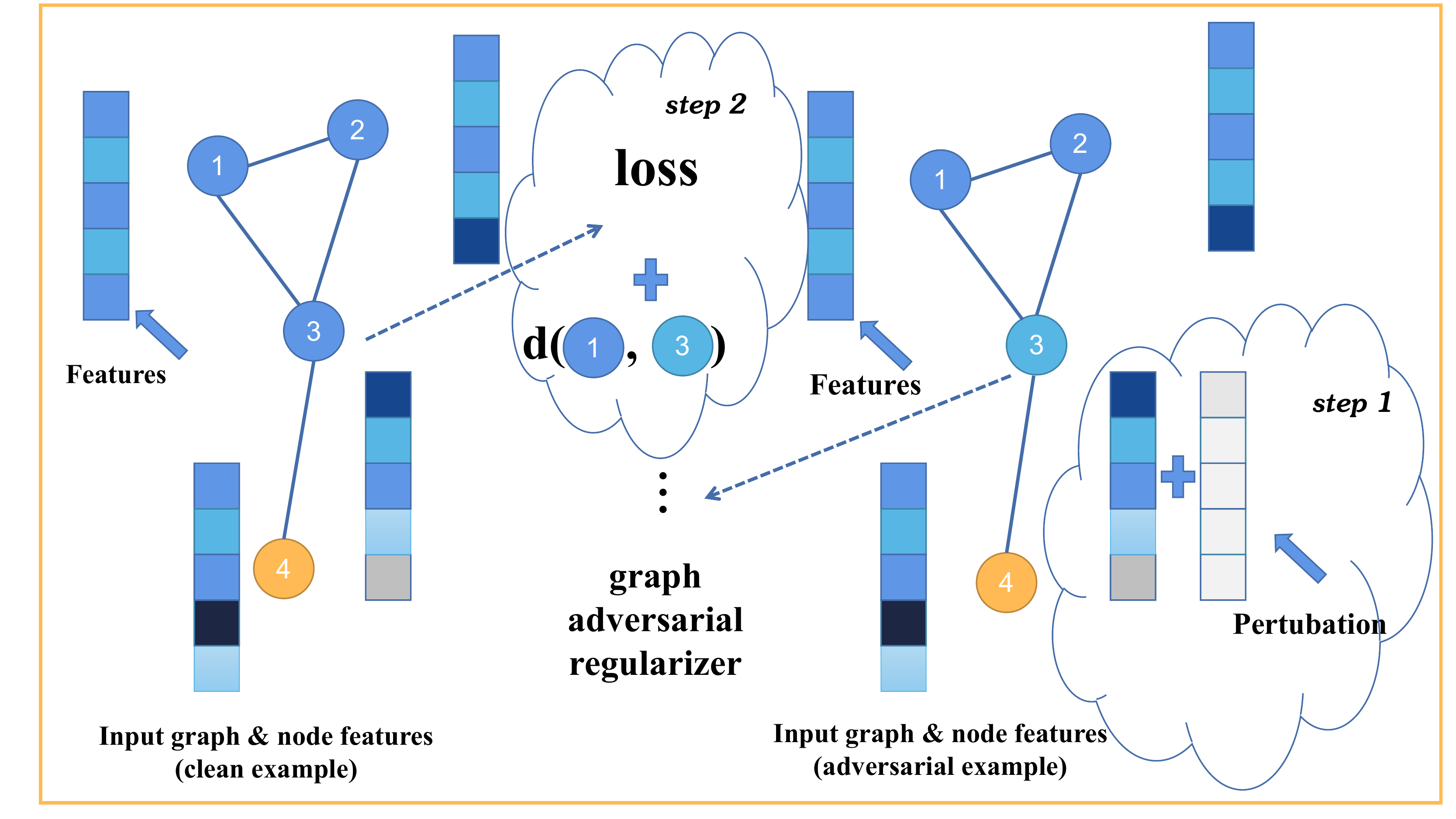}
    \caption{The training process of GraphAT: step 1) Producing graph adversarial instance and step 2) Optimizing model parameters through minimizing the loss and graph adversarial regularizer.} 
    \label{graphAT}
\end{figure}

Feng \textit{et al.}~\cite{DBLP:journals/tkde/FengHTC21} propose a dynamic regularization scheme called graph adversarial training (GraphAT), as shown in Fig.~\ref{graphAT}. The scheme breaks the smoothness of linked nodes to the greatest degree possible and generates network adversarial examples by perturbing the input of associated clean examples. Furthermore, it minimizes the graph neural network's objective function using an extra regularizer across adversarial graph samples. This promotes smoothness between adversarial and linked example predictions, making the model more robust to perturbations transmitted across the graph. However, the computing cost grows linearly with the neighbors sampled. In addition, the work focuses only on graph-based learning from a single graph, but future research objectives seek to determine the efficacy of graph-adversarial training on numerous graphs.

\begin{figure*}
    \centering
    \includegraphics[height=1.8in]{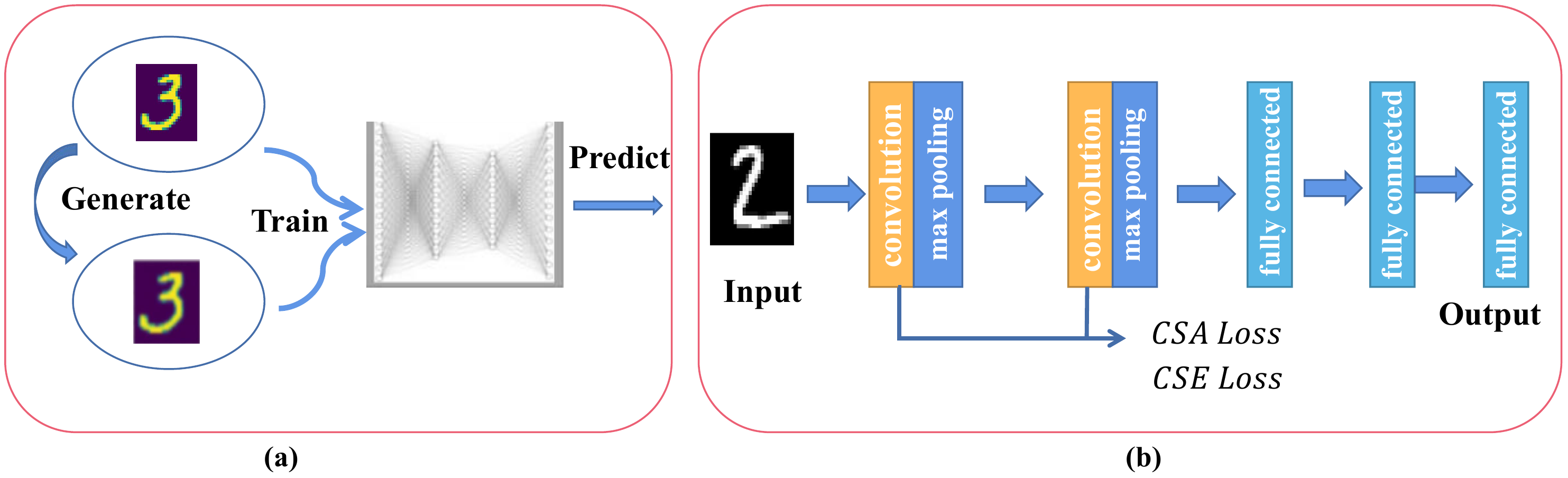}
    \caption{Comparison of conventional adversarial learning and CSA/CSE algorithms. (a) In adversarial learning, clean and adversarial examples are alternately supplied into the neural network. (b) To accomplish the min-max property, the framework for CSA and CSE algorithms applies the convolution parameter to the loss function by adding a normalization.} 
    \label{csa}
\end{figure*}

Shen \textit{et al.}~\cite{9712207} suggest a novel method for safeguarding a particular class from adversarial attacks, which is different from previous defensive approaches that try to increase the resilience of overall classes. They adopt cost-sensitive adversarial learning (CSA) and cost-sensitive adversarial extension (CSE) to include cost sensitivity in adversarial learning and enhance the model's adversarial robustness. Fig.~\ref{csa} depicts an overview of the adversarial learning system, as well as the CSA and CSE algorithms. The techniques have been tested on the MNIST and CIFAR datasets. However, the gain in robustness for more complicated datasets may be limited.

In view of human outstanding generalization ability, Chen \textit{et al.}~\cite{DBLP:conf/iccv/ChenPM0D021} argue that a resilient CNN should be able to endure changes in amplitude while concentrating on the phase spectrum. In order to do this, they provide a unique data enhancement approach dubbed Amplitude-Phase Recombination. APR integrates the phase spectrum of the current picture with the amplitude spectrum of an adversarial image to generate a new training sample with the same label as the current sample. This strategy allows the CNN to get more structured data from the phase components than from the amplitude components. APR is at the forefront of several generalization and calibration problems, such as adaption for surface fluctuations and common corruptions, adversarial assaults, and out-of-distribution detection.

\subsubsubsection{Feature Extraction}
To train a model, feature extraction focuses on the texture features of images. Sun \textit{et al.}~\cite{DBLP:conf/iccv/SunLXQKL021} are particularly interested in shape features and propose two edge-enabled pipelines, namely, EdgeNetRob and EdgeGANRob, to force CNNs to rely more on edge features, inspired by the fact that the visual system of humans ends up paying more attention on global features, such as shapes, for recognition, whereas CNN models are biased towards local features (e.g., textures) in images. EdgeNetRob and EdgeGANRob both use an edge detection technique to extract structural attributes from a picture. EdgeNetRob then trains downstream learning tasks by using the recovered edge features, while EdgeGANRob rebuilds a new image by filling in texture features using a learned GAN. These findings demonstrate adding edge features can increase the model's robustness while decreasing clean accuracy significantly.

{\small
\setlength{\tabcolsep}{2pt}
\begin{table*}[!t]
\caption{Regularization on Input Layer (\uppercase\expandafter{\romannumeral1})}
	\label{table7}
	\setlength{\tabcolsep}{3pt} 
	\renewcommand\arraystretch{1.5} 
	\begin{tabular}{|p{1cm}|p{1cm}|p{1.3cm}|p{3.3cm}|p{1cm}|p{1cm}|p{3.7cm}|p{3.5cm}|}
    \hline
    \centering{\textbf{Defense}} & \centering{\textbf{Strategy}} & \centering{\textbf{Attack type}} & \centering{\textbf{Description}} & \centering{\textbf{Input}} & \centering{\textbf{Inv-Metric}} & \centering{\textbf{Strength}} & \centering{\textbf{Weakness}}  \cr \hline
   AMM \cite{DBLP:journals/pami/YanGZ21}  & Add Noise & White-box, Black-box & A regularization method based on learning that uses an adversarial perturbation as a proxy. & Image & $\ell_p$ & Significantly improve the test set accuracy of various DNN architectures, and the generalization capability has been enhanced &Higher computational cost. \\ \hline
        
    GAN-based Deepfake \cite{9533868}& Add Noise & White-box, Black-box, Gray-box &  An adversarial face generating approach that incorporates random differentiable image alterations during DeepFake model training to safeguard people's faces. & Image & $\ell_1$ & Makes it easy to recognize induced fake images and videos, regardless of model or data, and uses the same technique in all scenarios. & DeepFake model training takes a significant amount of time. \\\hline

    PDA \cite{DBLP:journals/tip/YuLLYZ21}  & Add Noise & Black-box, Corruption& During training, gradually introduce various adversarial perturbations. & Image & $\ell_2$ & Spending less training time while maintaining a high level of accuracy on clean samples, regulating the perturbation boundaries, ensuring greater robustness. & It is difficult to achieve robustness against white-box adversarial attacks that are confined to multiple spaces. \\ \hline
    
   adMRL \cite{DBLP:conf/ijcnn/ChenCW21}   &Add Noise  &White-box & A novel MRL algorithm learning strategy for generalizing a meta-policy by meta-training an agent in a distorted environment with disturbed states.& Robotic trajectories & $\ell_p$ & Based on model-agnostic meta-learning, the agent may learn the initial parameters with improved generalization ability, as well as fight against additional ``bad'' samples. & Only test on FSGM and random noise attack.\\ \hline
    
HIRE-SNN \cite{2021HIRE}  & Add Noise& White-box, Black-box & Use the time steps of SNN training to efficiently input multiple noisy variations of the same image.& Image & $\ell_p$ &The robustness is not primarily derived from gradient masking, and the degradation in clean image accuracy is negligible. Improve inference latency and computation energy. &Improve model robustness at the expense of clean-image classification accuracy.\\ \hline

   ICAT \cite{DBLP:conf/ijcnn/XuWP21} & AT &White-box, Black-box  &Adversarial training with one additional induced class. & Image  & KL, CE  & Propose a new idea that the main impact of counteracting attacks is the alternation of prediction distributions. & Worse defense against black-box attacks over white-box attacks. \\ \hline
    
   TriATNE \cite{2021TriATNE}  & AT & White-box, Black-box &A novel framework for learning stable and robust node embeddings with three participants, where the producer and the seller compete to win customers. & Node & $\ell_p$ & TriATNE outperforms the baseline on all datasets in link prediction, showing good performance on homogeneous networks. & Learning using, e.g., knowledge graphs, continues to pose difficulties. \\ \hline
    
Graph-AT \cite{DBLP:journals/tkde/FengHTC21}   &AT  & White-box &Dynamic regularization according to the graph structure. & Node &  CE & Dynamic regularization; When learning to construct and resist perturbations, the influence of the connection instance is considered. & The computational overhead linearly rises as sampling more neighbors. Focus only on graph-based learning from one graph.\\\hline
    
  CSA\& CSE \cite{9712207}   & AL  & White-box & CSA is a cost-sensitive adversarial learning approach. CSE is an end-to-end learning strategy that can improve the model's adversarial robustness with no need for AT. & Image & $\ell_2$ & Cost-sensitive training can successfully defend specific categories against adversarial attacks and can be used in conjunction with AT to increase performance. & May not be successful in improving the adversarial robustness of the model in a more complex dataset.\\ \hline
    
   GAT \cite{DBLP:conf/iccv/PoursaeedJYBL21}  & AT&Unseen Attack, OOD &An strategy for improving the model's generalization to test data and OOD samples while also improving its robustness against unknown adversarial attacks. & Image &  $\ell_p$ & Achieve SOTA robustness without training, improve both robustness and generalization. Extend to classification, semantic segmentation and object detection. &  
  Maybe high computation cost. \\ \hline
          \end{tabular}
\end{table*}

\begin{table*}
\caption{Regularization on Input Layer (\uppercase\expandafter{\romannumeral2)}}
	\label{table7-2}
	\setlength{\tabcolsep}{3pt} 
	\renewcommand\arraystretch{1.5} 
	\begin{tabular}{|p{1cm}|p{1cm}|p{1.3cm}|p{3.4cm}|p{1cm}|p{1.2cm}|p{3.5cm}|p{3.5cm}|}
    \hline
    \centering{\textbf{Defense}} & \centering{\textbf{Strategy}} & \centering{\textbf{Attack type}} & \centering{\textbf{Introduction}} & \centering{\textbf{Input}} & \centering{\textbf{Invisibility Metric}} & \centering{\textbf{Strength}} & \centering{\textbf{Weakness}}  \cr \hline
    APR \cite{DBLP:conf/iccv/ChenPM0D021}  &AL&White-box, Black-box, Corruption, OOD & Recombine the phase spectrum of the current image with the distracter image amplitude spectrum to create a new training sample with the current image as the label. & Image & CE & Good adaptability to common corruption and surface changes, out-of-distribution detection, while maintaining good ability on clean images. & Need to investigate ways to represent part-whole hierarchies in neural networks based on the phase spectrum as well as more CNN models or convolution operations. \\ \hline
          
    ART \cite{DBLP:conf/ijcnn/YaoHHP21} & AL & Gaussian Noise & A neural network retraining strategy that indirectly improves a model's ability to maximize the least distance from all data examples to the decision. & Image and data points & $\ell_{\infty}$ & The negative impact on classification accuracy is small. Reduce compute resources spent on strengthening robust data samples and increment of model retraining time. & Experiments on small datasets only, which lack defensive performance experiments under strong attacks.\\ \hline
            
   Style-Mix \cite{DBLP:conf/cvpr/HongCK21}  & AL & White-box & A new data augmentation mix-up method that can generate different training samples by convex combinations of content and style characteristics. & Image & $\ell_2$ &  Performs better or similar to SOTA mix-up approaches and learns more robust classifiers against adversarial attacks. Enhance the generalization of model training. & The automatic selection of the mixed ratio slightly degrades the defensive performance of further separate foreground from background.
    \\ \hline
    
  MaxUp \cite{DBLP:conf/cvpr/GongRY021} & AL & Gaussian Noise & Introduce gradient-norm regularization for improving the loss function's smoothness in order to increase generalization performance and reduce over-fitting. & Image, video, 3D point cloud & $\ell_2$ & MaxUp can easily take any SOTA data enhancement scheme and significantly improve them by minimizing the worst-case (rather than average) risk on enhanced data. & Although it merely necessitates an additional forward pass, MaxUp incurs a non-negligible additional time cost, which may be reduced by using low-resolution images. \\\hline
    
EdgeNet-Rob \& EdgeGANRob \cite{DBLP:conf/iccv/SunLXQKL021}  & Feature Extract & White-box, Black-box, Corruption
& Two edge-enabled pipelines EdgeNet-Rob and EdgeGAN-Rob, make the CNNs rely more on edge features.& Image & CE & Edge features can improve the CNN model's robustness. Clean accuracy can be increased on datasets with clear edge information by repopulating the texture information. & Clean accuracy is slightly reduced. \\ \hline
     \end{tabular}
\end{table*}
}

\subsubsection{Regularization on Middle Layer}

Intermediate layer regularization can be achieved by operating on neurons, hidden layers, and Lipschitz condition constraints. Tables~\ref{middle1} and~\ref{middle2} briefly introduce the latest regularization methods on the middle layer and compare their advantages and disadvantages. 

\subsubsubsection{Layer Regularization}
On the middle layer, the most commonly adopted regularization method is layer regularization, where a regularization term is added to the hidden layer.

Adversarial Noise Propagation (ANP)~\cite{DBLP:journals/tip/LiuLYZLT21} is a simple yet strong training technique that, unlike classic adversarial defensive methods, does not manipulate solely the input layer (as discussed in Section~\ref{input layer}). During training, it injects noises into the hidden layers by propagating backward from the adversarial loss. This enables the learned parameters in each layer to produce accurate and consistent results for the benign instance and its distributed noisy surrogates, resulting in a high degree of resilience. Due to the fact that each layer helps improve the resilience of the model to differing degrees in this technique, it is necessary to build a more adaptable algorithm that takes into consideration the changing behavior of different levels. While perturbing the deep layers boosts the model's robustness, adding noises to the shallow layers has an adverse effect. Nevertheless, the root source of this issue is not studied in~\cite{DBLP:journals/tip/LiuLYZLT21}.

Zhang \textit{et al.}~\cite{DBLP:journals/tip/ZhangLLXYML21} provide a unique perspective of neuron sensitivity to explain adversarial resilience for deep models, as assessed by the magnitude of variance in neuronal activity in response to benign and adversarial situations. They suggest a Sensitive Neuron Stabilizing (SNS) approach after analyzing the behaviors of the model's intermediate layers and demonstrating dependence between adversarial resilience and neuron sensitivities. By stabilizing sensitive neurons (instead of obfuscation), the technique tries to increase the model's robustness to adversarial samples. However, dynamically updating sensitive neurons incurs a greater computational expense.

Yao  \textit{et al.}~\cite{DBLP:conf/ijcnn/YaoG21} offer a defensive model that makes use of a supervision method to enhance the model's robustness. The assumption is that supervision can increase the quality of feature maps for the hidden layer, hence increasing the resilience of the model. As a secondary classifier, a supervisory layer is appended to the main neural network model to perform this strategy. By utilizing the classification results of intermediate layers to assess perturbation properties and by continuously changing the loss function of the supervisory layer, the quality of the hidden layer's recovered features may be enhanced. This decreases the impact of adversarial perturbation and boosts the model's overall resilience. Under both black-box and gray-box threat models, the suggested technique can survive assaults from FGSM as well as C\&W, BIM, and DeepFool attack algorithms to a large degree. However, under a white-box threat model, it will only reduce the confidence of the attacker and cannot effectively defend against attacks, being considered a weakly supervised defense model.

Schwartz \textit{et al.}~\cite{DBLP:conf/ijcnn/SchwartzD21} propose a method called HLDR to improve adversarial robustness by using latent disparity regularization. The regularizer is defined to depend linearly on the disparity in representations created in the hidden layers based on benign and adversarial data. The regularizer penalizes the discrepancy and provides significant improvements in adversarial robustness while also reducing training time. However, the method only considers training programs with fine-tuned subsets constructed using one method. It is not yet understood how HLDR performance may be affected by adversarial, Gaussian or other forms of distortion, different training methods for fine-tuning subsets, and different objective functions.

\begin{figure}
    \centering
    \includegraphics[height=0.75in]{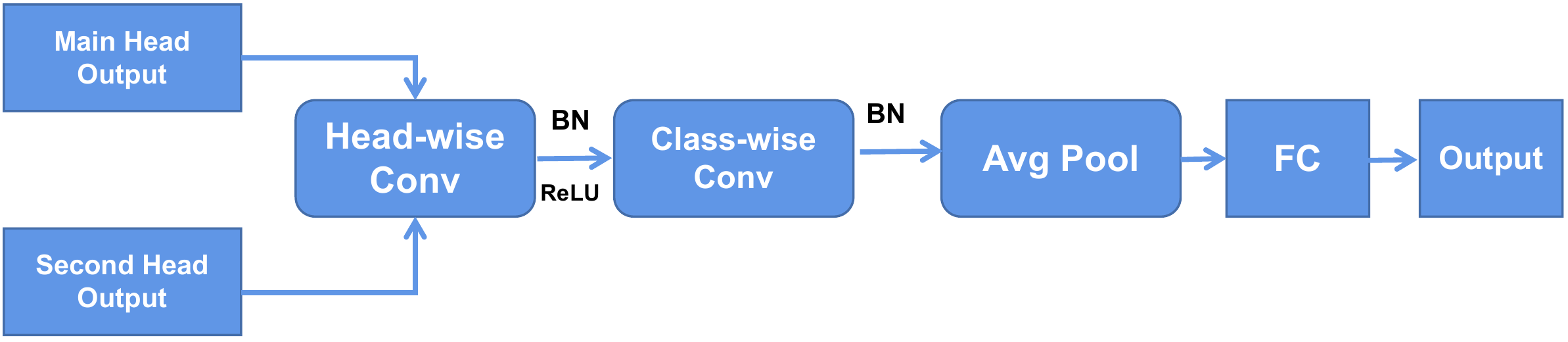}
    \caption{The workflow of DH-AT. DH-AT connects another head to an intermediate layer of the neural network, and utilizes a shallow CNN to integrate the outputs of the two heads.} 
    \label{dhat}
\end{figure}

Dual Head Adversarial Training (DH-AT)~\cite{DBLP:conf/ijcnn/JiangMEB21} is a novel defensive method that employs a dual-headed architecture to increase both clean accuracy and adversarial resilience as an enhanced form of Adversarial Training (AT) in both network structure and training strategy. The architecture of the lightweight CNN used by DH-AT is outlined in Fig.~\ref{dhat}. As seen in the diagram, DH-AT connects a second network head (or branch) to a network's intermediate layer before combining the outputs of both heads using a lightweight CNN. In order to attain both clean precision and resilience in the meantime, the training technique is modified to account for the relative significance of the two heads. A potential drawback is that DH-AT may require longer training time.

Li \textit{et al.}~\cite{DBLP:conf/iccv/LiMLYKWH21} propose an Embedding Regularized Classifier (ER-Classifier) to improve the adversarial resilience of classifiers. The intrinsic dimension of image data is substantially less than its pixel space dimension, and hostile examples often dwell outside the manifold of natural image data. The approach projects high-dimensional input images into a low-dimensional space and returns adversarial samples to the manifold of natural image data via regularization. This enhances classification accuracy when faced with adversarial scenarios. In addition, the framework may be utilized in conjunction with detection approaches to discover adversarial instances. Exploration of low-dimensional areas to improve the resiliency of DNNs is a potential innovation.

Similarly, Carbone \textit{et al.}~\cite{DBLP:conf/ijcnn/CarboneSB21} propose RP-Ensemble, a training approach based on the Manifold Hypothesis~\cite{2012Manifold, DBLP:conf/stoc/DasguptaF08}. Using projected representations of the original inputs, this method enhances the robustness of a pre-trained classifier against adversarial examples. Moreover, they develop the RP-Regularizer, a regularization term based on the norms of the loss gradients, which measures vulnerability, with the expectation over random projections of the inputs. This is done during training to capitalize on pertinent adversarial characteristics. The strategy is computationally efficient and independent of the attack norm type. However, the CIFAR-10 dataset produces less impressive results than the MNIST dataset.

CNNs usually ignore key auxiliary properties, whilst current adversarial training and regularization approaches overlook the independence of local features. Liu \textit{et al.}~\cite{DBLP:conf/iccv/LiuW00S21} introduce TENET Training, a group-wise inhibition-based regularization approach for enhancing feature diversity and network resilience. The suggested approach dynamically regularizes CNNs during learning by suppressing regions with the highest activation values that are the most discriminative. This allows the network to study more diverse aspects, which can more accurately depict pictures, even when they have been altered maliciously. TENET Training improves both robustness and generalization significantly in comparison to other SOTA methods.

While classifying hyperspectral pictures, a novel self-attention context network (SACNet)~\cite{DBLP:journals/tip/XuDZ21} is presented to strengthen the network's inherent resilience to adversarial samples. Existing adversarial learning algorithms are designed to recognize RGB pictures, but this strategy targets hyperspectral images (HSIs). In contrast to local feature extraction, global context information extraction necessitates the construction of associations between a specific pixel and all relevant pixels in the whole picture. This pixel's network prediction would be impacted by its neighboring pixels, and the overall loss would be dispersed across all neighboring pixels. Therefore, tackling these networks may need a greater degree of disturbance. The suggested SACNet has a very high temporal cost since self-attention learning and contextual encoding raise the computing burden of the overall framework. Future research should also examine if SACNet can defend against adaptive RGB images.

According to~\cite{DBLP:journals/tip/LiSGL21}, the downsampling method is primarily responsible for CNNs' poor noise resilience. They combine frequently employed CNN designs with a discrete wavelet transform (DWT) to produce wavelet-integrated convolutional networks (WaveCNets), which address the issue of aliasing in CNNs and enhance noise resistance in image classification. During downsampling, DWT separates the WaveCNets feature maps into low-frequency and high-frequency components. Low-frequency components are passed to succeeding layers to retain robust high-level characteristics, while high-frequency components are deleted to avoid noise transmission. Although WaveCNets regularly resists different forms of noise, its performance is inferior to that of well-trained defensive systems. The wavelet transform also adds computational complexity.

Yu \textit{et al.}~\cite{DBLP:conf/iccv/YuCXLWBM21} show that universal adversarial patches in prevalent CNNs often generate deep feature vectors that have very large norms. They suggest a simple but effective defensive technique based on a unique feature norm clipping (FNC) layer. This differentiable module can be dynamically introduced to various CNNs to prevent the adaptive creation of huge norm-deep feature vectors. An FNC investigation of the effective receptive field (ERF) reveals that the adversarial patch's impacts may be minimized naturally, resulting in enhanced classification accuracy against adversarial patch attacks. Experiments conducted on multiple datasets demonstrate that this proposed technique enhances the robustness of various CNNs against white-box universal patch assaults while retaining acceptable identification accuracy for clean samples and incurring a little computational cost. However, it remains unclear why this method is still valid for position-independent image patches.

Mok \textit{et al.}~\cite{DBLP:conf/iccv/MokNCY21} investigate the topic of constructing an adversarially resilient neural network with strong inherent resilience and robust training strategies. Adversarially Robust Architecture Rush (AdvRush) is an adversarial robustness-aware neural architecture search (NAS) approach based on the observation that the inherent robustness of a neural network depends on the smoothness of its input landscape regardless of the training procedure. Using a regularizer that prefers candidate architectures with a smoother input loss landscape, AdvRush selects a neural network that is resistant to adversarial inputs. The approach is very adaptable to many datasets. Future studies will examine the robustness of neural network architecture on multimodal datasets and broaden the search area to include activation functions.

Yeo \textit{et al.}~\cite{9711018} present a general framework for developing robust predictions based on the creation of a varied ensemble of different middle domains. The suggested method makes predictions by combining a variety of cues (called ``middle domains") into a single strong prediction. The concept is that predictions based on distinct cues react differently to a distribution adjustment. As a result, they can be combined into a robust final prediction. The method can change without manual modification or redesign, or any additional supervision or labeling already attached to the dataset. It can generalize to new non-adversarial and anti-corruption scenarios. On the other hand, the limitations of the method include its reliance on reasonable uncertainty estimates in the presence of distribution shifts and its use of only unimodal distributions for the study. Additionally, the selection of middle domains and the use of ensemble-based methods inevitably increase computational complexity.

\subsubsubsection{Lipschiz Condition}
To mitigate the sensitivity of the output to changes in the input, Amini \textit{et al.}~\cite{DBLP:journals/tmm/AminiG20} offer a new non-smooth regularization term in the optimization formulation and two non-smooth regularizers that construct direct linkages for the weight matrices in each neural network layer. These regularizers adjust the Lipschitz constant of the underlying architecture to make the mapping between inputs and outputs more stable, hence reducing the network's sensitivity to small perturbations. However, regular gradient-based learning methods become less useful when non-smooth variables are included.

It has been demonstrated in~\cite{DBLP:conf/nips/HeinA17} that limiting a neural network's Lipschitz constant gives certifiable robustness guarantees against local adversarial attacks and increases the model's generalization ability and interpretability. Therefore, Serrurier \textit{et al.}~\cite{DBLP:conf/cvpr/SerrurierMGBLB21} provide a novel optimal transport-based classification framework that takes into account the Lipschitz constant and the gradient norm preservation requirement. They use a regularized Kantorovich-Rubinstein formulation that includes a hinge loss term, which provides the desired robustness guarantees with little accuracy loss. One possible drawback is that the computation time increases during learning.

Some defensive approaches augment the standard training aim for intermediate layers with graduated penalties. For example, by using a layer-by-layer contrast penalty term, Gu and Rigazio~\cite{DBLP:journals/corr/GuR14} are able to retain the output of a DNN unaffected by input disturbances. FNC~\cite{DBLP:conf/iccv/YuCXLWBM21} is also a novel feature norm shear layer that can be placed flexibly into various networks to adaptively suppress the creation of big norm depth eigenvectors and enhance its overall performance.

{\small
\setlength{\tabcolsep}{2pt}
\begin{table*}
\caption{Robustness Regularization on Middle Layer (\uppercase\expandafter{\romannumeral1})}
	\label{middle1}
	\setlength{\tabcolsep}{3pt} 
	\renewcommand\arraystretch{1.5} 
	\begin{tabular}{|p{1cm}|p{1cm}|p{1.3cm}|p{3.7cm}|p{1cm}|p{1.2cm}|p{3.5cm}|p{3.5cm}|}
    \hline
    \centering{\textbf{Defense}} & \centering{\textbf{Strategy}} & \centering{\textbf{Attack type}} & \centering{\textbf{Description}} & \centering{\textbf{Input}} & \centering{\textbf{Invisibility Metric}} & \centering{\textbf{Strength}} & \centering{\textbf{Weakness}}  \cr \hline    
    
    KR \cite{DBLP:conf/cvpr/SerrurierMGBLB21}  & Lipschitz  &White-box &A new optimal transport-based classification framework that incorporates the Lipschitz constant and the gradient norm preservation constraint as a theoretical need.& Image & hinge-KR & The expected guarantee in terms of robustness is supplied without any major loss of accuracy. Use adversarial attacks to interpret a prediction. & The calculation time increases during learning.\\ \hline
 
  Inf-Norm\& Inf-Ind \cite{DBLP:journals/tmm/AminiG20} & Lipschitz  & White-box & Two novel robust training approaches for CNN architectures are proposed. Both techniques require modifying the network structure using an approximate non-smooth regularization term that influences the spectral constants in the convoluted and fully connected layers. & Image & $\ell_2$, $\ell_{\infty}$ & Improve the architecture's robustness by making network mapping more reliable and interpretable. Gradually increasing complexity through warm-starting. & The use of non-smooth terms restricts the application of traditional gradient-based learning techniques.\\ \hline
    
    ANP \cite{DBLP:journals/tip/LiuLYZLT21}  & Layer Reg. &White-box, Black-box & A powerful training algorithm that adds noises to the hidden neural network layers in a layer-wise manner.& Image & RMS Distance & Easily integrated with adversarial training methods, and efficiently perform using the basic backward-forward approach, introducing no additional computations and memory consumption. & Since each layer contributes to the robustness of the model to varying degrees, a more adaptive algorithm needs to be designed to take into account the heterogeneous behavior of the different layers. 
    \\ \hline

    SNS \cite{DBLP:journals/tip/ZhangLLXYML21} & Layer Reg. &White-box, Black-box, Gaussian Noise &Stabilize neuron sensitivities toward benign and adversarial examples. & Image & $\ell_1$ & SNS performs well against black-box attacks on different datasets, such as SPSA and NATTack, but does not achieve security through obfuscation.
    & Updating sensitive neurons dynamically incurs higher computational costs. \\ \hline

  HLDR \cite{DBLP:conf/ijcnn/SchwartzD21}  & Layer Reg. &White-box &A training technique and goal function for arbitrary neural network designs that are adversarially robust. & Image & $\ell_1$ & More effective than other advanced techniques to protect neural networks from adversarial evasion attacks; only requires as many parameters as the original model; training fairly quickly. &  
  The effect of distortion, different training methods, and attack strategies on HLDR performance is unclear. \\ \hline
    
    ER-Classifier \cite{DBLP:conf/iccv/LiMLYKWH21} & Layer Reg. &White-box, Black-box &A new end-to-end robust DNN defensive scheme that enhances the classifier's adversarial robustness by embedding regularization.& Image & WD &  In addition to improving the classification accuracy of adversarial samples, the framework can also be combined with assays to detect adversarial samples. & Further research on low-dimensional space needs to be done to increase the robustness of DNNs.\\ \hline
    
    RP \cite{DBLP:conf/ijcnn/CarboneSB21}  & Reg. &White-box &Propose RP-Ensemble, a training technique consisting of projected versions of the original inputs and RP-Regularizer, a regularization term to the training objective. & Image & $\ell_1$ & Totally independent of the attack's selected norm, and computationally efficient. & On CIFAR-10, the results were less significant than those for MNIST. \\ \hline

    DH-AT \cite{DBLP:conf/ijcnn/JiangMEB21} & Reg. &White-box &An upgraded variation of AT that connects a second head to one of the network's intermediate layers. & Image & CE, KL & Be readily added with minor changes into existing adversarial training techniques, and can achieve clean precision and robustness at the same time. & Incurs more training time. \\ \hline
     \end{tabular}
\end{table*}
    
\begin{table*}[!t]
\caption{Robustness Regularization on Middle Layer (\uppercase\expandafter{\romannumeral2})}
	\label{middle2}
	\setlength{\tabcolsep}{3pt} 
	\renewcommand\arraystretch{1.5} 
	\begin{tabular}{|p{0.85cm}|p{1cm}|p{1.3cm}|p{3.8cm}|p{1cm}|p{1.2cm}|p{3.5cm}|p{3.5cm}|}
    \hline
    \centering{\textbf{Defense}} & \centering{\textbf{Strategy}} & \centering{\textbf{Attack type}} & \centering{\textbf{Description}} & \centering{\textbf{Input}} & \centering{\textbf{Invisibility Metric}} & \centering{\textbf{Strength}} & \centering{\textbf{Weakness}}  \cr \hline    
   TENET \cite{DBLP:conf/iccv/LiuW00S21} &Layer Reg. &White-box &A regularization strategy based on group-wise inhibition for increasing feature diversity and network robustness. & Image & CE & 
   Enables the network to explore varied and richer features, resulting in a more accurate picture representation, even when malicious alterations are introduced.
   & Maybe high computation cost. 
   \\ \hline
    

  SACNet \cite{DBLP:journals/tip/XuDZ21}  & Layer Reg. & White-box& A new self-attention context network in which all of the loss at one pixel is shared by all of the pixels that are connected to it. Attacking such a network requires a higher level of perturbation. & Hyper-spectral Image & Context enhanced features & The HSI's global contextual information may considerably enhance the resilience of DNNs against adversarial attacks.& SACNet has a higher time cost since self-attention learning, and context encoding raises the computing overhead of the whole framework. \\ \hline
    
FNC \cite{DBLP:conf/iccv/YuCXLWBM21} &Layer Reg.& White-box& Feature norm clipping layers, which are differentiable modules that may be arbitrarily placed in different CNNs, are used to prevent the creation of excessively large norm deep feature vectors.&Image & $\ell_p$ & Does not introduce trainable parameters and has only very low computational overhead, effectively improving the robustness of white-box generic patch attacks by different CNNs.
    & The identification accuracy of the clean samples affects slightly. The reason why remaining valid for the location-independent patch for a single image is unknown.\\ \hline
    
  Wave-CNets \cite{DBLP:journals/tip/LiSGL21}  &Layer Reg. & White-box, Black-box& For noise-resistant image classification, DWT is applied to the feature maps during downsampling to minimize aliasing. & Image & Corruption Error &Separation of aliasing effects, or the differentiation between low- and high-frequency information, may be stymied without resorting to adversarial training. & It is less resistant to attackers than specially trained defense methods and the wavelet transform introduces additional computations.\\ \hline
    
   Super-vision Layer \cite{DBLP:conf/ijcnn/YaoG21}&Layer Reg. & Black-box, Gray-box, White-box& A defensive model in which a supervisory layer is introduced as an auxiliary classifier to the basic neural network model. & Image & $\ell_p$ & The quality of the features recovered by the hidden layer is enhanced by the black box and gray box threat models. 
   &  In the case of white-box defense, it will only reduce the confidence of the attacker and make them unable to protect against assaults effectively.
   \\ \hline
    
  Adv-Rush \cite{DBLP:conf/iccv/MokNCY21}  & Layer Reg. & White-box, Black-box& A unique NAS adversarial robustness-aware neural architecture search technique that employs a regularization term generated from the neural network's loss landscape curvature data.
  & Image & $\ell_2$ & 
  AdvRush successfully discovered a neural process for adversarial robustness, and it is
  highly transferable on different datasets. & For large regularization strength, the searched architecture experiences a significant drop in clean accuracy. \\ \hline 
    
    Cross-Domain Ensembles \cite{9711018} &Layer Reg. & White-box, Corruption& A general framework for making robust predictions based on creating a diverse ensemble of various middle domains. & Image & $\ell_1$ & No manual modification or redesign is required when making a change between middle domains. No additional supervision or labeling is required than what the dataset already comes with; Can be extended to a whole new non-adversary and anti-corruption. & Depends on reasonable uncertainty estimation in the presence of distribution transfer; Selection of intermediate domain; Using only unimodal distribution; Integration-based methods can increase computational complexity.\\ \hline  
     \end{tabular}
\end{table*}
}

\subsubsection{Regularization on Output Layer}
The output layer can be regularized through defense distillation, label smoothing, or by modifying loss functions. An overview of the possible regularization methods in the output layer is provided and compared in Table~\ref{outl}. 

\subsubsubsection{Distillation}
Many recent works enhance the resilience of a student network with a teacher network by using Knowledge distillation (KD)~\cite{DBLP:conf/aaai/LiLWZWL22} in combination with adversarial training. They build on adversarial training by using a teacher network that has already been pre-trained in the form of adversaries.
According to~\cite{DBLP:conf/iccv/ZiZMJ21}, adversarial training approaches are more successful on large models and less effective on small models. To solve this issue, Zi \textit{et al.}~\cite{DBLP:conf/iccv/ZiZMJ21} propose Robust Soft Label Adversarial Distillation (RSLAD), a unique method for training small, robust student models by distilling from adversarially trained large models. 
This strategy replaces hard labels with robust soft labels inside supervision loss terms.
It employs the huge, robust instructor model's robust soft labels to assist student learning in both natural and adversarial situations. However, one downside of the strategy is that when the teacher network grows too complicated for the student network to learn from, the student's robustness tends to decline.

Attention Guided Knowledge Distillation and Bi-directional Metric Learning (AGKD-BML) is a new adversarial training-based model proposed by Wang \textit{et al.}~\cite{DBLP:conf/iccv/WangDYLL21}. By leveraging knowledge distillation, the approach is made up of two parts: Bidirectional attack metric learning (BML) and attention-guided knowledge distillation (AGKD). To enhance the attention map for adversarial instances and repair damaged intermediate features, the AGKD module extracts clean image attention information to the student model. The BML component employs bidirectional metric learning to standardize the feature space representation. The combination of these two modules consistently surpasses cutting-edge techniques, including single-directional metric learning (SML)~\cite{DBLP:conf/nips/MaoZYVR19}, Bilateral~\cite{DBLP:conf/iccv/WangZ19}, and feature scattering (FS)~\cite{DBLP:conf/nips/ZhangW19}.
The authors further demonstrate that the model's robustness is not generated from gradient obfuscation, but rather from a slight drop in clean accuracy.

Wu \textit{et al.}~\cite{9693248} build a distilled differentiator using activation-based network pruning to decrease attack transferability, meanwhile retaining accuracy. As a two-phase defense, they use an ensemble structure of diverse differentiators. In the first step, the student model is utilized to narrow down the possible differentiators to be developed. In the second stage, a small, predetermined number of differentiators are employed to properly evaluate clean or reject hostile inputs. This solution fits the criteria for defense rate, model accuracy, and scalability. Through small-scale integration models, the architecture retains scalability, efficacy, and comparable clean input accuracy while being more efficient and simpler to implement. However, ensemble-based methods, such as boosting~\cite{DBLP:conf/stoc/AlonGHM21} and bagging~\cite{DBLP:conf/ic3/KumarS22}, would inevitably increase the computational complexity compared to non-ensemble models, due to their need for extensive model training and a combination of multiple prediction results into a final result.

\subsubsubsection{Loss Function}
Many training techniques have been developed to enhance performance by modifying or adding new regularization terms to the models' loss functions.

Mustafa \textit{et al.}~\cite{DBLP:journals/pami/MustafaKHGSS21} explain the proximity of distinct classes of samples in the learned feature space of deep learning models is the primary reason for the vulnerabilities in DNNs. As a proactive protection against adversarial attacks, they suggest a distance-based training technique, Prototype Conformity Loss (PCL), to solve this issue. This strategy aims to maximally segregate the learned feature representations at many layers of a DNN so that there is little intersection between any two classes in the decision layer as well as the intermediate feature space. Such a strategy assures that an adversarial instance with a limited perturbation budget can no longer deceive the network. The strategy significantly increases the model's robustness with a shorter training period, without diminishing the classification accuracy of clean pictures. However, the model is less resistant to white-box attacks than it is to black-box attacks since gradient masking is not utilized in its defense.

Bernhard \textit{et al.}~\cite{DBLP:conf/ijcnn/BernhardMD21} create an innovative technique known as the ``luring effect'' to prevent transferability between two models to pave a new path for robustness in a realistic black-box situation. They provide a deception-based method that is applicable to any pre-trained model and needs no labeled dataset. The target model is reinforced with a neural network designed to have an appealing effect and trained with a loss function that employs logit sequence order.

\begin{figure}[!t]
    \centering
    \includegraphics[height=1.5in]{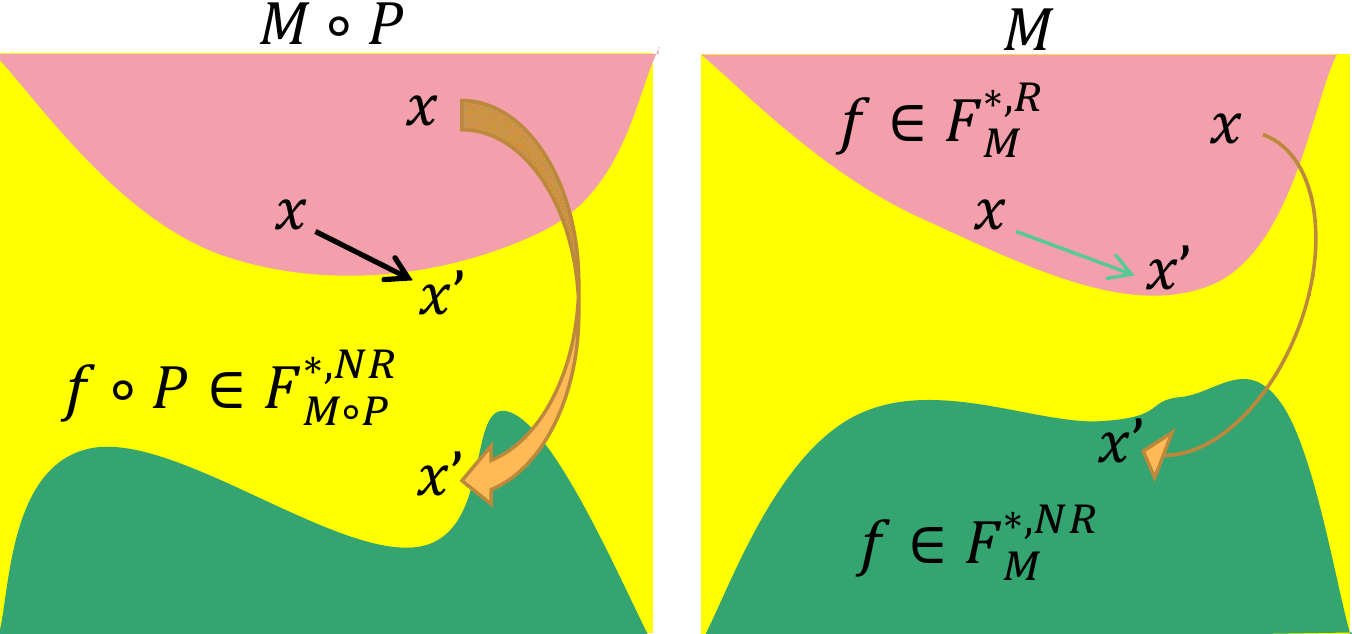}
    \caption{On the left, $x'$ dupes $M\circ P$ by flipping the non-robust feature $f\circ P$ (to the green class). On the right, however, $f$ can be a robust feature, in which case $M$ will not be deceived (still in the pink class).} 
    \label{luring}
\end{figure}

Fig.~\ref{luring} illustrates the two cases of the Luring effect. On the left, $x'$ dupes $M\circ P$ by flipping the non-robust feature $f\circ P$. However, on the right of the figure, $f$ can be a robust feature, in which case $M$ will not be deceived (still in the pink class), or a non-robust feature, in which case $f\circ P$ toggle will not be tracked. Even with massive adversarial perturbations, the approach can successfully limit the efficiency of cutting-edge transfer black-box attacks. It may be effectively coupled with existing defense strategies. The benefits of the method for robustness are more pronounced on the SVHN and MNIST datasets, but the CIFAR-10 results only perform well within the range of defense schemes that need a pre-trained model.

Like conventional method for improving adversarial resilience by limiting logit norms to tiny values~\cite{DBLP:conf/bmvc/ShafahiGNHDG19}, Kanai \textit{et al.}~\cite{DBLP:conf/ijcnn/KanaiYYTI21} introduce a function named bounded logit function (BLF), which employs a bounded activation function shortly before the softmax to confine the logit norms. BLF is constrained by finite values. Moreover, its pre-logit at the maximum or minimum points is constrained by finite values. Consequently, introducing BLF right prior to softmax gives finite values to the optimal logit and pre-logit. Despite its simplicity, the approach successfully enhances robustness in adversarial training compared to alternative logit regularization methods. However, although BLF is effective against powerful non-target attacks, it is useless against target attacks. Furthermore, the approach reveals the softmax cross-entropy's fragility as well as the efficiency of logit regularization. However, it is unclear why a small logit enhances robustness.

\subsubsubsection{Boundary Estimation} 
Liu \textit{et al.}~\cite{liu2021training} concentrate on constructing certifiers to identify certified areas of the input neighborhood where the model produces the right prediction and employing such certifiers to train a model to be verifiably resilient to adversarial attacks. They developed a stronger certifier, the polyhedral envelope certifier (PEC), as well as a regularization scheme named polyhedral envelope regularization (PER), which can be applied to networks of different architectures with general activation functions. Different from some earlier methods, such as COnvex Layer-wise adversarial Training (COLT) developed in~\cite{DBLP:conf/iclr/BalunovicV20}, PER has minimum computing cost and offers improved robustness guarantees and precision on clean data in a variety of circumstances. However, the method can be restrictive for large models, particularly for deeper networks. The boundary of the output logarithm inevitably becomes less tight, irrespective of the linearization method being used. In addition, the linear approximation is in favor of the $\ell_{\infty}$ norm over other $\ell_p$ norms, performing better in the case of $\ell_{\infty}$ than it performs in the case of~$\ell_2$.

\subsubsection{Regularization Across Layers -- Gradient Masking}\label{gradient masking}
Gradient masking here refers to a typical approach for protecting against white-box attacks that depend on model gradients. This method includes adding an extra training layer to the model, which decreases its sensitivity to tiny changes in the input data. This is often accomplished by using random noise or perturbations to obfuscate gradient information, or by employing a gradient near or at zero to neutralize or mitigate gradient-based attacks. However, gradient masking does not alter the decision boundaries. Instead, it just makes it more challenging for an attacker to influence the model using gradient information. This means that gradient masking is generally not effective against black-box attacks. The attacker can simply self-train an agent model to mimic the defense model, e.g., by observing the real discriminant labels of the input samples.

Many defense approaches have been developed based on gradient masking~\cite{DBLP:journals/access/LeeBY21}. 
Some are directly designed to perform gradient masking, such as replacing the smooth sigmoid function with a hard threshold~\cite{DBLP:conf/iccv/HowardPALSCWCTC19}. Some strategies add regularization terms with a gradient penalty component, making the model less susceptible to tiny input perturbations~\cite{DBLP:conf/ijcnn/SchwartzD21, DBLP:journals/tkde/FengHTC21}. However, this strategy has the potential to significantly diminish the model's precision and learning capability. Defensive distillation substitutes the last layer with a soft maximum function and a temperature junction to regulate the degree of distillation after the training process~\cite{9693248}.

Gradient masking has used regularizers or smoothing labels to make the model less susceptible to input perturbations. Some of these strategies include blurring or masking gradient data, akin to gradient masking. HLDR~\cite{DBLP:conf/ijcnn/SchwartzD21}, for instance, adds a regularization term to penalize the difference between benign and adversarial data representations in the hidden layer. GraphAT~\cite{DBLP:journals/tkde/FengHTC21} aims to minimize an adversarial graph regularizer and reduce prediction divergence between a disturbed target instance and its related instances. Robust CNN Training with Inf-Norm and Inf-Ind regularization~\cite{DBLP:journals/tmm/AminiG20} is used to improve the total Lipschitz constant and consistency of input-output maps. However, these methods do not use a gradient step to update the penalty parameter, which can lead to ambiguity in the adversary. In other words, they also blur the gradient.
The RP-Regularizer~\cite{DBLP:conf/ijcnn/CarboneSB21} integrates the specification of the loss gradient, intended as a metric of vulnerability, with the expectations of stochastic predictions of the inputs. To prevent steep gradients caused by binary masks, in TENET Training~\cite{DBLP:conf/iccv/LiuW00S21}, researchers propose Rectified Reverse Function (RRF) to smooth the group inversion mapping. MaxUp~\cite{DBLP:conf/cvpr/GongRY021} adds a gradient-norm smooth regularization for Gaussian perturbations. The regularization procedure in ER-Classifier~\cite{DBLP:conf/iccv/LiMLYKWH21} may assist in eliminating adversarial distortion effects and returning adversarial instances to normal data manifolds. GAT~\cite{DBLP:conf/iccv/PoursaeedJYBL21} provides variety and realism to adversarial training examples to close the distributional gap between adversarial and actual samples. AdvRush~\cite{DBLP:conf/iccv/MokNCY21} introduces regularizers for candidate architectures that smooth input loss landscapes.


Defensive distillation is an additional kind of gradient masking~\cite{DBLP:journals/ral/YunLL21}, which substitutes the last layer with a protective soft maximum function and a temperature junction to modify the level of distillation after the training process. RSLAD~\cite{DBLP:conf/iccv/ZiZMJ21} is a strategy that uses resilient soft labels created by an adversarially-trained teacher model to guide the training of students on both clean and adversarial samples. Distilled Differentiator~\cite{9693248} utilizes an ensemble structure built on specialized classifiers called differentiators and activation-based network pruning to limit attack transferability while maintaining precision.

\subsubsection{Summary}

In the past few years, researchers have made many contributions to the field of regularization methods for adversarial defense from four perspectives: Input layer, middle layer, output layer, and across layers. Regularization methods at the input layer can be divided into noise addition, adversarial learning, and feature extraction. Amongst these, adversarial learning, which is the most effective and widely used defense method, has developed many variants in recent years. Regularization methods at the middle layer focus on changing the model structure to improve its inherent robustness. The output layer regularization can be classified as distillation, decision boundary estimation, and loss functions design, and it effectively improves the robustness of the models from different perspectives. Last but not least, certain defense techniques attempt to combat adversarial attacks across neural network layers by implementing gradient masking, which is ineffective in the face of black-box attacks.

{\small
\setlength{\tabcolsep}{2pt}
\begin{table*}
\caption{Robustness Regularization on Output Layer}
	\label{outl}
	\setlength{\tabcolsep}{3pt} 
	\renewcommand\arraystretch{1.5} 
	\begin{tabular}{|p{1cm}|p{1cm}|p{1.3cm}|p{3.4cm}|p{0.8cm}|p{1.3cm}|p{3.5cm}|p{4cm}|}
    \hline
    \centering{\textbf{Defense}} & \centering{\textbf{Strategy}} & \centering{\textbf{Attack type}} & \centering{\textbf{Description}} & \centering{\textbf{Input}} & \centering{\textbf{Invisibility Metric}} & \centering{\textbf{Strength}} & \centering{\textbf{Weakness}}  \cr \hline
    
    AGKD-BML \cite{DBLP:conf/iccv/WangDYLL21} &Distil-lation & White-box, Black-box&Attention Guided Knowledge Distillation and Bi-directional Metric Learning are used to create an adversarial training-based model. & Image & $\ell_2$ & By integrating AGKD and BML, achieve SOTA robustness under different attacks. Robustness does not come from gradient obfuscation. &  AGKD-BML trained on a 7-step attack achieves much lower performance against the regular attacks. The clean accuracy has decreased a little.\\ \hline

    Distilled Differentiator \cite{9693248}& Distil-lation& White-box, Black-box&Activation-based network pruning is used to maximize the transferability of attacks and retrain their precision. & Image & $\ell_p$ & Effectively defend against targeted and non-targeted attacks, maintaining scalability, effectiveness, and comparable clean accuracy through a small-scale ensemble model.& Ensemble-based methods will inevitably increase computational complexity. \\ \hline 

   RSLAD \cite{DBLP:conf/iccv/ZiZMJ21}& Distil-lation& White-box, Black-box& A unique adversarial robustness distillation approach for training robust small student models, whereby robust soft labels are substituted for hard labels in all supervision loss terms.& Image & KL & Improve the robustness of small models against SOTA attacks, especially automated attacks; more effectively than previous adversarial training and distillation techniques. & When the teacher's network grows too complex for the student to understand, the student's resilience declines. \\ \hline

    PCL \cite{DBLP:journals/pami/MustafaKHGSS21} & Loss  &White-box, Black-box& A proactive defense against adversarial assaults, a novel distance-based training technique that aims to maximally segregate the learned feature representations at different depth layers of the deep model.& Image & PCL &  Greatly enhance the robustness of the learning model, even against the strongest white-box attacks, without clean accuracy decline, not due to obfuscated gradients, and require a shorter training time. & Robustness against white-Box settings is higher than black-box settings. \\ \hline

    Luring effect \cite{DBLP:conf/ijcnn/BernhardMD21}  &     Loss   & Black-box&A novel strategy to thwart transferability of black-box attack between two models. & Image & CE & Luring effect can be implemented at a low cost for any pre-trained model, and can successfully limit the efficiency of even the most advanced transfer-based black box attacks with large adversarial perturbations, and be effectively combined with existing defense schemes. & The advances in robustness are more evident on SVHN and MNIST, but the CIFAR-10 findings are especially encouraging in the context of a defensive system that needs a pre-trained model. \\ \hline
  
    BLF \cite{DBLP:conf/ijcnn/KanaiYYTI21}  &     Loss &White-box, Black-box&The insertion of a new limited function shortly prior to softmax improves adversarial robustness.& Image & $\ell_{\infty}$ & BLF is easily combined with AT, proving that BLF is superior to logits squeezing without AT, and is superior to or comparable to logit compression, label smoothing, and TRADES when AT is used. & Logit regularization methods without AT are insufficiently resistant to targeted assaults. It is unclear why small logit can improve robustness. \\ \hline

    PER \cite{liu2021training} &    Bound&White-box & To promote bigger adversarial-free zones in the vicinity of the input data, hence enhancing the proven robustness of the models. & Image & $\ell_p$ & Suitable for different architectures and networks with generic activation functions; the computational overhead of PER is negligible; achieve better robustness guarantees and accuracy for clean data. & Regardless of the linearization approach, the boundary of the output logarithm for large models unavoidably grows looser over deeper networks. Furthermore, the linear approximation implicitly prefers the $\ell_{\infty}$ norm over the  $\ell_p$ norm.
    \\ \hline
    
     \end{tabular}
\end{table*}
}

\subsection{Summary and Lessons Learned }
The latest adversarial defense techniques, especially those published in the past few years, have been primarily focused on adversarial detection and robustness enhancement techniques. Adversarial detection methods are simpler and more effective than modifying the original model and input images, but they can be easily bypassed by attackers. On the other hand, robustness enhancement techniques aim to improve the accuracy of the model and reduce the success rate of attacks. However, there are still challenges, such as reducing data dependency, avoiding gradient masking effects, improving model generalization, reducing the cost of model training, and improving resistance to unknown high-intensity attacks.

Combining detection and robustness-enhancing defense methods is a promising direction for future research. Existing methods, such as GAT~\cite{DBLP:conf/iccv/PoursaeedJYBL21}, XEnsemble~\cite{DBLP:journals/tdsc/WeiL21}, APR~\cite{DBLP:conf/iccv/ChenPM0D021}, and ER-Classifier~\cite{DBLP:conf/iccv/LiMLYKWH21}, have already contributed to this perspective. Future research is anticipated to focus on developing methods that combine the advantages of both detection and robustness enhancement, while addressing the challenges mentioned above. This will help to improve the overall effectiveness of adversarial defenses.

\begin{figure*}[t]
        \centering
        \includegraphics[height=4.8in]{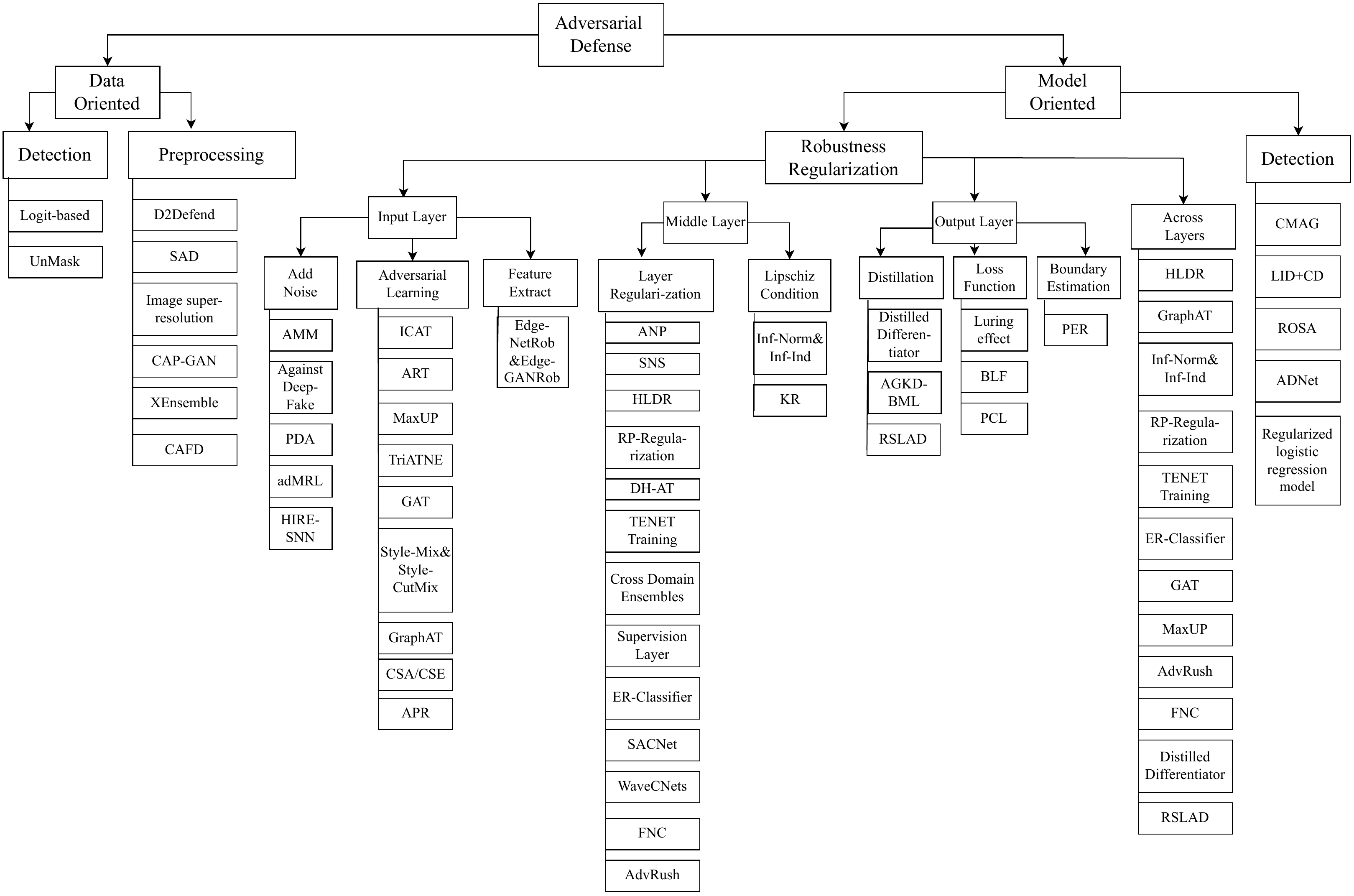}
        \caption{Anatomy of the recent breakthroughs in adversarial defenses since 2021. We classify adversarial defense methods as data-oriented and model-oriented methods. From the model-oriented angle, we further divide defense methods into robustness regularization in different layers and detection.} 
        \label{fig:defense}
\end{figure*} 

\section{Lessons Learned and Open Issues}
\label{future}

In this section, we summarize the important lessons learned from the comprehensive analysis of the recent research outcomes in adversarial attacks and defenses, devise the remaining open challenges and point to research opportunities in this rapidly growing, important area.

\subsection{Lessons and Challenges of Adversarial Attacks}

\begin{figure}
    \centering
    \includegraphics[width=3.5in]{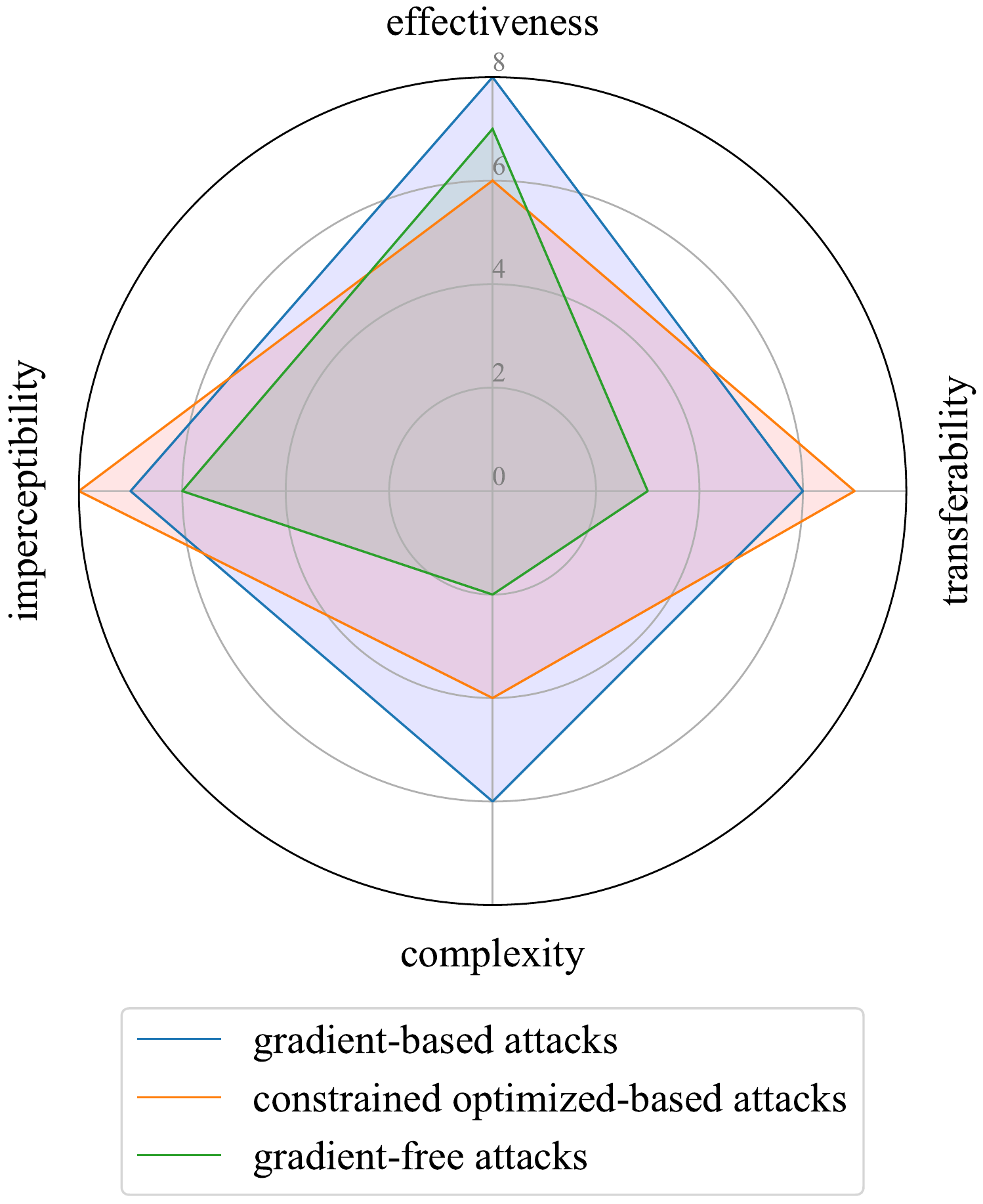}
    \caption{The balance of four factors in adversarial attacks.}
    \label{fig: spider web}
\end{figure}

It is important to strike a balance between effectiveness, imperceptibility, complexity, and transferability in adversarial attacks, among which there are obvious trade-offs. As illustrated in Fig.~\ref{fig: spider web}, 
\begin{itemize}
     \item Gradient-based attack methods, such as LAFEAT~\cite{DBLP:conf/cvpr/YuG021}, DSNGD~\cite{DBLP:conf/ijcnn/SchwinnNRZETB21}, SGA.~\cite{DBLP:journals/tkde/LiXCXHZ23}, are known for their high ASRs and good transferability. However, the methods have limitations of high computational and time costs, as well as the issue of ``gradient saturation," which reduces their effectiveness. Gradient-based attacks limit the perturbation to a certain size during the generation of the perturbation, guaranteeing invisibility.

    \item  Constrained optimization-based attack methods, such as GF-Attack~\cite{9720115}, AdMRL~\cite{DBLP:conf/ijcnn/ChenCW21}, and 
    SSAH~\cite{DBLP:conf/cvpr/LuoL0WXS22}, 
    have good transferability. However, they are known for high computational and time costs, making it difficult to use them in time-sensitive applications. Constraint optimization-based attack methods can guarantee small visibility of the attack by constraining the strength of the perturbation, providing greater stealth.
    
    \item  Search-based attack methods, such as FeaCP~\cite{DBLP:journals/tifs/LiQHQLD21}, are highly transferable and can be extended to other domains beyond image classification. Nevertheless, for more complex datasets, such as ImageNet~\cite{DBLP:conf/ijcnn/YaoHHP21}, searching for the optimal adversarial sample needs significantly more iterations and more computational cost. It can be difficult to find an appropriate search start point for search~\cite{2021396}. Current search-based methods are primarily applied to the initialization or optimization of other adversarial sample generation algorithms, such as 
    Square attack~\cite{DBLP:conf/eccv/AndriushchenkoC20}. The search-based perturbation does not make use of gradient information, and the perturbation magnitude of each search step is controlled to fall under a fixed range to ensure a certain invisibility.
\end{itemize}

Future efforts are expected to reduce attack costs, improve the transferability of attacks across different datasets and models, and extend the attacks to more deep-learning tasks. 

{\small
\setlength{\tabcolsep}{2pt}
\begin{table*}
\caption{Lessons Learned and Open Issues}
	\label{lesson}
	\setlength{\tabcolsep}{3pt} 
	\renewcommand\arraystretch{1.5} 
	\begin{tabular}{|p{1cm}|p{8cm}|p{8cm}|}
    \hline
    \centering{\textbf{Field}} &  \centering{\textbf{Lessons Learned}} &\centering{\textbf{ Open Issues}} \cr
    \hline
    Attack & 
        $\sbullet[.75]$ Gradient-based attack methods: High ASRs and good transferability, but high computational and time costs, as well as the issue of ``gradient saturation''.
        
        $\sbullet[.75]$ Constrained optimization-based attack: Good transferability but high computational and time costs.
        
        $\sbullet[.75]$ Search-based attack methods: Highly transferable but difficult to find an appropriate search start point.
    & 
    $\sbullet[.75]$ Reducing attack costs.
    
    $\sbullet[.75]$ Improving transferability across different datasets and models.
    
   $\sbullet[.75]$  Extending to more deep learning tasks.
    \\ \hline 
    
    Defense &
     $\sbullet[.75]$   Trade-off between defense effectiveness and cost: Training the model in a complex environment or structural improvements greatly boosts robustness with a cost of additional computational complexity.

     $\sbullet[.75]$  Unanswered Root Cause of Robustness Loss: For bounding the Lipschitz Continuity of the gradient of DNNs, including Gradient Clip, Weight Decay, and gradient masking, which is controversial in the case of adversarial defenses.

     $\sbullet[.75]$  Scalability and Generalizability: It is generally hard to apply a defense method that is effective on one DNN model or dataset, to other complex DNN models or datasets.

 $\sbullet[.75]$  Data-driven: Detection methods for adversarial samples are more data-driven, but current research on detection techniques is limited by the lack of consensus on adversarial samples at the mathematical level.
    & 
    $\sbullet[.75]$ Developing methods that combine both detection and robustness enhancement.
    
    $\sbullet[.75]$ Addressing challenges such as reducing data dependency, avoiding gradient masking effects, improving model generalization, reducing the cost of model training, and improving resistance to unknown high-intensity attacks.
    
    $\sbullet[.75]$ Overall improving the effectiveness of adversarial defense techniques.

    $\sbullet[.75]$ A combination of detection and defense approaches is the research trend.
    \\ \hline   
     \end{tabular}
\end{table*}
}

\subsection{Lessons and Challenges of Defenses }
It is crucial to provide effective and reliable countermeasures to adversarial attacks for an apparent reason. Existing adversarial defenses can benefit from continuing development to address the following challenges. 

\subsubsection{Trade-off between Defense Effectiveness and Overhead}
There is clearly a trade-off between defense effectiveness and defense overhead, e.g., training time and/or complexity. 
  
    One way to make a model more robust against potential threats is to train the model in a complex environment.
    For example, AdMRL~\cite{DBLP:conf/ijcnn/ChenCW21} enables agents to learn the initial parameters with better generalization ability by accumulating knowledge in difficult environments. PDA~\cite{DBLP:journals/tip/YuLLYZ21} uses different magnitudes of adversarial samples to increase the diversity of data.
    AMM~\cite{DBLP:journals/pami/YanGZ21} focuses on the minimum (instance-specific) margin, which is often considered a key factor in determining a model's generalization capability. A principled regularizer is derived to improve the model's performance on unseen samples with certain types of distortions. 
    However, the requirements of training time and memory space are greatly increased, as a result of iterative updating processes and the use of higher-order gradients in the optimization process. 

    Another way to boost robustness is structural improvements, e.g., adding more layers or units to a neural network~\cite{DBLP:conf/ijcnn/YaoG21}, typically at the expense of additional computational complexity. 
    The increased complexity can also lead to higher memory requirements and longer inference times, which can impact the overall performance of the model. 
    For instance, the upgraded model in DH-AT~\cite{DBLP:conf/ijcnn/JiangMEB21} contains two heads for robustness and clean accuracy independently, at the expense of higher training time. The framework for classification based on optimum transport in~\cite{DBLP:conf/cvpr/SerrurierMGBLB21} contains a KR regularization approach and more accurate constant evaluation of convolution and pooling layers, but almost triples the training time. Because self-attentional learning and context coding raise the computing overhead of the overall framework, SACNet~\cite{DBLP:journals/tip/XuDZ21} also has a somewhat lengthy execution time. Due to the employment of numerous models, several ensemble-based methods~\cite{9693248,DBLP:conf/ijcnn/CarboneSB21} generally increase computational complexity.
    
    There have been many noteworthy attempts to reduce training costs, including transfer learning, partial training/updating, optimizing training epochs, and parallel training. 
    For example, an adaptive retraining process is used in ART~\cite{DBLP:conf/ijcnn/YaoHHP21}. The Luring effect~\cite{DBLP:conf/ijcnn/BernhardMD21} can be used to improve a trained model at a low cost, since it does not require labeled datasets. It is also possible to introduce additional regularization items, such as HLDR~\cite{DBLP:conf/ijcnn/SchwartzD21}, or layers, such as FNC~\cite{DBLP:conf/iccv/YuCXLWBM21}, to neural networks for performance improvements without increasing model parameters.
    In HIRE-SNN~\cite{2021HIRE}, the weight updating only takes place after $T$ steps, thereby reducing the training cost while allowing different adversarial image variants to train the model. 
    The iterative approach is only used to the unfrozen layer for the Distilled Differentiator~\cite{9693248}. PDA~\cite{DBLP:journals/tip/YuLLYZ21} selects the iterative steps that best balance robustness, precision, and computing cost. To parallelize training, RP-Ensemble~\cite{DBLP:conf/ijcnn/CarboneSB21} creates classifiers in separately projected subspaces. In addition, models may be trained concurrently in Distilled Differentiator~\cite{9693248}. These efforts mitigate the trade-off between defense efficacy and expense to some degree, but they do not eliminate the trade-off.

\subsubsection{Unanswered Root Cause of Robustness Loss}
Lipschitz continuity is a mathematical concept used to measure the degree of continuity between two functions~\cite{2000Nonlinear}. 
More specifically, a function $f$ from $S \subset \mathbb{R}^{n}$ into $\mathbb{R}^{m}$ is Lipschitz continuous at $x \in S$ if there exists such a constant $C$ that
$$\left \|  f(y) - f(x) \right \|  \le C \left \|  y - x \right \|, $$
for all $y \in S$ close to $x$.
In the context of deep learning, Lipschitz continuity is used to measure the robustness of DNNs by assessing how small changes in the inputs affect the outputs of the DNNs. In other words, a Lipschitz-continuous function has a fixed-ratio bound on the distances between the corresponding outputs of two points close to each other in the input space~\cite{2000Nonlinear}. A DNN is said to be Lipschitz-continuous if small changes in its inputs can only cause small changes in its outputs, implicating the stability of a DNN in the face of noisy data or unexpected inputs.

There are several methods for bounding the Lipschitz Continuity of the gradient of DNNs. A few common methods are Gradient Clip~\cite{DBLP:conf/nips/ChenWH20}, which involves clipping the gradients to ensure they do not exceed a threshold, and Weight Decay~\cite{DBLP:conf/nips/LobachevaKCMV21}, which involves adding a regularization term to the loss function of a DNN to limit the magnitude of the gradients. 

Another popular method for bounding the Lipschitz Continuity of the gradient of DNNs is gradient masking~\cite{DBLP:journals/access/LeeBY21}, which involves adding a penalty term to the loss function to prevent abrupt changes in gradients.
However, gradient masking is controversial in the case of adversarial defenses~\cite{DBLP:journals/access/LeeBY21}.
Some researchers have argued that although gradient masking-based defense techniques, e.g., ER-Classifier~\cite{DBLP:conf/iccv/LiMLYKWH21} and Inf-Norm\&Inf-Ind\cite{DBLP:journals/tmm/AminiG20}, deliver effective defense against adversarial attacks in some cases, they do not address the root cause of adversarial attacks~\cite{DBLP:journals/access/LeeBY21}, which is the lack of Lipschitz Continuity.
As a consequence, gradient masking approaches are vulnerable to attacks that are independent of the gradients of the models under attack, such as high-intensity black-box attacks~\cite{DBLP:journals/access/LeeBY21}. 

\subsubsection{Scalability and Generalizability}
To defend against (new) adversarial attacks, one approach is to design more effective and robust neural network models~\cite{DBLP:conf/iclr/MadryMSTV18}, and the other is to block malicious inputs before it is fed into the model~\cite{DBLP:journals/tie/DuCSWFK20}. Since most heuristic defense strategies are unable to defend against adaptive white-box attacks, many researchers have begun to focus on provable defense mechanisms that guarantee a certain level of defense performance, irrespective of the attacker's method of attack~\cite{2021TriATNE}. 

Scalability has been a key issue to the majority of existing provable defense approaches, e.g., the PGD method developed in~\cite{DBLP:conf/iclr/MadryMSTV18}. Proof-based defense strategies are effective in defending against less sophisticated ``shadow'' neural networks, but are ineffective in the case of more advanced ``deep'' neural networks~\cite{DBLP:journals/access/DhilleswararaoB22}. Moreover, while provable defense methods work satisfactorily on small-scale datasets, e.g., the CIFAR-10 dataset with only ten classes, they deteriorate on more difficult tasks, such as classification on the ImageNet dataset that consists of a thousand major classes~\cite{DBLP:conf/ijcnn/YaoHHP21}.

Generalizability, also known as transferability, is another major concern of provable defense approaches. 
Specifically, it is generally hard to apply a defense method that is effective on one DNN model or dataset, to other DNN models or datasets~\cite{DBLP:conf/ijcnn/CarboneSB21}. 
One approach may yield satisfactory results on homogeneous networks, but performs poorly on heterogeneous networks~\cite{2021TriATNE}.

\subsubsection{Dependence on Data}
Adversarial sample detection has traditionally relied on data-driven methods. However, there is a lack of agreement on the mathematical definition of adversarial samples, limiting current research in this field~\cite{1315}. Attackers can easily bypass detections by exploiting the knowledge of the detection mechanisms, rendering the detection mechanisms ineffective~\cite{DBLP:conf/iwcmc/UsamaALQA19}. To overcome this challenge, a mixed approach that integrates both detection and defense strategies can offer a promising solution. The defense component aims to enhance accuracy and decrease attack success rate, while reducing data dependence and increasing resilience to high-intensity attacks. These are important research areas in pursuit of robust defense mechanisms. 

\section{Conclusion}
\label{conclusion}
In this survey, we have provided a comprehensive overview of the recent advancements in adversarial attacks and defenses in the field of machine learning and deep neural networks. We have analyzed both the attack techniques, including those based on constrained optimization and gradient-based optimization, and their adaptations to different threat models, such as white-box, gray-box, and black-box attacks. We have reviewed the latest defense strategies against adversarial examples, including detection and robustness improvement, which mainly focus on enhancing robustness through regularization, data augmentation, and structure optimization. Moreover, the transferability of adversarial attacks has been thoroughly investigated, providing deeper insights into the workings of deep learning models. It is expected that this survey serves as a foundation for future research in this rapidly evolving field, and provides useful information for researchers and security practitioners. 

\bibliographystyle{IEEEtran}
\bibliography{references}

\begin{thebibliography}{100}
\providecommand{\url}[1]{#1}
\csname url@samestyle\endcsname
\providecommand{\newblock}{\relax}
\providecommand{\bibinfo}[2]{#2}
\providecommand{\BIBentrySTDinterwordspacing}{\spaceskip=0pt\relax}
\providecommand{\BIBentryALTinterwordstretchfactor}{4}
\providecommand{\BIBentryALTinterwordspacing}{\spaceskip=\fontdimen2\font plus
\BIBentryALTinterwordstretchfactor\fontdimen3\font minus
  \fontdimen4\font\relax}
\providecommand{\BIBforeignlanguage}[2]{{%
\expandafter\ifx\csname l@#1\endcsname\relax
\typeout{** WARNING: IEEEtran.bst: No hyphenation pattern has been}%
\typeout{** loaded for the language `#1'. Using the pattern for}%
\typeout{** the default language instead.}%
\else
\language=\csname l@#1\endcsname
\fi
#2}}
\providecommand{\BIBdecl}{\relax}
\BIBdecl

\bibitem{DBLP:journals/access/HeLH23}
Q.~He, J.~Liu, and Z.~Huang, ``{WSRC:} {W}eakly supervised faster {RCNN} toward
  accurate traffic object detection,'' \emph{{IEEE} Access}, vol.~11, pp.
  1445--1455, 2023.

\bibitem{DBLP:conf/dsmlai/KaurS21a}
B.~Kaur and S.~Singh, ``Object detection using deep learning: {A} review,'' in
  \emph{Proc. {Int. Conf. Data Sci. Mach. Learn. Artif. Intell. (DSMLAI'21)},
  {\normalfont Windhoek Namibia, August 9 - 12, 2021}}, pp. 328--334.

\bibitem{DBLP:journals/apin/XuGLBLZ23}
C.~Xu, W.~Gao, T.~Li, N.~Bai, {GAN}g Li, and Y.~Zhang, ``Teacher-student
  collaborative knowledge distillation for image classification,'' \emph{Appl.
  Intell.}, vol.~53, no.~2, pp. 1997--2009, 2023.

\bibitem{DBLP:conf/icitee/Liu21}
L.~Liu, ``Improved image classification accuracy by convolutional neural
  networks,'' in \emph{Proc. {Int. Conf. Inf. Technol. Electr. Eng.
  (ICITEE'21)}, {\normalfont Changde, Hunan, China, October 29 - 31, 2021}},
  pp. 1--5.

\bibitem{DBLP:conf/acl/AngelovaA022}
G.~Angelova, E.~Avramidis, and S.~M{\"{o}}ller, ``Using neural machine
  translation methods for sign language translation,'' in \emph{Proc. {Annu.
  Meet. Assoc. Comput. Ling.: Stud. Res. Workshop (ACL'22)}, {\normalfont
  Dublin, Ireland, May 22-27, 2022}}, pp. 273--284.

\bibitem{DBLP:journals/taccess/Ananthanarayana21}
T.~Ananthanarayana, P.~Srivastava, A.~Chintha, A.~Santha, B.~Landy, J.~Panaro,
  A.~Webster, N.~R. Kotecha, S.~Sah, T.~Sarchet, R.~W. Ptucha, and I.~Nwogu,
  ``Deep learning methods for sign language translation,'' \emph{{ACM} Trans.
  Accessible Comput.}, vol.~14, no.~4, pp. 1--30, 2021.

\bibitem{DBLP:journals/air/ChanBLPC23}
J.~Y. Chan, K.~T. Bea, S.~M.~H. Leow, S.~W. Phoong, and W.~K. Cheng, ``State of
  the art: {A} review of sentiment analysis based on sequential transfer
  learning,'' \emph{Artif. Intell. Rev.}, vol.~56, no.~1, pp. 749--780, 2023.

\bibitem{DBLP:journals/apin/SunTDB23}
Z.~Sun, L.~Tian, Q.~Du, and J.~A. Bhutto, ``Sample hardness guided softmax loss
  for face recognition,'' \emph{Appl. Intell.}, vol.~53, no.~3, pp. 2640--2655,
  2023.

\bibitem{DBLP:journals/csur/AbdullahA23}
T.~Abdullah and A.~Ahmet, ``Deep learning in sentiment analysis: {R}ecent
  architectures,'' \emph{{ACM} Comput. Surv.}, vol.~55, no.~8, pp. 1--37, 2023.

\bibitem{DBLP:conf/5gwf/KelkarD21}
A.~Kelkar and C.~Dick, ``{NVIDIA} aerial {GPU} hosted {AI}-on-5{G},'' in
  \emph{Proc. {IEEE} 5{G} {W}orld {F}orum ({WF-5G'21}), {\normalfont Montreal,
  QC, Canada, October 13-15, 2021}}, pp. 64--69.

\bibitem{10.1145/3487890}
Y.~Hu, W.~Kuang, Z.~Qin, K.~Li, J.~Zhang, Y.~Gao, W.~Li, and K.~Li,
  ``Artificial intelligence security: {T}hreats and counter measures,''
  \emph{ACM Comput. Surv.}, vol.~55, no.~20, pp. 1--36, 2021.

\bibitem{DBLP:conf/iclr/MadryMSTV18}
A.~Madry, A.~Makelov, L.~Schmidt, D.~Tsipras, and A.~Vladu, ``Towards deep
  learning models resistant to adversarial attacks,'' in \emph{Proc. {Int.
  Conf. Learn. Representations (ICLR'18)}, {\normalfont Vancouver, BC, Canada,
  April 30 - May 3, 2018}}, pp. 1--23.

\bibitem{DBLP:conf/eccv/AndriushchenkoC20}
M.~Andriushchenko, F.~Croce, N.~Flammarion, and M.~Hein, ``Square attack: {A}
  query-efficient black-box adversarial attack via random search,'' in
  \emph{Proc. {Eur. Conf. Comput. Vision (ECCV'20)}, {\normalfont Glasgow, UK,
  August 23-28, 2020}}, pp. 484--501.

\bibitem{DBLP:conf/sp/Carlini017}
N.~Carlini and D.~A. Wagner, ``Towards evaluating the robustness of neural
  networks,'' in \emph{Proc. {IEEE} {Symp. Secur. Privacy (SP'17) },
  {\normalfont Jose, CA, USA, May 22-26, 2017}}, pp. 39--57.

\bibitem{DBLP:conf/ijcnn/YaoG21}
Z.~Yao and J.~Gao, ``Adversarial example defense based on the supervision,'' in
  \emph{Proc. {Int. Joint Conf. Neural Networks (IJCNN'21)}, {\normalfont
  Shenzhen, China, July 18-22, 2021}}, pp. 1--8.

\bibitem{DBLP:conf/cvpr/DuanMQCYHY21}
R.~Duan, X.~Mao, A.~K. Qin, Y.~Chen, S.~Ye, Y.~He, and Y.~Yang, ``Adversarial
  laser beam: Effective physical-world attack to {DNN}s in a blink,'' in
  \emph{Proc. {IEEE Conf. Comput. Vis. Pattern Recognit. (CVPR'21)},
  {\normalfont Virtual Event, Nashville, TN, USA, June 19-25, 2021}}, pp.
  16\,062--16\,071.

\bibitem{9918564}
L.~Xu, X.~Zheng, X.~Li, Y.~Zhang, L.~Liu, and H.~Ma, ``{WiCAM: I}mperceptible
  adversarial attack on deep learning based {W}i{F}i sensing,'' in \emph{Proc.
  {Annu. {IEEE} Int. Conf. Sens. Commun. Networking ({SECON}'22)}, {\normalfont
  Stockholm, Sweden, September 20-23, 2022}}, pp. 10--18.

\bibitem{9609969}
B.~Kim, Y.~E. Sagduyu, K.~Davaslioglu, T.~Erpek, and S.~Ulukus, ``Channel-aware
  adversarial attacks against deep learning-based wireless signal
  classifiers,'' \emph{IEEE Trans. Wirel. Commun.}, vol.~21, no.~6, pp.
  3868--3880, 2022.

\bibitem{9816059}
G.~Apruzzese, R.~Vladimirov, A.~Tastemirova, and P.~Laskov, ``Wild networks:
  {E}xposure of {5G} network infrastructures to adversarial examples,''
  \emph{{IEEE} Trans. Network Serv. Manage.}, vol.~19, no.~4, pp. 5312--5332,
  2022.

\bibitem{9734753}
L.~Zhang, S.~Lambotharan, G.~Zheng, G.~Liao, A.~Demontis, and F.~Roli, ``A
  hybrid training-time and run-time defense against adversarial attacks in
  modulation classification,'' \emph{IEEE Wirel. Commun. Lett.}, vol.~11,
  no.~6, pp. 1161--1165, 2022.

\bibitem{DBLP:journals/csur/MirskyL21}
Y.~Mirsky and W.~Lee, ``The creation and detection of deepfakes: {A} survey,''
  \emph{{ACM} Comput. Surv.}, vol.~54, no.~1, pp. 7:1--7:41, 2022.

\bibitem{DBLP:conf/ivsp/YinUD21}
Z.~Yin, K.~Uchida, and S.~Deng, ``Improving adversarial attacks on face
  recognition using a modified image translation model,'' in \emph{Proc. {Int.
  Conf. Image, Video Signal Process. (IVSP'21)}, {\normalfont Singapore, March
  19-21, 2021}}, pp. 26--31.

\bibitem{DBLP:journals/tcps/JiangHZPA21}
W.~Jiang, Z.~He, J.~Zhan, W.~Pan, and D.~Adhikari, ``Research progress and
  challenges on application-driven adversarial examples: {A} survey,''
  \emph{{ACM} Trans. Cyber-Phys. Syst.}, vol.~5, no.~4, pp. 1--25, 2021.

\bibitem{DBLP:conf/kdd/FursovMKKRGBK0B21}
I.~Fursov, M.~Morozov, N.~Kaploukhaya, E.~Kovtun, R.~R. Castro, G.~Gusev,
  D.~Babaev, I.~Kireev, A.~Zaytsev, and E.~Burnaev, ``Adversarial attacks on
  deep models for financial transaction records,'' in \emph{Proc. {ACM SIGKDD
  Conf. Knowl. Discovery Data Min. (KDD'21)}, {\normalfont Virtual Event,
  Singapore, August 14-18, 2021}}, pp. 2868--2878.

\bibitem{10.1145/3440084.3441213}
S.~Khan and M.~R. Rabbani, ``Chatbot as islamic finance expert ({CaIFE}):
  {W}hen finance meets artificial intelligence,'' in \emph{Proc. {Int. Symp.
  Comput. Commun. (ISCC'21)}, {\normalfont Newcastle upon Tyne, United Kingdom,
  2021}}, pp. 1--5.

\bibitem{2019A}
G.~E. Kaiqiang and T.~Chen, ``A survey of attack and defense on human-computer
  interaction security,'' \emph{Telecommun. Sci.}, vol.~35, pp. 100--116, 2019.

\bibitem{DBLP:conf/chi/ParkAHL21}
H.~Park, D.~Ahn, K.~Hosanagar, and J.~Lee, ``Human-{AI} interaction in human
  resource management: {U}nderstanding why employees resist algorithmic
  evaluation at workplaces and how to mitigate burdens,'' in \emph{Proc. {ACM
  CHI Conf. Hum. Factors Comput. Syst. (CHI'21)}, {\normalfont Virtual Event,
  Yokohama, Japan, May 8-13, 2021}}, pp. 1--15.

\bibitem{9259112}
Y.~Lin, H.~Zhao, X.~Ma, Y.~Tu, and M.~Wang, ``Adversarial attacks in modulation
  recognition with convolutional neural networks,'' \emph{IEEE Trans. Reliab.},
  vol.~70, no.~1, pp. 389--401, 2021.

\bibitem{9893902}
Y.~Ye, Y.~Chen, and M.~Liu, ``Multiuser adversarial attack on deep learning for
  {OFDM} detection,'' \emph{IEEE Wirel. Commun. Lett.}, vol.~11, no.~12, pp.
  2527--2531, 2022.

\bibitem{DBLP:conf/ijcai/XiaoLZHLS18}
C.~Xiao, B.~Li, J.~Y. Zhu, W.~He, M.~Liu, and D.~Song, ``Generating adversarial
  examples with adversarial networks,'' in \emph{Proc. {Int. Joint Conf. Artif.
  Intell. (IJCAI'18)}, {\normalfont Stockholm, Sweden, July 13-19, 2018}}, pp.
  3905--3911.

\bibitem{9756577}
P.~Freitas~de Araujo~Filho, G.~Kaddoum, M.~Naili, E.~T. Fapi, and Z.~Zhu,
  ``Multi-objective {GAN}-based adversarial attack technique for modulation
  classifiers,'' \emph{{IEEE} Commun. Lett.}, vol.~26, no.~7, pp. 1583--1587,
  2022.

\bibitem{9144305}
Y.~Shi, K.~Davaslioglu, and Y.~E. Sagduyu, ``Generative adversarial network in
  the air: Deep adversarial learning for wireless signal spoofing,'' \emph{IEEE
  Trans. Cognit. Commun. Networking}, vol.~7, no.~1, pp. 294--303, 2021.

\bibitem{9390408}
H.~Lv, M.~Wen, R.~Lu, and J.~Li, ``An adversarial attack based on incremental
  learning techniques for unmanned in 6{G} scenes,'' \emph{IEEE Trans. Veh.
  Technol.}, vol.~70, no.~6, pp. 5254--5264, 2021.

\bibitem{9531421}
P.~Huang, X.~Zhang, S.~Yu, and L.~Guo, ``{IS-WARS: I}ntelligent and stealthy
  adversarial attack to {Wi-Fi}-based human activity recognition systems,''
  \emph{IEEE Trans. Dependable Secure Comput.}, vol.~19, no.~6, pp. 3899--3912,
  2022.

\bibitem{DBLP:journals/corr/GoodfellowSS14}
I.~J. Goodfellow, J.~Shlens, and C.~Szegedy, ``{E}xplaining and harnessing
  adversarial examples,'' in \emph{Proc. {Int. Conf. Learn. Representations
  (ICLR'15)}, {\normalfont San Diego, CA, USA, May 7-9, 2015 }}, pp. 1--11.

\bibitem{10015738}
F.~Xiao, Y.~Huang, Y.~Zuo, W.~Kuang, and W.~Wang, ``Over-the-air adversarial
  attacks on deep learning {Wi-Fi} fingerprinting,'' \emph{IEEE Internet Things
  J.}, pp. 1--12, 2023 (early access).

\bibitem{9477416}
D.~Xu, H.~Yang, C.~Gu, Z.~Chen, Q.~Xuan, and X.~Yang, ``Adversarial examples
  detection of radio signals based on multifeature fusion,'' \emph{{IEEE}
  Trans. Circuits Syst.}, vol.~68, no.~12, pp. 3607--3611, 2021.

\bibitem{9887932}
Z.~Wang, W.~Liu, and H.~M. Wang, ``{GAN} against adversarial attacks in radio
  signal classification,'' \emph{IEEE Commun. Lett.}, vol.~26, no.~12, pp.
  2851--2854, 2022.

\bibitem{9542973}
R.~Sahay, C.~G. Brinton, and D.~J. Love, ``A deep ensemble-based wireless
  receiver architecture for mitigating adversarial attacks in automatic
  modulation classification,'' \emph{IEEE Trans. Cognit. Commun. Networking},
  vol.~8, no.~1, pp. 71--85, 2022.

\bibitem{9916315}
K.~W. McClintick, J.~Harer, B.~Flowers, W.~C. Headley, and A.~M. Wyglinski,
  ``Countering physical eavesdropper evasion with adversarial training,''
  \emph{{IEEE} {Open J. Commun. Soc. }}, vol.~3, pp. 1820--1833, 2022.

\bibitem{DBLP:conf/cvpr/Moosavi-Dezfooli17}
M.~D. Seyed~Mohsen, A.~Fawzi, O.~Fawzi, and P.~Frossard, ``Universal
  adversarial perturbations,'' in \emph{Proc. {IEEE Conf. Comput. Vis. Pattern
  Recognit. (CVPR'17)}, {\normalfont Honolulu, HI, USA, July 21-26, 2017}}, pp.
  86--94.

\bibitem{9328496}
A.~M. Sadeghzadeh, S.~Shiravi, and R.~Jalili, ``Adversarial network traffic:
  {T}owards evaluating the robustness of deep-learning-based network traffic
  classification,'' \emph{{IEEE} Trans. Network Serv. Manage.}, vol.~18, no.~2,
  pp. 1962--1976, 2021.

\bibitem{9747933}
E.~Nowroozi, Y.~Mekdad, M.~H. Berenjestanaki, M.~Conti, and A.~E. Fergougui,
  ``Demystifying the transferability of adversarial attacks in computer
  networks,'' \emph{{IEEE} Trans. Network Serv. Manage.}, vol.~19, no.~3, pp.
  3387--3400, 2022.

\bibitem{DBLP:conf/eurosp/PapernotMJFCS16}
N.~Papernot, P.~D. McDaniel, S.~Jha, M.~Fredrikson, Z.~B. Celik, and A.~Swami,
  ``The limitations of deep learning in adversarial settings,'' in \emph{Proc.
  {IEEE Eur. Symp. Secur. Privacy (EuroS{\&}P'16)}, {\normalfont
  Saarbr{\"{u}}cken, Germany, March 21-24, 2016}}, pp. 372--387.

\bibitem{DBLP:conf/iclr/KurakinGB17}
A.~Kurakin, I.~J. Goodfellow, and S.~Bengio, ``Adversarial machine learning at
  scale,'' in \emph{Proc. {Int. Conf. Learn. Representations (ICLR'17)},
  {\normalfont Toulon, France, April 24-26, 2017}}, pp. 1--17.

\bibitem{DBLP:journals/corr/SzegedyZSBEGF13}
C.~Szegedy, W.~Zaremba, I.~Sutskever, J.~Bruna, D.~Erhan, I.~J. Goodfellow, and
  R.~Fergus, ``Intriguing properties of neural networks,'' in \emph{Proc. {Int.
  Conf. Learn. Representations (ICLR'14)}, {\normalfont Banff, AB, Canada,
  April 14-16, 2014}}, pp. 1--10.

\bibitem{DBLP:conf/cvpr/Moosavi-Dezfooli16}
M.~D. Seyed~Mohsen, A.~Fawzi, and P.~Frossard, ``Deep{F}ool: {A} simple and
  accurate method to fool deep neural networks,'' in \emph{Proc. {IEEE Conf.
  Comput. Vis. Pattern Recognit. (CVPR'16)}, {\normalfont Las Vegas, NV, USA,
  June 27-30, 2016}}, pp. 2574--2582.

\bibitem{9674195}
C.~Zhang, X.~Costa~Pérez, and P.~Patras, ``Adversarial attacks against deep
  learning-based network intrusion detection systems and defense mechanisms,''
  \emph{IEEE ACM Trans. Networking}, vol.~30, no.~3, pp. 1294--1311, 2022.

\bibitem{9448103}
D.~Han, Z.~Wang, Y.~Zhong, W.~Chen, J.~Yang, S.~Lu, X.~Shi, and X.~Yin,
  ``Evaluating and improving adversarial robustness of machine learning-based
  network intrusion detectors,'' \emph{IEEE J. Sel. Areas Commun.}, vol.~39,
  no.~8, pp. 2632--2647, 2021.

\bibitem{DBLP:conf/www/TariqJW22}
S.~Tariq, S.~Jeon, and S.~S. Woo, ``Am {I} a real or fake celebrity?
  {E}valuating face recognition and verification {API}s under {D}eep{f}ake
  impersonation attack,'' in \emph{Proc, {World Wide Web (WWW'22)},
  {\normalfont Virtual Event, Lyon, France, April 25 - 29, 2022}}, pp.
  512--523.

\bibitem{DBLP:journals/tissec/QinPLRB22}
L.~Qin, F.~Peng, M.~Long, R.~Ramachandra, and C.~Busch, ``Vulnerabilities of
  unattended face verification systems to facial components-based presentation
  attacks: {A}n empirical study,'' \emph{{ACM} Trans. Privacy Secur.}, vol.~25,
  no.~1, pp. 1--28, 2022.

\bibitem{DBLP:journals/access/IlioudiDWS22}
A.~Ilioudi, A.~Dabiri, B.~J. Wolf, and B.~D. Schutter, ``Deep learning for
  object detection and segmentation in videos: {T}oward an integration with
  domain knowledge,'' \emph{{IEEE} Access}, vol.~10, pp. 34\,562--34\,576,
  2022.

\bibitem{DBLP:conf/kdd/LiuGJ0D21}
B.~Liu, Y.~Guo, J.~Jiang, J.~Tang, and W.~Deng, ``Multi-view correlation based
  black-box adversarial attack for 3{D} object detection,'' in \emph{Proc. {ACM
  SIGKDD Conf. Knowl. Discovery Data Min. (KDD'21)}, {\normalfont Virtual
  Event, Singapore, August 14-18, 2021}}, pp. 1036--1044.

\bibitem{DBLP:conf/icons2/OrthSPD22}
A.~Orth, T.~C. Stewart, M.~Picard, and M.~A. Drouin, ``Towards a laser warning
  system in the visible spectrum using a neuromorphic camera,'' in \emph{Proc.
  {Int. Conf. Neuromorph. Syst. (ICONS'22)}, {\normalfont Knoxville, TN, USA,
  July 27 - 29, 2022}}, pp. 1--4.

\bibitem{DBLP:journals/iotj/YangLZLT21}
X.~Yang, W.~Liu, S.~Zhang, W.~Liu, and D.~Tao, ``Targeted attention attack on
  deep learning models in road sign recognition,'' \emph{{IEEE} Internet Things
  J.}, vol.~8, no.~6, pp. 4980--4990, 2021.

\bibitem{DBLP:conf/raid/MaLW0H0L22}
B.~Ma, X.~Lin, X.~Wang, B.~Liu, Y.~He, W.~Ni, and R.~P. Liu, ``New cloaking
  region obfuscation for road network-indistinguishability and location
  privacy,'' in \emph{Proc.{Int. Symp. Res. Attacks, Intrusions Defense
  (RAID'22)}, {\normalfont Limassol, Cyprus, October 26-28, 2022}}, pp.
  160--170.

\bibitem{DBLP:journals/access/XuWZC22}
J.~Xu, H.~Wang, J.~Zhang, and L.~Cai, ``Robust hand gesture recognition based
  on {RGB-D} data for natural human-computer interaction,'' \emph{{IEEE}
  Access}, vol.~10, pp. 54\,549--54\,562, 2022.

\bibitem{DBLP:conf/chi/WadleyKKSWCGHMS22}
G.~Wadley, V.~Kostakos, P.~Koval, W.~Smith, S.~Webber, A.~L. Cox, J.~J. Gross,
  K.~H{\"{o}}{\"{o}}k, R.~L. Mandryk, and P.~Slov{\'{a}}k, ``The future of
  emotion in human-computer interaction,'' in \emph{Proc. {ACM CHI Conf. Hum.
  Factors Comput. Syst. (CHI'22)}, {\normalfont New Orleans, LA, USA, April 29
  - May 5 2022}}, pp. 1--6.

\bibitem{DBLP:conf/hai/NalepkaCKKPR22}
P.~Nalepka, N.~Caruana, D.~M. Kaplan, R.~W. Kallen, E.~Pellicano, and M.~J.
  Richardson, ``Neurodiverse human-machine interaction and collaborative
  problem-solving in social {VR},'' in \emph{Proc. {Int. Conf. Hum.-Agent
  Interact. (HAI'22)}, {\normalfont Christchurch, New Zealand, December 5-8,
  2022}}, pp. 293--295.

\bibitem{DBLP:conf/ACMdis/Alves-OliveiraL21}
P.~A. Oliveira, M.~L. Lupetti, M.~Luria, D.~L{\"{o}}ffler, M.~Gamboa,
  L.~Albaugh, W.~Kamino, A.~K. Ostrowski, D.~Puljiz, P.~R. Cu{\'{e}}llar,
  M.~Scheunemann, M.~Suguitan, and D.~Lockton, ``Collection of metaphors for
  human-robot interaction,'' in \emph{Proc. {ACM Conf. Designing Interact.
  Syst. (DIS'21)}, {\normalfont Virtual Event, USA, 28 June, July 2, 2021}},
  pp. 1366--1379.

\bibitem{2022Credit}
A.~K. Nandi, K.~K. Randhawa, S.~C. Hong, M.~Seera, and C.~P. Lim, ``Credit card
  fraud detection using a hierarchical behavior-knowledge space model,''
  \emph{Public Lib. Sci.}, vol.~17, no.~1, pp. 1--16, 2022.

\bibitem{DBLP:conf/icbct/ZhangHZ21}
Y.~Zhang, M.~Huang, and X.~Zhang, ``Occupation mechanism for eliminating
  double-spending attacks on trusted transaction blockchain,'' in \emph{Proc.
  {Int. Conf. Blockchain Technol. (ICBCT'21) }, {\normalfont Shanghai, China,
  March 26-28, 2021}}, pp. 14--19.

\bibitem{2022Stealthy}
H.~Badrsimaei, R.~A. Hooshmand, and S.~Nobakhtian, ``Stealthy and profitable
  data injection attack on real time electricity market with network model
  uncertainties,'' \emph{Electr. Power Syst. Res.}, vol. 205, pp. 1--13, 2022.

\bibitem{DBLP:conf/icaif/ShearerBBW21}
M.~Shearer, D.~Byrd, T.~H. Balch, and M.~P. Wellman, ``Stability effects of
  arbitrage in exchange traded funds: {A}n agent-based model,'' in \emph{Proc.
  {ACM Int. Conf. AI Finance (ICAIF'21)}, {\normalfont Virtual Event, November
  3 - 5, 2021}}, pp. 1--9.

\bibitem{2020Managing}
C.~Gga, E.~Lgfc, C.~Cm, D.~Oc, E.~Lb, and C.~Edab, ``Managing a pool of rules
  for credit card fraud detection by a game theory based approach,''
  \emph{{Future Gener. Comput. Syst.}}, vol. 102, pp. 549--561, 2020.

\bibitem{DBLP:conf/pris2/ChenJT22}
R.~Chen, C.~H. Ju, and F.~S. Tu., ``A credit scoring ensemble framework using
  adaboost and multi-layer ensemble classification,'' in \emph{Proc. {Int.
  Conf. Pattern Recognit. Intell. Syst. (PRIS'22) }, {\normalfont Wuhan, China,
  July 29-31, 2022}}, pp. 72--79.

\bibitem{DBLP:journals/algorithms/KogaT22}
K.~Koga and K.~Takemoto, ``Simple black-box universal adversarial attacks on
  deep neural networks for medical image classification,'' \emph{Algorithms},
  vol.~15, no.~5, pp. 144--162, 2022.

\bibitem{DBLP:conf/niss/BengagBM21}
A.~Bengag, A.~Bengag, and O.~Moussaoui, ``Classification of security attacks in
  {WBAN} for medical healthcare,'' in \emph{Proc. {Int. Conf. Networking, Inf.
  Syst. Secur. (NISS'21)}, {\normalfont Kenitra, Morocco, April 1-2, 2021}},
  pp. 1--5.

\bibitem{DBLP:conf/aaai/Swenor22}
A.~Swenor, ``Using random perturbations to mitigate adversarial attacks on
  {NLP} models,'' in \emph{Proc. {Assoc. Adv. Artif. Intell. (AAAI'21) },
  {\normalfont Virtual Event, British Columbia, Canada, February 22 - March 1,
  2022}}, pp. 13\,142--13\,143.

\bibitem{9557814}
W.~Wang, R.~Wang, L.~Wang, Z.~Wang, and A.~Ye, ``Towards a robust deep neural
  network against adversarial texts: {A} survey,'' \emph{{IEEE} Trans. Knowl.
  Data Eng.}, vol.~35, no.~3, pp. 3159--3179, 2023.

\bibitem{DBLP:journals/access/TasoojiM22}
T.~K. Tasooji and H.~J. Marquez, ``A secure decentralized event-triggered
  cooperative localization in multi-robot systems under cyber attack,''
  \emph{{IEEE} Access}, vol.~10, pp. 128\,101--128\,121, 2022.

\bibitem{DBLP:conf/atal/Youssef21}
Y.~M. Youssef, ``Inducing rules about distributed robotic systems for fault
  detection {\&} diagnosis,'' in \emph{Proc. {Auton. Agents Multi-Agent Syst.
  (AAMAS'21)}, {\normalfont Virtual Event, United Kingdom, May 3-7, 2021}}, pp.
  1845--1847.

\bibitem{8804390}
V.~Santhanam and L.~S. Davis, ``A generic improvement to deep residual networks
  based on gradient flow,'' \emph{IEEE Trans. Neural Networks Learn. Syst.},
  vol.~31, no.~7, pp. 2490--2499, 2020.

\bibitem{9409715}
H.~Xu, M.~Yang, L.~Deng, Y.~Qian, and C.~Wang, ``Neutral cross-entropy loss
  based unsupervised domain adaptation for semantic segmentation,''
  \emph{{IEEE} Trans. Image Process.}, vol.~30, pp. 4516--4525, 2021.

\bibitem{9252132}
Y.~Zhong and W.~Deng, ``Towards transferable adversarial attack against deep
  face recognition,'' \emph{IEEE Trans. Inf. Forensics Secur.}, vol.~16, pp.
  1452--1466, 2021.

\bibitem{2019An}
S.~Makki, Z.~Assaghir, Y.~Taher, R.~Haque, and H.~Zeineddine, ``An experimental
  study with imbalanced classification approaches for credit card fraud
  detection,'' \emph{IEEE Access}, vol.~7, no.~99, pp. 93\,010--93\,022, 2019.

\bibitem{DBLP:conf/acmturc/ChenZLYMW19}
L.~Chen, Z.~Zhang, Q.~Liu, L.~Yang, Y.~Meng, and P.~Wang, ``A method for online
  transaction fraud detection based on individual behavior,'' in \emph{Proc.
  {ACM} {Turing Award Celebration Conf. China (TUR-C'19)}, {\normalfont
  Chengdu, China, May 17-19, 2019}}, pp. 1--8.

\bibitem{DBLP:journals/pvldb/CaoYCZLQ19}
S.~Cao, X.~Yang, C.~Chen, J.~Zhou, X.~Li, and Y.~Qi, ``Tit{A}nt: {O}nline
  real-time transaction fraud detection in ant financial,'' in \emph{Proc.
  {Int. Conf. Very Large Date Bases Endowment (VLDB'19)}, {\normalfont Los
  Angeles, USA, August 26-30, 2019}}, pp. 2082--2093.

\bibitem{DBLP:conf/mc/FeineMM20}
J.~Feine, S.~Morana, and A.~Maedche, ``A chatbot response generation system,''
  in \emph{Proc. Mensch und Comput. (MuC'20), {\normalfont Tagungsband,
  Magdebug, Germany, September 6-9, 2020}}, pp. 333--341.

\bibitem{DBLP:journals/eswa/BadueGCACFJBPMV21}
C.~Badue, R.~Guidolini, R.~V. Carneiro, P.~Azevedo, V.~B. Cardoso, A.~Forechi,
  L.~F.~R. Jesus, R.~F. Berriel, T.~M. Paix{\~{a}}o, F.~W. Mutz,
  L.~de~Paula~Veronese, T.~O. Santos, and A.~F.~D. Souza, ``Self-driving cars:
  {A} survey,'' \emph{Expert Syst. Appl.}, vol. 165, pp. 113\,816--113\,843,
  2021.

\bibitem{DBLP:conf/cscs2/ForsterBOS22}
D.~F{\"{o}}rster, T.~Bruckschl{\"{o}}gl, J.~L. Omer, and T.~Schipper,
  ``Challenges and directions for automated driving security,'' in \emph{Proc.
  {Int. Conf. Control Syst. Comput. Sci. (CSCS'22)}, {\normalfont Ingolstadt,
  Germany, December 8, 2022}}, pp. 1--11.

\bibitem{DBLP:journals/access/SalloumGVS22}
S.~A. Salloum, T.~Gaber, S.~Vadera, and K.~Shaalan, ``A systematic literature
  review on phishing email detection using natural language processing
  techniques,'' \emph{{IEEE} Access}, vol.~10, pp. 65\,703--65\,727, 2022.

\bibitem{DBLP:conf/codaspy/QachfarVM22}
F.~Z. Qachfar, R.~M. Verma, and A.~Mukherjee, ``Leveraging synthetic data and
  {PU} learning for phishing email detection,'' in \emph{Proc. {ACM Conf. Data
  Appl. Secur. Privacy (CODASPY'22)}, {\normalfont Baltimore, MD, USA, April 24
  - 27, 2022}}, pp. 29--40.

\bibitem{DBLP:journals/tcad/SunYRYH22}
Q.~Sun, X.~Yao, A.~A. Rao, B.~Yu, and S.~Hu, ``Counteracting adversarial
  attacks in autonomous driving,'' \emph{{IEEE} Trans. Comput. Aided Des.
  Integr. Circuits Syst.}, vol.~41, no.~12, pp. 5193--5206, 2022.

\bibitem{DBLP:conf/cui/SeymourCS22}
W.~Seymour, M.~Cot{\'{e}}, and J.~M. Such, ``When it's not worth the paper it's
  written on: {A} provocation on the certification of skills in the {A}lexa and
  {G}oogle assistant ecosystems,'' in \emph{Proc. {Conversational User
  Interfaces (CUI'22)}, {\normalfont Glasgow, United Kingdom, July 26 - 28,
  2022}}, pp. 1--5.

\bibitem{2018A}
Q.~Liu, P.~Li, W.~Zhao, W.~Cai, S.~Yu, and V.~C.~M. Leung, ``A survey on
  security threats and defensive techniques of machine learning: {A} data
  driven view,'' \emph{IEEE Access}, vol.~6, pp. 12\,103--12\,117, 2018.

\bibitem{DBLP:conf/icccv/DomyatiM22}
A.~Domyati and Q.~Memon, ``Robust detection of cardiac disease using machine
  learning algorithms: Robust detection,'' in \emph{Proc. {Int. Conf. Control
  Comput. Vision (ICCCV'22)}, {\normalfont Xiamen, China, August 19-21, 2022}},
  pp. 52--55.

\bibitem{DBLP:journals/tecs/MorrisEKIAS22}
J.~Morris, K.~Ergun, B.~Khaleghi, M.~Imani, B.~Aksanli, and T.~Simunic,
  ``Hydrea: {U}tilizing hyperdimensional computing for a more robust and
  efficient machine learning system,'' \emph{{ACM} Trans. Embedded Comput.
  Syst.}, vol.~21, no.~6, pp. 1--25, 2022.

\bibitem{DBLP:conf/ccs/PapernotMGJCS17}
N.~Papernot, P.~D. McDaniel, I.~J. Goodfellow, S.~Jha, Z.~B. Celik, and
  A.~Swami, ``Practical black-box attacks against machine learning,'' in
  \emph{Proc. {{ACM} Asia Conf. Comput. Commu. Secu. (AsiaCCS'17)},
  {\normalfont Abu Dhabi, United Arab Emirates, April 2-6, 2017}}, pp.
  506--519.

\bibitem{DBLP:journals/cybersec/SunTZ18}
L.~Sun, M.~Tan, and Z.~Zhou, ``A survey of practical adversarial example
  attacks,'' \emph{Cybersecur.}, vol.~1, no.~1, pp. 9--17, 2018.

\bibitem{DBLP:journals/access/AkhtarM18}
N.~Akhtar and A.~S. Mian, ``Threat of adversarial attacks on deep learning in
  computer vision: {A} survey,'' \emph{{IEEE} Access}, vol.~6, pp.
  14\,410--14\,430, 2018.

\bibitem{REN2020346}
K.~Ren, T.~Zheng, Z.~Qin, and X.~Liu, ``Adversarial attacks and defenses in
  deep learning,'' \emph{Eng.}, vol.~6, no.~3, pp. 346--360, 2020.

\bibitem{DBLP:journals/wicomm/KongXWHNL21}
Z.~Kong, J.~Xue, Y.~Wang, L.~Huang, Z.~Niu, and F.~Li, ``A survey on
  adversarial attack in the age of artificial intelligence,'' \emph{Wirel.
  Commun. Mob. Comput.}, vol. 2021, pp. 1--22, 2021.

\bibitem{app9050909}
S.~Qiu, Q.~Liu, S.~Zhou, and C.~Wu, ``Review of {A}rtificial {I}ntelligence
  adversarial attack and defense technologies,'' \emph{Appl. Sci.}, vol.~9,
  no.~5, pp. 909--939, 2019.

\bibitem{DBLP:journals/corr/abs-2209-14262}
D.~Wang, W.~Yao, T.~Jiang, G.~Tang, and X.~Chen, ``A survey on physical
  adversarial attack in computer vision,'' \emph{CoRR}, vol. abs/2209.14262,
  2022.

\bibitem{DBLP:journals/corr/abs-2211-01671}
X.~Wei, B.~Pu, J.~Lu, and B.~Wu, ``Physically adversarial attacks and defenses
  in computer vision: {A} survey,'' \emph{CoRR}, vol. abs/2211.01671, 2022.

\bibitem{DBLP:journals/pr/QianHWZ22}
Z.~Qian, K.~Huang, Q.~F. Wang, and X.~Y. Zhang, ``A survey of robust
  adversarial training in pattern recognition: {F}undamental, theory, and
  methodologies,'' \emph{Pattern Recognit.}, vol. 131, pp. 108\,889--108\,928,
  2022.

\bibitem{DBLP:journals/compsec/GallagherPCPMK22}
M.~Gallagher, N.~Pitropakis, C.~Chrysoulas, P.~Papadopoulos, A.~Mylonas, and
  S.~K. Katsikas, ``Investigating machine learning attacks on financial time
  series models,'' \emph{Comput. Secur.}, vol. 123, pp. 102\,933--102\,947,
  2022.

\bibitem{DBLP:conf/icaif/GoldblumSPG21}
M.~Goldblum, A.~Schwarzschild, A.~B. Patel, and T.~Goldstein, ``Adversarial
  attacks on machine learning systems for high-frequency trading,'' in
  \emph{Proc. {ACM Int. Conf. AI Finance (ICAIF'21)}, {\normalfont Virtual
  Event, November 3 - 5, 2021}}, pp. 2:1--2:9.

\bibitem{2017The}
A.~Kirilenko, A.~S. Kyle, M.~Samadi, and T.~Tuzun, ``The {F}lash {C}rash:
  {H}igh frequency trading in an electronic market,'' \emph{The Journal of
  Finance}, vol.~72, no.~3, pp. 967--998, 2017.

\bibitem{article}
H.~Liang, E.~He, Y.~Zhao, Z.~Jia, and H.~Li, ``Adversarial attack and defense:
  {A} survey,'' \emph{Electronics}, vol.~11, no.~8, pp. 1283--1302, 2022.

\bibitem{Amirkhani2022ASO}
A.~Amirkhani, M.~P. Karimi, and A.~Banitalebi-Dehkordi, ``A survey on
  adversarial attacks and defenses for object detection and their applications
  in autonomous vehicles,'' \emph{The Visual Computer}, pp. 1--15, 2022.

\bibitem{DBLP:conf/cvpr/CarliniF20}
N.~Carlini and H.~Farid, ``Evading {D}eepfake -- {I}mage detectors with white-
  and black-box attacks,'' in \emph{Proc. {IEEE Conf. Comput. Vis. Pattern
  Recognit. (CVPR'20)}, {\normalfont Seattle, WA, USA, June 14-19, 2020}}, pp.
  2804--2813.

\bibitem{DBLP:conf/naacl/XuZJL21}
Y.~Xu, X.~Zhong, A.~J. Yepes, and J.~H. Lau, ``Grey-box adversarial attack and
  defence for sentiment classification,'' in \emph{Proc. {N. Am. Chapter Assoc.
  Computat. Ling. (NAACL'21)}, {\normalfont Virtual Event, June 6-11, 2021}},
  pp. 4078--4087.

\bibitem{2015Basic}
W.~Ford, ``Basic iterative methods,'' \emph{Numer. Linear Algebra Appl.s}, pp.
  469--490, 2015.

\bibitem{DBLP:conf/cvpr/YuG021}
Y.~Yu, X.~Gao, and C.~Z. Xu, ``{LAFEAT:} piercing through adversarial defenses
  with latent features,'' in \emph{Proc. {IEEE Conf. Comput. Vis. Pattern
  Recognit. (CVPR'21)}, {\normalfont Virtual Event, Nashville, TN, USA, June
  19-25, 2021}}, pp. 5735--5745.

\bibitem{DBLP:journals/tip/CheBZLLTGC21}
Z.~Che, A.~Borji, G.~Zhai, S.~Ling, J.~Li, Y.~Tian, G.~Guo, and P.~L. Callet,
  ``Adversarial attack against deep saliency models powered by non-redundant
  priors,'' \emph{{IEEE} Trans. Image Process.}, vol.~30, pp. 1973--1988, 2021.

\bibitem{DBLP:conf/ijcnn/SchwinnNRZETB21}
L.~Schwinn, A.~Nguyen, R.~Raab, D.~Zanca, B.~M. Eskofier, D.~Tenbrinck, and
  M.~Burger, ``Dynamically sampled nonlocal gradients for stronger adversarial
  attacks,'' in \emph{Proc. {Int. Joint Conf. Neural Networks (IJCNN'21)},
  {\normalfont Shenzhen, China, July 18-22, 2021}}, pp. 1--8.

\bibitem{DBLP:journals/pami/ChenHSYH22}
S.~Chen, Z.~He, C.~Sun, J.~Yang, and X.~Huang, ``Universal adversarial attack
  on attention and the resulting dataset {DA}mage{N}et,'' \emph{{IEEE} Trans.
  Pattern Anal. Mach. Intell.}, vol.~44, no.~4, pp. 2188--2197, 2022.

\bibitem{DBLP:journals/tec/SuVS19}
J.~Su, D.~V. Vargas, and K.~Sakurai, ``One pixel attack for fooling deep neural
  networks,'' \emph{{IEEE} Int. Conf. Big Data Intell. Comput.}, vol.~23,
  no.~5, pp. 828--841, 2019.

\bibitem{DBLP:conf/sigir/Wen0L21}
Z.~Wen, Y.~Fang, and Z.~Liu, ``Meta-inductive node classification across
  graphs,'' in \emph{Proc. {Int. ACM SIGIR Conf. Res. Dev. Inf. Retr.
  (SIGIR'21)}, {\normalfont Virtual Event, Canada, July 11-15, 2021}}, pp.
  1219--1228.

\bibitem{DBLP:conf/jist/Ugai21}
T.~Ugai, ``Fuzzy search of knowledge graph with link prediction,'' in
  \emph{Proc. {Int. Joint Conf. Knowl. Graphs (IJCKG'21)}, {\normalfont Virtual
  Event, Thailand, December 6 - 8, 2021}}, pp. 121--125.

\bibitem{DBLP:journals/tcss/ChenLSL20}
J.~Chen, X.~Lin, Z.~Shi, and Y.~Liu, ``Link prediction adversarial attack via
  iterative gradient attack,'' \emph{{IEEE} Trans. Comput. Soc. Syst.}, vol.~7,
  no.~4, pp. 1081--1094, 2020.

\bibitem{DBLP:journals/tkde/LiXCXHZ23}
J.~Li, T.~Xie, L.~Chen, F.~Xie, X.~He, and Z.~Zheng, ``Adversarial attack on
  large scale graph,'' \emph{{IEEE} Trans. Knowl. Data Eng.}, vol.~35, no.~1,
  pp. 82--95, 2023.

\bibitem{DBLP:conf/icml/DaiLTHWZS18}
H.~Dai, H.~Li, T.~Tian, X.~Huang, L.~Wang, J.~Zhu, and L.~Song, ``Adversarial
  attack on graph structured data,'' in \emph{Proc. {Int. Conf. Mach. Learn.
  (ICML'18)}, {\normalfont Stockholmsm{\"{a}}ssan, Stockholm, Sweden, July
  10-15, 2018}}, pp. 1123--1132.

\bibitem{7876373}
J.~Liu, Y.~Li, G.~Ling, R.~Li, and Z.~Zheng, ``Community detection in
  location-based social networks: {A}n entropy-based approach,'' in \emph{Proc.
  {Int. Conf. Comput. Inf. Technol. (CIT'16)}, {\normalfont Nadi, Fiji,
  December 8-10, 2016}}, pp. 452--459.

\bibitem{DBLP:journals/datamine/WangLSLYZ20}
J.~Wang, M.~Luo, F.~Suya, J.~Li, Z.~Yang, and Q.~Zheng, ``Scalable attack on
  graph data by injecting vicious nodes,'' \emph{Data Min. Knowl. Discovery},
  vol.~34, no.~5, pp. 1363--1389, 2020.

\bibitem{9667076}
H.~Wang, S.~Wang, Z.~Jin, Y.~Wang, C.~Chen, and M.~Tistarelli,
  ``Similarity-based gray-box adversarial attack against deep face
  recognition,'' in \emph{Proc. {Int. Conf. Autom. Face Gesture Recognit.
  (FG'21)}, {\normalfont Jodhpur, India, December 15-18, 2021}}, pp. 1--8.

\bibitem{DBLP:journals/pami/WangLLL22}
H.~Wang, G.~Li, X.~Liu, and L.~Lin, ``A {H}amiltonian {M}onte {C}arlo method
  for probabilistic adversarial attack and learning,'' \emph{{IEEE} Trans.
  Pattern Anal. Mach. Intell.}, vol.~44, no.~4, pp. 1725--1737, 2022.

\bibitem{DBLP:journals/neco/Hinton02}
G.~E. Hinton, ``Training products of experts by minimizing contrastive
  divergence,'' \emph{Neural Comput.}, vol.~14, no.~8, pp. 1771--1800, 2002.

\bibitem{9149635}
Y.~Xiang, Y.~Xu, Y.~Li, W.~Ma, Q.~Xuan, and Y.~Liu, ``Side-channel gray-box
  attack for {DNN}s,'' \emph{{IEEE} Trans. Circuits Syst. II Express Briefs},
  vol.~68, no.~1, pp. 501--505, 2021.

\bibitem{10.1145/3511808.3557238}
Z.~Liu, Y.~Luo, L.~Wu, S.~Li, Z.~Liu, and S.~Z. Li, ``Are gradients on graph
  structure reliable in gray-box attacks?'' in \emph{Proc. {ACM Int. Conf. Inf.
  Knowl. Manage. (CIKM'22)}, {\normalfont Atlanta, GA, USA, 2022}}, pp.
  1360--–1368.

\bibitem{DBLP:conf/ijcnn/ChenCW21}
S.~Chen, Z.~Chen, and D.~Wang, ``Adaptive adversarial training for meta
  reinforcement learning,'' in \emph{Proc. {Int. Joint Conf. Neural Networks
  (IJCNN'21)}, {\normalfont Shenzhen, China, July 18-22, 2021}}, pp. 1--8.

\bibitem{DBLP:conf/cvpr/WuSLK21}
W.~Wu, Y.~Su, M.~R. Lyu, and I.~King, ``Improving the transferability of
  adversarial samples with adversarial transformations,'' in \emph{Proc. {IEEE
  Conf. Comput. Vis. Pattern Recognit. (CVPR'21)}, {\normalfont Virtual Event,
  Nashville, TN, USA, June 19-25, 2021}}, pp. 9024--9033.

\bibitem{DBLP:conf/cvpr/DongLPS0HL18}
Y.~Dong, F.~Liao, T.~Pang, H.~Su, J.~Zhu, X.~Hu, and J.~Li, ``Boosting
  adversarial attacks with momentum,'' in \emph{Proc. {IEEE Conf. Comput. Vis.
  Pattern Recognit. (CVPR'18)}, {\normalfont Salt Lake City, UT, USA, June
  18-22, 2018}}, pp. 9185--9193.

\bibitem{9609659}
Y.~Dong, S.~Cheng, T.~Pang, H.~Su, and J.~Zhu, ``Query-efficient black-box
  adversarial attacks guided by a transfer-based prior,'' \emph{{IEEE} Trans.
  Pattern Anal. Mach. Intell.}, vol.~44, no.~12, pp. 9536--9548, 2022.

\bibitem{DBLP:conf/cvpr/LuoL0WXS22}
C.~Luo, Q.~Lin, W.~Xie, B.~Wu, J.~Xie, and L.~Shen, ``Frequency-driven
  imperceptible adversarial attack on semantic similarity,'' in \emph{Proc.
  {IEEE Conf. Comput. Vis. Pattern Recognit. (CVPR'22)}, {\normalfont New
  Orleans, LA, USA, June 18-24, 2022}}, pp. 15\,315--15\,324.

\bibitem{DBLP:journals/tifs/BonnetFB22}
B.~Bonnet, T.~Furon, and P.~Bas, ``Generating adversarial images in quantized
  domains,'' \emph{{IEEE} Trans. Inf. Forensics Secur.}, vol.~17, pp. 373--385,
  2022.

\bibitem{9720115}
H.~Chang, Y.~Rong, T.~Xu, W.~Huang, H.~Zhang, P.~Cui, X.~Wang, W.~Zhu, and
  J.~Huang, ``Adversarial attack framework on graph embedding models with
  limited knowledge,'' \emph{{IEEE} Trans. Knowl. Data Eng.}, vol.~14, pp.
  1--14, 2022.

\bibitem{2021Hypergraph}
S.~Bai, F.~Zhang, and P.~Torr, ``Hypergraph convolution and hypergraph
  attention,'' \emph{Pattern Recognit.}, vol. 110, no.~6, pp.
  107\,637--107\,644, 2021.

\bibitem{DBLP:journals/tsipn/ChenTTLKH19}
P.~Y. Chen, C.~C. Tu, P.~S. Ting, Y.~Y. Lo, D.~Koutra, and A.~O.~H. III,
  ``Identifying influential links for event propagation on {T}witter: {A}
  network of networks approach,'' \emph{{IEEE} Trans. Signal Inf. Process.
  Networks}, vol.~5, no.~1, pp. 139--151, 2019.

\bibitem{DBLP:conf/ijcnn/FengFXWHX21}
S.~Feng, F.~Feng, X.~Xu, Z.~Wang, Y.~Hu, and L.~Xie, ``Digital watermark
  perturbation for adversarial examples to fool deep neural {N}etworks,'' in
  \emph{Proc. {Int. Joint Conf. Neural Networks (IJCNN'21)}, {\normalfont
  Shenzhen, China, July 18-22, 2021}}, pp. 1--8.

\bibitem{9970367}
N.~Aafaq, N.~Akhtar, W.~Liu, M.~Shah, and A.~Mian, ``Language model agnostic
  gray-box adversarial attack on image captioning,'' \emph{{IEEE} Trans. Inf.
  Forensics Secur.}, vol.~18, pp. 626--638, 2023.

\bibitem{DBLP:journals/tifs/LiQHQLD21}
Q.~Li, Y.~Qi, Q.~Hu, S.~Qi, Y.~Lin, and J.~S. Dong, ``Adversarial adaptive
  neighborhood with feature importance-aware convex interpolation,''
  \emph{{IEEE} Trans. Inf. Forensics Secur.}, vol.~16, pp. 2447--2460, 2021.

\bibitem{DBLP:journals/access/AriasNGTV22}
F.~X. Arias, M.~Z. N{\'{u}}{\~{n}}ez, A.~G. Adames, N.~T. Flores, and M.~V.
  Lombardo, ``Sentiment analysis of public social media as a tool for
  health-related topics,'' \emph{{IEEE} Access}, vol.~10, pp. 74\,850--74\,872,
  2022.

\bibitem{DBLP:conf/cvpr/DuanM00QY20}
R.~Duan, X.~Ma, Y.~Wang, J.~Bailey, A.~K. Qin, and Y.~Yang, ``Adversarial
  camouflage: {H}iding physical-world attacks with natural styles,'' in
  \emph{Proc. {IEEE Conf. Comput. Vis. Pattern Recognit. (CVPR'20)},
  {\normalfont Seattle, WA, USA, June 13-19, 2020}}, pp. 997--1005.

\bibitem{DBLP:journals/tcyb/YangLDT20}
E.~Yang, T.~Liu, C.~Deng, and D.~Tao, ``Adversarial examples for hamming space
  search,'' \emph{{IEEE} Trans. Cybern.}, vol.~50, no.~4, pp. 1473--1484, 2020.

\bibitem{9526915}
M.~Liu, L.~Chen, X.~Du, L.~Jin, and M.~Shang, ``Activated gradients for deep
  neural networks,'' \emph{IEEE Trans. Neural Networks Learn. Syst.}, pp.
  1--13, 2021 (early access).

\bibitem{DBLP:conf/cvpr/JiaS0Y21}
S.~Jia, Y.~Song, C.~Ma, and X.~Yang, ``Io{U} attack: Towards temporally
  coherent black-box adversarial attack for visual object tracking,'' in
  \emph{Proc. {IEEE Conf. Comput. Vis. Pattern Recognit. (CVPR'21)},
  {\normalfont Virtual Event, Nashville, TN, USA, June 19-25, 2021}}, pp.
  6709--6718.

\bibitem{DBLP:journals/tip/ZhangTLWT20}
Y.~Zhang, X.~Tian, Y.~Li, X.~Wang, and D.~Tao, ``Principal component
  adversarial example,'' \emph{{IEEE} Trans. Image Process.}, vol.~29, pp.
  4804--4815, 2020.

\bibitem{DBLP:conf/iccv/DuanCNYQH21}
R.~Duan, Y.~Chen, D.~Niu, Y.~Yang, A.~K. Qin, and Y.~He, ``Adv{D}rop:
  Adversarial attack to {DNN}s by dropping information,'' in \emph{Proc. {IEEE
  Int. Conf. Comput. Vision (ICCV'21)}, {\normalfont Montreal, QC, Canada,
  October 10-17, 2021}}, pp. 7486--7495.

\bibitem{DBLP:conf/icml/AthalyeC018}
A.~Athalye, N.~Carlini, and D.~A. Wagner, ``Obfuscated gradients give a false
  sense of security: {C}ircumventing defenses to adversarial examples,'' in
  \emph{Proc. {Int. Conf. Mach. Learn. (ICML'18)}, {\normalfont
  Stockholmsm{\"{a}}ssan, Stockholm, Sweden, July 10-15, 2018}}, vol.~80, pp.
  274--283.

\bibitem{DBLP:conf/ndss/Xu0Q18}
W.~Xu, D.~Evans, and Y.~Qi, ``Feature squeezing: {D}etecting adversarial
  examples in deep neural networks,'' in \emph{Proc. {Network Distrib. Syst.
  Symp. (NDSS'18)}, {\normalfont San Diego, California, USA, February 18-21,
  2018}}, pp. 1--15.

\bibitem{DBLP:conf/eccv/ChengLCTCLZ22}
Z.~Cheng, J.~Liang, H.~Choi, G.~Tao, Z.~Cao, D.~Liu, and X.~Zhang, ``Physical
  attack on monocular depth estimation with optimal adversarial patches,'' in
  \emph{Proc. {Eur. Conf. Comput. Vision (ECCV'22)}, {\normalfont Tel Aviv,
  Israel, October 23-27, 2022}}, vol. 13698, pp. 514--532.

\bibitem{DBLP:journals/tifs/ShenYZXLH21}
M.~Shen, H.~Yu, L.~Zhu, K.~Xu, Q.~Li, and J.~Hu, ``Effective and robust
  physical-world attacks on deep learning face recognition systems,''
  \emph{{IEEE} Trans. Inf. Forensics Secur.}, vol.~16, pp. 4063--4077, 2021.

\bibitem{DBLP:journals/pami/WeiGY23}
X.~Wei, Y.~Guo, and J.~Yu, ``Adversarial sticker: {A} stealthy attack method in
  the physical world,'' \emph{{IEEE} Trans. Pattern Anal. Mach. Intell.},
  vol.~45, no.~3, pp. 2711--2725, 2023.

\bibitem{DBLP:conf/cvpr/EykholtEF0RXPKS18}
K.~Eykholt, I.~Evtimov, E.~Fernandes, B.~Li, A.~Rahmati, C.~Xiao, A.~Prakash,
  T.~Kohno, and D.~Song, ``Robust physical-world attacks on deep learning
  visual classification,'' in \emph{Proc. {IEEE Conf. Comput. Vis. Pattern
  Recognit. (CVPR'18)}, {\normalfont Salt Lake City, UT, USA, June 18-22,
  2018}}, pp. 1625--1634.

\bibitem{DBLP:conf/aaai/LiuLFMZXT19}
A.~Liu, X.~Liu, J.~Fan, Y.~Ma, A.~Zhang, H.~Xie, and D.~Tao,
  ``Perceptual-sensitive {GAN} for generating adversarial patches,'' in
  \emph{Proc. {Assoc. Adv. Artif. Intell. (AAAI'19) }, {\normalfont Honolulu,
  Hawaii, USA, January 27 - February 1, 2019}}, pp. 1028--1035.

\bibitem{DBLP:conf/iscas/0250CWGHZW021}
W.~Wang, Y.~Chai, Z.~Wu, L.~Ge, X.~Han, B.~Zhang, C.~Wang, and Y.~Li,
  ``Generating adversarial patches using data-driven {M}ulti{D}-{WGAN},'' in
  \emph{Proc. {IEEE Int. Symp. Circuits Syst. (ISCAS'21)}, {\normalfont Daegu,
  South Korea, May 22-28, 2021}}, pp. 1--5.

\bibitem{DBLP:journals/tip/WangLBL22}
J.~Wang, A.~Liu, X.~Bai, and X.~Liu, ``Universal adversarial patch attack for
  automatic checkout using perceptual and attentional bias,'' \emph{{IEEE}
  Trans. Image Process.}, vol.~31, pp. 598--611, 2022.

\bibitem{DBLP:conf/iccv/RutaMF0JFGC21}
D.~Ruta, S.~Motiian, B.~Faieta, Z.~Lin, H.~Jin, A.~Filipkowski, A.~Gilbert, and
  J.~P. Collomosse, ``{ALADIN:} {A}ll layer adaptive instance normalization for
  fine-grained style similarity,'' in \emph{Proc. {IEEE Int. Conf. Comput.
  Vision (ICCV'21)}, {\normalfont Montreal, QC, Canada, October 10-17, 2021}},
  pp. 11\,906--11\,915.

\bibitem{DBLP:journals/iotj/BaiLZ22}
T.~Bai, J.~Luo, and J.~Zhao, ``Inconspicuous adversarial patches for fooling
  image-recognition systems on mobile devices,'' \emph{{IEEE} Internet Things
  J.}, vol.~9, no.~12, pp. 9515--9524, 2022.

\bibitem{DBLP:conf/eccv/LiuWLCZY20}
A.~Liu, J.~Wang, X.~Liu, B.~Cao, C.~Zhang, and H.~Yu, ``Bias-based universal
  adversarial patch attack for automatic check-out,'' in \emph{Proc. {Eur.
  Conf. Comput. Vision (ECCV'20)}, {\normalfont Glasgow, UK, August 23-28,
  2020}}, vol. 12358, pp. 395--410.

\bibitem{DBLP:conf/iccv/YuCXLWBM21}
C.~Yu, J.~Chen, Y.~Xue, Y.~Liu, W.~Wan, J.~Bao, and H.~Ma, ``Defending against
  universal adversarial patches by clipping feature norms,'' in \emph{Proc.
  {IEEE Int. Conf. Comput. Vision (ICCV'21)}, {\normalfont Montreal, QC,
  Canada, October 10-17, 2021}}, pp. 16\,414--16\,422.

\bibitem{DBLP:conf/icassp/KimLR22}
T.~Kim, H.~J. Lee, and Y.~M. Ro, ``Map: Multispectral adversarial patch to
  attack person detection,'' in \emph{Proc. {Int. Conf. Acoust. Speech Signal
  Process. (ICASSP'22)}, {\normalfont Virtual Event, Singapore, 23-27 May
  2022}}, pp. 4853--4857.

\bibitem{DBLP:conf/cw/TarchounAKM21}
B.~Tarchoun, I.~Alouani, A.~B. Khalifa, and M.~A. Mahjoub, ``Adversarial
  attacks in a multi-view setting: {An} empirical study of the adversarial
  patches inter-view transferability,'' in \emph{Proc. {Int. Conf. Cyberworlds
  (CW'21)}, {\normalfont Caen, France, September 28-30, 2021}}, pp. 299--302.

\bibitem{DBLP:conf/eccv/SuZCYCG18}
D.~Su, H.~Zhang, H.~Chen, J.~Yi, P.~Y. Chen, and Y.~Gao, ``Is robustness the
  cost of accuracy? - {A} comprehensive study on the robustness of 18 deep
  image classification models,'' in \emph{Proc. {Eur. Conf. Comput. Vision
  ({ECCV}'18)}, {\normalfont Munich, Germany, September 8-14, 2018}}, vol.
  11216, pp. 644--661.

\bibitem{DBLP:journals/corr/SimonyanZ14a}
K.~Simonyan and A.~Zisserman, ``Very deep convolutional networks for
  large-scale image recognition,'' in \emph{Proc. {Int. Conf. Learn.
  Representations (ICLR'15)}, {\normalfont San Diego, CA, USA, May 7-9, 2015}},
  pp. 1--14.

\bibitem{DBLP:conf/emnlp/WallaceSS20}
E.~Wallace, M.~Stern, and D.~Song, ``Imitation attacks and defenses for
  black-box machine translation systems,'' in \emph{Proc. {Conf. Empirical
  Methods Nat. Lang. Process. (EMNLP'20)}, {\normalfont Virtual Event, Punta
  Cana, Dominican Republic, November 16-20, 2020}}, pp. 5531--5546.

\bibitem{9157681}
H.~Zheng, Z.~Zhang, J.~Gu, H.~Lee, and A.~Prakash, ``Efficient adversarial
  training with transferable adversarial examples,'' in \emph{Proc. {IEEE Conf.
  Comput. Vis. Pattern Recognit. (CVPR'20)}}, Los Alamitos, CA, USA, June 2020,
  pp. 1178--1187.

\bibitem{9762031}
Y.~Wang, Y.~Tan, T.~Baker, N.~Kumar, and Q.~Zhang, ``Deep fusion: {C}rafting
  transferable adversarial examples and improving robustness of industrial
  artificial intelligence of things,'' \emph{{IEEE} Trans. Ind. Inf.}, pp.
  1--10, 2022 (early access).

\bibitem{9612007}
Y.~Zhou, M.~Kantarcioglu, and B.~Xi, ``Exploring the effect of randomness on
  transferability of adversarial samples against deep neural networks,''
  \emph{{IEEE} {Trans. Dependable Secure Comput.}}, vol.~20, no.~1, pp. 83--99,
  2023.

\bibitem{DBLP:journals/neco/HochreiterS97}
S.~Hochreiter and J.~Schmidhuber, ``Long short-term memory,'' \emph{Neural
  Comput.}, vol.~9, no.~8, pp. 1735--1780, 1997.

\bibitem{DBLP:conf/ijcnn/BernhardMD21}
R.~Bernhard, P.~A. Mo{\"{e}}llic, and J.~M. Dutertre, ``Luring transferable
  adversarial perturbations for deep neural networks,'' in \emph{Proc. {Int.
  Joint Conf. Neural Networks (IJCNN'21)}, {\normalfont Shenzhen, China, July
  18-22, 2021}}, pp. 1--8.

\bibitem{DBLP:conf/iccv/ZiZMJ21}
B.~Zi, S.~Zhao, X.~Ma, and Y.~G. Jiang, ``Revisiting adversarial robustness
  distillation: {R}obust soft labels make student better,'' in \emph{Proc.
  {IEEE Int. Conf. Comput. Vision (ICCV'21)}, {\normalfont Montreal, QC,
  Canada, October 10-17, 2021}}, pp. 16\,423--16\,432.

\bibitem{DBLP:conf/ijcnn/YanLDBX21}
X.~Yan, Y.~Li, T.~Dai, Y.~Bai, and S.~T. Xia, ``D2{D}efend: {D}ual-domain based
  defense against adversarial examples,'' in \emph{Proc. {Int. Joint Conf.
  Neural Networks (IJCNN'21)}, {\normalfont Shenzhen, China, July 18-22,
  2021}}, pp. 1--8.

\bibitem{DBLP:conf/ispacs/EndoT19}
H.~Endo and A.~Taguchi, ``Color image enhancement by using hue-saturation
  gradient,'' in \emph{Proc. {Int. Symp. Intell. Signal Process. Commun. Syst.
  (ISPACS'19)}, {\normalfont Taipei, Taiwan, December 3-6, 2019}}, pp. 1--2.

\bibitem{DBLP:conf/codaspy/LiuKK21}
G.~Liu, I.~Khalil, and A.~Khreishah, ``Using single-step adversarial training
  to defend iterative adversarial examples,'' in \emph{Proc. {ACM Conf. Data
  Appl. Secur. Privacy (CODASPY'21)}, {\normalfont Virtual Event, USA, April
  26-28, 2021}}, pp. 17--27.

\bibitem{DBLP:conf/cvpr/HeRF19}
Z.~He, A.~S. Rakin, and D.~Fan, ``Parametric noise injection: {T}rainable
  randomness to improve deep neural network robustness against adversarial
  attack,'' in \emph{{Proc. IEEE Conf. Comput. Vis. Pattern Recognit.
  (CVPR'19)}, {\normalfont Long Beach, CA, USA, June 16-20, 2019}}, pp.
  588--597.

\bibitem{DBLP:journals/csur/CaiXXWLP21}
Z.~Cai, Z.~Xiong, H.~Xu, P.~Wang, W.~Li, and Y.~Pan, ``Generative adversarial
  networks: {A} survey toward private and secure applications,'' \emph{{ACM}
  Comput. Surv.}, vol.~54, no.~6, pp. 1--38, 2022.

\bibitem{9552529}
X.~Du and C.~M. Pun, ``Robust audio patch attacks using physical sample
  simulation and adversarial patch noise generation,'' \emph{{IEEE} Trans.
  Multimedia}, vol.~24, pp. 4381--4393, 2022.

\bibitem{9797884}
J.~Deng, L.~Dong, R.~Wang, R.~Yang, and D.~Yan, ``Decision-based attack to
  speaker recognition system via local low-frequency perturbation,''
  \emph{{IEEE} Signal Process. Lett.}, vol.~29, pp. 1432--1436, 2022.

\bibitem{9599559}
Y.~Ding, H.~Lin, L.~Liu, Z.~Ling, and Y.~Hu, ``Robustness of speech spoofing
  detectors against adversarial post-processing of voice conversion,''
  \emph{{IEEE/ACM} Trans. Audio, Speech, Lang. Process.}, vol.~29, pp.
  3415--3426, 2021.

\bibitem{9627642}
S.~Liu, N.~Lu, C.~Chen, and K.~Tang, ``Efficient combinatorial optimization for
  word-level adversarial textual attack,'' \emph{{IEEE/ACM} Trans. Audio,
  Speech, Lang. Process.}, vol.~30, pp. 98--111, 2022.

\bibitem{9696006}
B.~R. Manoj, M.~Sadeghi, and E.~G. Larsson, ``Downlink power allocation in
  massive {MIMO} via deep learning: {A}dversarial attacks and training,''
  \emph{IEEE Trans. Cognit. Commun. Networking}, vol.~8, no.~2, pp. 707--719,
  2022.

\bibitem{9570812}
P.~M. Santos, B.~R. Manoj, M.~Sadeghi, and E.~G. Larsson, ``Universal
  adversarial attacks on neural networks for power allocation in a massive
  {MIMO} system,'' \emph{IEEE Wirel. Commun. Lett.}, vol.~11, no.~1, pp.
  67--71, 2022.

\bibitem{9814883}
Y.~Guo, Q.~Li, W.~Zuo, and H.~Chen, ``An intermediate-level attack framework on
  the basis of linear regression,'' \emph{IEEE Trans. Pattern Anal. Mach.
  Intell.}, vol.~45, no.~3, pp. 2726--2735, 2023.

\bibitem{DBLP:journals/tkde/FengHTC21}
F.~Feng, X.~He, J.~Tang, and T.~S. Chua, ``Graph adversarial training:
  Dynamically regularizing based on graph structure,'' \emph{{IEEE} Trans.
  Knowl. Data Eng.}, vol.~33, no.~6, pp. 2493--2504, 2021.

\bibitem{DBLP:journals/tmm/AminiG20}
S.~Amini and S.~Ghaemmaghami, ``Towards improving robustness of deep neural
  networks to adversarial perturbations,'' \emph{{IEEE} Trans. Multim.},
  vol.~22, no.~7, pp. 1889--1903, 2020.

\bibitem{DBLP:conf/iclr/HendrycksG17}
D.~Hendrycks and K.~Gimpel, ``A baseline for detecting misclassified and
  out-of-distribution examples in neural networks,'' in \emph{Proc. {Int. Conf.
  Learn. Representations (ICLR'17)}, {\normalfont Toulon, France, April 24-26,
  2017}}, pp. 1--12.

\bibitem{DBLP:journals/tcyb/LiLY20a}
H.~Li, G.~Li, and Y.~Yu, ``{ROSA: R}obust salient object detection against
  adversarial attacks,'' \emph{{IEEE} Trans. Cybern.}, vol.~50, no.~11, pp.
  4835--4847, 2020.

\bibitem{9449325}
K.~Han, Y.~Li, and B.~Xia, ``A cascade model-aware generative adversarial
  example detection method,'' \emph{Tsinghua Sci. Technol.}, vol.~26, no.~6,
  pp. 800--812, 2021.

\bibitem{DBLP:conf/ictai/NgoWKPAL19}
C.~P. Ngo, A.~A. Winarto, C.~K.~L. Kou, S.~Park, F.~Akram, and H.~K. Lee,
  ``Fence {GAN: T}owards better anomaly detection,'' in \emph{Proc. {Int. Conf.
  Tools Artif. Intell. (ICTAI'21)}, {\normalfont Portland, OR, USA, November
  4-6, 2019}}, pp. 141--148.

\bibitem{DBLP:journals/access/XieYWCL21}
C.~Xie, K.~Yang, A.~Wang, C.~Chen, and W.~Li, ``A mura detection method based
  on an improved generative adversarial network,'' \emph{{IEEE} Access},
  vol.~9, pp. 68\,826--68\,836, 2021.

\bibitem{9825039}
X.~Zhang and Y.~Wang, ``{ADN}et: {A} neural network model for adversarial
  example detection based on steganalysis and attention mechanism,'' in
  \emph{Proc. {Int. Conf. Comput. Vision, Image Deep Learning (CVIDL'22)},
  {\normalfont San Diego, CA, USA, October 30, 2021 - November 3, 2021}}, pp.
  55--60.

\bibitem{9378303}
S.~Freitas, S.~T. Chen, Z.~J. Wang, and D.~H. Chau, ``Un{M}ask: {A}dversarial
  detection and defense through robust feature alignment,'' in
  \emph{Proc.{IEEE} Int. Conf. Big Data (Big Data'20), {\normalfont Atlanta,
  GA, USA, December 10-13, 2020}}, pp. 1081--1088.

\bibitem{9506292}
Y.~Wang, L.~Xie, X.~Liu, J.~Yin, and T.~Zheng, ``Model-agnostic adversarial
  example detection through logit distribution learning,'' in \emph{Proc. {IEEE
  Int. Conf. Image Process. (ICIP'21)}}, 2021, pp. 3617--3621.

\bibitem{9052913}
M.~Esmaeilpour, P.~Cardinal, and A.~L. Koerich, ``Detection of adversarial
  attacks and characterization of adversarial subspace,'' in \emph{Proc. {Int.
  Conf. Acoust. Speech Signal Process. (ICASSP'20)}, {\normalfont Barcelona,
  Spain, May 4-8, 2020}}, pp. 3097--3101.

\bibitem{DBLP:conf/cacml/LiuWD22}
H.~Liu, F.~Wang, and J.~Du, ``Research on adversarial attack technology for
  object detection in physical world based on vision,'' in \emph{Proc. {Asia
  Conf. Algorithms, Comput. Mach. Learn. (CACM'22)}, {\normalfont Hangzhou,
  China, March 25-27, 2022}}, pp. 638--648.

\bibitem{DBLP:journals/access/Wang22e}
S.~Wang, ``Joint learning of discriminative metric space from multi-context
  visual scene for unsupervised salient object detection,'' \emph{{IEEE}
  Access}, vol.~10, pp. 126\,089--126\,099, 2022.

\bibitem{DBLP:conf/iclr/SongKNEK18}
Y.~Song, T.~Kim, S.~Nowozin, S.~Ermon, and N.~Kushman, ``Pixeldefend:
  {L}everaging generative models to understand and defend against adversarial
  examples,'' in \emph{Proc. {Int. Conf. Learn. Representations (ICLR'18)},
  {\normalfont Vancouver, BC, Canada, April 30 - May 3, 2018}}, pp. 1--20.

\bibitem{DBLP:conf/cvpr/LiuLLXLWW19}
Z.~Liu, Q.~Liu, T.~Liu, N.~Xu, X.~Lin, Y.~Wang, and W.~Wen, ``Feature
  distillation: {DNN}-oriented {JPEG} compression against adversarial
  examples,'' in \emph{Proc. {IEEE Conf. Comput. Vis. Pattern Recognit.
  (CVPR'19)}, {\normalfont Long Beach, CA, USA, June 16-20, 2019}}, pp.
  860--868.

\bibitem{DBLP:conf/sp/LingJZWWLW19}
X.~Ling, S.~Ji, J.~Zou, J.~Wang, C.~Wu, B.~Li, and T.~Wang, ``{DEEPSEC:} {A}
  uniform platform for security analysis of deep learning model,'' in
  \emph{Proc. {IEEE} {Symp. Secur. Privacy (SP'19)}, {\normalfont San
  Francisco, CA, USA, May 19-23, 2019}}, pp. 673--690.

\bibitem{DBLP:conf/ijcnn/KangTCK21}
M.~Kang, T.~Q. Tran, S.~J. Cho, and D.~Kim, ``{CAP-GAN: T}owards adversarial
  robustness with cycle-consistent attentional purification,'' in \emph{Proc.
  {Int. Joint Conf. Neural Networks (IJCNN'21)}, {\normalfont Shenzhen, China,
  July 18-22, 2021}}, pp. 1--8.

\bibitem{DBLP:journals/corr/DziugaiteGR16}
G.~K. Dziugaite, Z.~Ghahramani, and D.~M. Roy, ``A study of the effect of {JPG}
  compression on adversarial images,'' \emph{CoRR}, vol. abs/1608.00853, 2016.

\bibitem{DBLP:conf/iccv/Zhou0P0WYL21}
D.~Zhou, N.~Wang, C.~Peng, X.~Gao, X.~Wang, J.~Yu, and T.~Liu, ``Removing
  adversarial noise in class activation feature space,'' in \emph{Proc. {IEEE
  Int. Conf. Comput. Vision (ICCV'21)}, {\normalfont Montreal, QC, Canada,
  October 10-17, 2021}}, pp. 7858--7867.

\bibitem{DBLP:conf/icassp/JinSZDZ19}
G.~Jin, S.~Shen, D.~Zhang, F.~Dai, and Y.~Zhang, ``{APE-{GAN}: A}dversarial
  perturbation elimination with {{GAN}},'' in \emph{Proc. {Int. Conf. Acoust.
  Speech Signal Process. (ICASSP'19)}, {\normalfont Brighton, United Kingdom,
  May 12-17, 2019}}, pp. 3842--3846.

\bibitem{DBLP:journals/tdsc/WeiL21}
W.~Wei and L.~Liu, ``Robust deep learning ensemble against deception,''
  \emph{{IEEE} Trans. Dependable Secur. Comput.}, vol.~18, no.~4, pp.
  1513--1527, 2021.

\bibitem{DBLP:conf/nips/BastaniILVNC16}
O.~Bastani, Y.~Ioannou, L.~Lampropoulos, D.~Vytiniotis, A.~V. Nori, and
  A.~Criminisi, ``Measuring neural net robustness with constraints,'' in
  \emph{Proc. {Conf. Workshop Neural Inf. Process. Syst. (NIPS'16)},
  {\normalfont Barcelona, Spain, December 5-10, 2016}}, pp. 2613--2621.

\bibitem{DBLP:conf/icml/EtmannLMS19}
C.~Etmann, S.~Lunz, P.~Maass, and C.~Sch{\"{o}}nlieb, ``On the connection
  between adversarial robustness and saliency map interpretability,'' in
  \emph{Proc. {Int. Conf. Mach. Learn. (ICML'19)}, {\normalfont Long Beach,
  California, {USA}, June 9-15 2019}}, vol.~97, pp. 1823--1832.

\bibitem{DBLP:conf/nips/JordanLD19}
M.~Jordan, J.~Lewis, and A.~G. Dimakis, ``Provable certificates for adversarial
  examples: {F}itting a ball in the union of polytopes,'' in \emph{Proc. {Annu.
  Conf. Neural Inf. Process. Syst. (NeurIPS'19)}, {\normalfont Vancouver, BC,
  Canada, December 8-14, 2019, }}, pp. 14\,059--14\,069.

\bibitem{DBLP:conf/ijcnn/ZhuWZ21}
Y.~Zhu, X.~Wei, and Y.~Zhu, ``Efficient adversarial defense without adversarial
  training: {A} batch normalization approach,'' in \emph{Proc. {Int. Joint
  Conf. Neural Networks (IJCNN'21)}, {\normalfont Shenzhen, China, July 18-22,
  2021}}, pp. 1--8.

\bibitem{DBLP:journals/tip/MustafaKHSS20}
A.~Mustafa, S.~H. Khan, M.~Hayat, J.~Shen, and L.~Shao, ``Image
  super-resolution as a defense against adversarial attacks,'' \emph{{IEEE}
  Trans. Image Process.}, vol.~29, pp. 1711--1724, 2020.

\bibitem{DBLP:journals/corr/abs-2207-08089}
T.~Blau, R.~{GAN}z, B.~Kawar, A.~M. Bronstein, and M.~Elad, ``Threat
  model-agnostic adversarial defense using diffusion models,'' \emph{CoRR},
  vol. abs/2207.08089, 2022.

\bibitem{DBLP:conf/ipccc/QinY21}
Y.~Qin and C.~Yue, ``Key-based input transformation defense against adversarial
  examples,'' in \emph{Proc. {Int. Perform. Comput. Commun. Conf. (IPCCC'21)},
  {\normalfont Austin, TX, USA, October 29-31, 2021}}, pp. 1--10.

\bibitem{DBLP:journals/jota/MarchiT22}
A.~D. Marchi and A.~Themelis, ``Proximal gradient algorithms under local
  {L}ipschitz gradient continuity,'' \emph{J. Optim. Theory Appl.}, vol. 194,
  no.~3, pp. 771--794, 2022.

\bibitem{DBLP:journals/ijcv/AnirudhTKB20}
R.~Anirudh, J.~J. Thiagarajan, B.~Kailkhura, and P.~T. Bremer, ``Mimic{GAN}:
  Robust projection onto image manifolds with corruption mimicking,''
  \emph{Int. J. Comput. Vis.}, vol. 128, no.~10, pp. 2459--2477, 2020.

\bibitem{DBLP:journals/nca/XiaoZYWZW22}
J.~Xiao, S.~Zhang, Y.~Yao, Z.~Wang, Y.~Zhang, and Y.~F. Wang, ``Generative
  adversarial network with hybrid attention and compromised normalization for
  multi-scene image conversion,'' \emph{Neural Comput. Appl.}, vol.~34, no.~9,
  pp. 7209--7225, 2022.

\bibitem{DBLP:journals/corr/SmilkovTKVW17}
D.~Smilkov, N.~Thorat, B.~Kim, F.~B. Vi{\'{e}}gas, and M.~Wattenberg,
  ``Smooth{G}rad: {R}emoving noise by adding noise,'' \emph{CoRR}, vol.
  abs/1706.03825, 2017.

\bibitem{DBLP:journals/pami/YanGZ21}
Z.~Yan, Y.~Guo, and C.~Zhang, ``Adversarial margin maximization networks,''
  \emph{{IEEE} Trans. Pattern Anal. Mach. Intell.}, vol.~43, no.~4, pp.
  1129--1139, 2021.

\bibitem{9533868}
C.~Yang, L.~Ding, Y.~Chen, and H.~Li, ``Defending against {GAN}-based
  {D}eep{F}ake attacks via transformation-aware adversarial faces,'' in
  \emph{Proc. {Int. Joint Conf. Neural Networks (IJCNN'21)}, {\normalfont
  Shenzhen, China, July 18-22, 2021}}, pp. 1--8.

\bibitem{DBLP:journals/tip/YuLLYZ21}
H.~Yu, A.~Liu, G.~Li, J.~Yang, and C.~Zhang, ``Progressive diversified
  augmentation for general robustness of {DNN}s: {A} unified approach,''
  \emph{{IEEE} Trans. Image Process.}, vol.~30, pp. 8955--8967, 2021.

\bibitem{2021HIRE}
S.~Kundu, M.~Pedram, and P.~A. Beerel, ``{HIRE-SNN}: {H}arnessing the
  adversarial robustness of energy-efficient deep spiking neural networks via
  training with crafted input noise,'' in \emph{Proc. {IEEE Int. Conf. Comput.
  Vision (ICCV'21)}, {\normalfont Montreal, QC, Canada, October 10-17, 2021}},
  pp. 5189--5198.

\bibitem{DBLP:conf/ijcnn/XuWP21}
Z.~Xu, J.~Wang, and J.~Pu, ``Defense against adversarial attacks with an
  induced class,'' in \emph{Proc. {Int. Joint Conf. Neural Networks
  (IJCNN'21)}, {\normalfont Shenzhen, China, July 18-22, 2021}}, pp. 1--8.

\bibitem{DBLP:conf/ijcnn/YaoHHP21}
R.~Yao, C.~Huang, Z.~Hu, and K.~Pei, ``Adaptive retraining for neural network
  robustness in classification,'' in \emph{Proc. {Int. Joint Conf. Neural
  Networks (IJCNN'21)}, {\normalfont Shenzhen, China, July 18-22, 2021}}, pp.
  1--8.

\bibitem{DBLP:conf/cvpr/GongRY021}
C.~Gong, T.~Ren, M.~Ye, and Q.~Liu, ``Max{U}p: {L}ightweight adversarial
  training with data augmentation improves neural network training,'' in
  \emph{Proc. {IEEE Conf. Comput. Vis. Pattern Recognit. (CVPR'21)},
  {\normalfont Virtual Event, Nashville, TN, USA, June 19-25, 2021}}, pp.
  2474--2483.

\bibitem{2021TriATNE}
Q.~Liu, C.~Long, J.~Zhang, M.~Xu, and P.~Lv, ``Tri{ATNE}: {T}ripartite
  adversarial training for network embeddings,'' \emph{{IEEE} Trans. Cybern.},
  vol.~52, no.~9, pp. 9634--9645, 2022.

\bibitem{DBLP:conf/iccv/PoursaeedJYBL21}
O.~Poursaeed, T.~Jiang, H.~Yang, S.~J. Belongie, and S.~N. Lim, ``Robustness
  and generalization via generative adversarial training,'' in \emph{Proc.
  {IEEE Int. Conf. Comput. Vision (ICCV'21)}, {\normalfont Montreal, QC,
  Canada, October 10-17, 2021}}, pp. 15\,691--15\,700.

\bibitem{DBLP:conf/cvpr/HongCK21}
M.~Hong, J.~Choi, and G.~Kim, ``{StyleMix: S}eparating content and style for
  enhanced data augmentation,'' in \emph{Proc. {IEEE Conf. Comput. Vis. Pattern
  Recognit. (CVPR'21)}, {\normalfont Virtual Event, Nashville, TN, USA, June
  19-25, 2021}}, pp. 14\,862--14\,870.

\bibitem{DBLP:conf/iccv/YunHCOYC19}
S.~Yun, D.~Han, S.~Chun, S.~J. Oh, Y.~Yoo, and J.~Choe, ``{CutMix:
  R}egularization strategy to train strong classifiers with localizable
  features,'' in \emph{Proc. {IEEE Int. Conf. Comput. Vision (ICCV'19)},
  {\normalfont Seoul, South Korea, October 27 - November 2, 2019}}, pp.
  6022--6031.

\bibitem{9712207}
H.~Shen, S.~Chen, R.~Wang, and X.~Wang, ``Adversarial learning with
  cost-sensitive classes,'' \emph{IEEE Trans. Cybern.}, pp. 1--12, 2022 (early
  access).

\bibitem{DBLP:conf/iccv/ChenPM0D021}
G.~Chen, P.~Peng, L.~Ma, J.~Li, L.~Du, and Y.~Tian, ``Amplitude-phase
  recombination: Rethinking robustness of convolutional neural networks in
  frequency domain,'' in \emph{Proc. {IEEE Int. Conf. Comput. Vision
  (ICCV'21)}, {\normalfont Montreal, QC, Canada, October 10-17, 2021}}, pp.
  448--457.

\bibitem{DBLP:conf/iccv/SunLXQKL021}
M.~Sun, Z.~Li, C.~Xiao, H.~Qiu, B.~Kailkhura, M.~Liu, and B.~Li, ``Can shape
  structure features improve model robustness under diverse adversarial
  settings?'' in \emph{Proc. {IEEE Int. Conf. Comput. Vision (ICCV'21)},
  {\normalfont Montreal, QC, Canada, October 10-17, 2021}}, pp. 7506--7515.

\bibitem{DBLP:journals/tip/LiuLYZLT21}
A.~Liu, X.~Liu, H.~Yu, C.~Zhang, Q.~Liu, and D.~Tao, ``Training robust deep
  neural networks via adversarial noise propagation,'' \emph{{IEEE} Trans.
  Image Process.}, vol.~30, pp. 5769--5781, 2021.

\bibitem{DBLP:journals/tip/ZhangLLXYML21}
C.~Zhang, A.~Liu, X.~Liu, Y.~Xu, H.~Yu, Y.~Ma, and T.~Li, ``Interpreting and
  improving adversarial robustness of deep neural networks with neuron
  sensitivity,'' \emph{{IEEE} Trans. Image Process.}, vol.~30, pp. 1291--1304,
  2021.

\bibitem{DBLP:conf/ijcnn/SchwartzD21}
D.~M. Schwartz and G.~Ditzler, ``Bolstering adversarial robustness with latent
  disparity regularization,'' in \emph{Proc. {Int. Joint Conf. Neural Networks
  (IJCNN'21)}, {\normalfont Shenzhen, China, July 18-22, 2021}}, pp. 1--8.

\bibitem{DBLP:conf/ijcnn/JiangMEB21}
Y.~Jiang, X.~Ma, S.~M. Erfani, and J.~Bailey, ``Dual head adversarial
  training,'' in \emph{Proc. {Int. Joint Conf. Neural Networks (IJCNN'21)},
  {\normalfont Shenzhen, China, July 18-22, 2021}}, pp. 1--8.

\bibitem{DBLP:conf/iccv/LiMLYKWH21}
Y.~Li, M.~R. Min, T.~C.~M. Lee, W.~Yu, E.~Kruus, W.~Wang, and C.~J. Hsieh,
  ``Towards robustness of deep neural networks via regularization,'' in
  \emph{Proc. {IEEE Int. Conf. Comput. Vision (ICCV'21)}, {\normalfont
  Montreal, QC, Canada, October 10-17, 2021}}, pp. 7476--7485.

\bibitem{DBLP:conf/ijcnn/CarboneSB21}
G.~Carbone, G.~Sanguinetti, and L.~Bortolussi, ``Random projections for
  improved adversarial robustness,'' in \emph{Proc. {Int. Joint Conf. Neural
  Networks (IJCNN'21)}, {\normalfont Shenzhen, China, July 18-22, 2021}}, pp.
  1--7.

\bibitem{2012Manifold}
C.~R. Genovese, M.~Perone-Pacifico, and V.~L. Wasserman, ``Manifold estimation
  and singular deconvolution under {H}ausdorff loss,'' \emph{Ann. Stat.},
  vol.~40, pp. 941--963, 2012.

\bibitem{DBLP:conf/stoc/DasguptaF08}
S.~Dasgupta and Y.~Freund, ``Random projection trees and low dimensional
  manifolds,'' in \emph{Proc. {ACM Symp. Theory Comput. (STOC'08)},
  {\normalfont Victoria, British Columbia, Canada, May 17-20, 2008}}, pp.
  537--546.

\bibitem{DBLP:conf/iccv/LiuW00S21}
H.~Liu, H.~Wu, W.~Xie, F.~Liu, and L.~Shen, ``Group-wise inhibition based
  feature regularization for robust classification,'' in \emph{Proc. {IEEE Int.
  Conf. Comput. Vision (ICCV'21)}, {\normalfont Montreal, QC, Canada, October
  10-17, 2021}}, pp. 468--476.

\bibitem{DBLP:journals/tip/XuDZ21}
Y.~Xu, B.~Du, and L.~Zhang, ``Self-attention context network: {A}ddressing the
  threat of adversarial attacks for hyperspectral image classification,''
  \emph{{IEEE} Trans. Image Process.}, vol.~30, pp. 8671--8685, 2021.

\bibitem{DBLP:journals/tip/LiSGL21}
Q.~Li, L.~Shen, S.~Guo, and Z.~Lai, ``Wave{C}{N}et: {W}avelet integrated {CNNs}
  to suppress aliasing effect for noise-robust image classification,''
  \emph{{IEEE} Trans. Image Process.}, vol.~30, pp. 7074--7089, 2021.

\bibitem{DBLP:conf/iccv/MokNCY21}
J.~Mok, B.~Na, H.~Choe, and S.~Yoon, ``Adv{R}ush: {S}earching for adversarially
  robust neural architectures,'' in \emph{Proc. {IEEE Int. Conf. Comput. Vision
  (ICCV'21)}, {\normalfont Montreal, QC, Canada, October 10-17, 2021}}, pp.
  12\,302--12\,312.

\bibitem{9711018}
T.~Yeo, O.~Kar, and A.~Zamir, ``Robustness via cross-domain ensembles,'' in
  \emph{Proc. {IEEE Int. Conf. Comput. Vision (ICCV'21)}}, Los Alamitos, CA,
  USA, 2021, pp. 12\,169--12\,179.

\bibitem{DBLP:conf/nips/HeinA17}
M.~Hein and M.~Andriushchenko, ``Formal guarantees on the robustness of a
  classifier against adversarial manipulation,'' in \emph{Proc. {Conf. Workshop
  Neural Inf. Process. Syst. (NIPS'17)}, {\normalfont Long Beach, CA, {USA},
  December 4-9, 2017}}, pp. 2266--2276.

\bibitem{DBLP:conf/cvpr/SerrurierMGBLB21}
M.~Serrurier, F.~Mamalet, A.~G. Sanz, T.~Boissin, J.~M. Loubes, and E.~del
  Barrio, ``Achieving robustness in classification using optimal transport with
  hinge regularization,'' in \emph{Proc. {IEEE Conf. Comput. Vis. Pattern
  Recognit. (CVPR'21)}, {\normalfont Virtual Event, Nashville, TN, USA, June
  19-25, 2021}}, pp. 505--514.

\bibitem{DBLP:journals/corr/GuR14}
S.~Gu and L.~Rigazio, ``Towards deep neural network architectures robust to
  adversarial examples,'' in \emph{Proc. {Int. Conf. Learn. Representations
  (ICLR'15)}, {\normalfont San Diego, CA, USA, May 7-9, 2015}}, pp. 1--9.

\bibitem{DBLP:conf/aaai/LiLWZWL22}
G.~Li, X.~Li, Y.~Wang, S.~Zhang, Y.~Wu, and D.~Liang, ``Knowledge distillation
  for object detection via rank mimicking and prediction-guided feature
  imitation,'' in \emph{Proc. {Assoc. Adv. Artif. Intell. (AAAI'22), }
  {\normalfont Virtual Event, Palo Alto, California USA, February 22 - March 1,
  2022}}, pp. 1306--1313.

\bibitem{DBLP:conf/iccv/WangDYLL21}
H.~Wang, Y.~Deng, S.~Yoo, H.~Ling, and Y.~Lin, ``{AGKD-BML: D}efense against
  adversarial attack by attention guided knowledge distillation and
  bi-directional metric learning,'' in \emph{Proc. {IEEE Int. Conf. Comput.
  Vision (ICCV'21)}, {\normalfont Montreal, QC, Canada, October 10-17, 2021}},
  pp. 7638--7647.

\bibitem{DBLP:conf/nips/MaoZYVR19}
C.~Mao, Z.~Zhong, J.~Yang, C.~Vondrick, and B.~Ray, ``Metric learning for
  adversarial robustness,'' in \emph{Proc. {Annu. Conf. Neural Inf. Process.
  Syst. (NeurIPS'19)}, {\normalfont Vancouver, BC, Canada, December 8-14,
  2019}}, pp. 478--489.

\bibitem{DBLP:conf/iccv/WangZ19}
J.~Wang and H.~Zhang, ``Bilateral adversarial training: {T}owards fast training
  of more robust models against adversarial attacks,'' in \emph{Proc. {IEEE
  Int. Conf. Comput. Vision (ICCV'19)}, {\normalfont Seoul, South Korea,
  October 27 - November 2, 2019}}, pp. 6628--6637.

\bibitem{DBLP:conf/nips/ZhangW19}
H.~Zhang and J.~Wang, ``Defense against adversarial attacks using feature
  scattering-based adversarial training,'' in \emph{Proc. {Annu. Conf. Neural
  Inf. Process. Syst. (NeurIPS'19)}, {\normalfont Vancouver, BC, Canada,
  December 8-14, 2019}}, pp. 1829--1839.

\bibitem{9693248}
B.~Wu, S.~Wang, X.~Yuan, C.~Wang, C.~Rudolph, and X.~Yang, ``Defeating
  misclassification attacks against transfer learning,'' \emph{{IEEE} Trans.
  Dependable Secure Comput.}, no.~01, pp. 1--16, 2022 (early access).

\bibitem{DBLP:conf/stoc/AlonGHM21}
N.~Alon, A.~Gonen, E.~Hazan, and S.~Moran, ``Boosting simple learners,'' in
  \emph{Proc. {ACM Symp. Theory Comput. (STOC'21)}, {\normalfont Virtual Event,
  Italy, June 21-25, 2021}}, pp. 481--489.

\bibitem{DBLP:conf/ic3/KumarS22}
R.~Kumar and G.~Subbiah, ``Explainable machine learning for malware detection
  using ensemble bagging algorithms,'' in \emph{Proc. {Int. Conf. Contemp.
  Comput. (IC3'22)}, {\normalfont Noida, India, August 4-6, 2022}}, pp.
  453--460.

\bibitem{DBLP:journals/pami/MustafaKHGSS21}
A.~Mustafa, S.~H. Khan, M.~Hayat, R.~Goecke, J.~Shen, and L.~Shao, ``Deeply
  supervised discriminative learning for adversarial defense,'' \emph{{IEEE}
  Trans. Pattern Anal. Mach. Intell.}, vol.~43, no.~9, pp. 3154--3166, 2021.

\bibitem{DBLP:conf/bmvc/ShafahiGNHDG19}
A.~Shafahi, A.~Ghiasi, M.~Najibi, F.~Huang, J.~P. Dickerson, and T.~Goldstein,
  ``Batch-wise logit-similarity: {G}eneralizing logit-squeezing and
  label-smoothing,'' in \emph{Proc. {British Mach. Vision Conf. (BMVC'19)},
  {\normalfont Cardiff, UK, September 9-12, 2019}}, pp. 72--83.

\bibitem{DBLP:conf/ijcnn/KanaiYYTI21}
S.~Kanai, M.~Yamada, S.~Yamaguchi, H.~Takahashi, and Y.~Ida, ``Constraining
  logits by bounded function for adversarial robustness,'' in \emph{Proc. {Int.
  Joint Conf. Neural Networks (IJCNN'21)}, {\normalfont Shenzhen, China, July
  18-22, 2021}}, pp. 1--8.

\bibitem{liu2021training}
C.~Liu, M.~Salzmann, and S.~S{\"u}sstrunk, ``Training provably robust models by
  polyhedral envelope regularization,'' \emph{IEEE Trans. Neural Networks
  Learn. Syst.}, pp. 1--15, 2021 (early access).

\bibitem{DBLP:conf/iclr/BalunovicV20}
M.~Balunovic and M.~T. Vechev, ``Adversarial training and provable defenses:
  {B}ridging the gap,'' in \emph{Proc. {Int. Conf. Learn. Representations
  (ICLR'20)}, {\normalfont Addis Ababa, Ethiopia, April 26-30, 2020}}, pp.
  1--19.

\bibitem{DBLP:journals/access/LeeBY21}
H.~Lee, H.~Bae, and S.~Yoon, ``Gradient masking of label smoothing in
  adversarial robustness,'' \emph{{IEEE} Access}, vol.~9, pp. 6453--6464, 2021.

\bibitem{DBLP:conf/iccv/HowardPALSCWCTC19}
A.~Howard, R.~Pang, H.~Adam, Q.~V. Le, M.~Sandler, B.~Chen, W.~Wang, L.~C.
  Chen, M.~Tan, G.~Chu, V.~Vasudevan, and Y.~Zhu, ``Searching for
  {M}obile{N}et{V}3,'' in \emph{Proc. {IEEE Int. Conf. Comput. Vision
  (ICCV'19)}, {\normalfont Seoul, South Korea, October 27 - November 2, 2019}},
  pp. 1314--1324.

\bibitem{DBLP:journals/ral/YunLL21}
P.~Yun, Y.~Liu, and M.~Liu, ``In defense of knowledge distillation for task
  incremental learning and its application in {3D} object detection,''
  \emph{{IEEE} Robotics Autom. Lett.}, vol.~6, no.~2, pp. 2012--2019, 2021.

\bibitem{2021396}
X.~Liu, H.~Jiang, X.~Su, and J.~Feng, ``Adversarial examples generation
  algorithm based on decision boundary search,'' \emph{{J. Electron. Sci.
  Technol.}}, vol.~51, pp. 721--727, 2022.

\bibitem{2000Nonlinear}
E.~Wegert and M.~A. Efendiev, ``Nonlinear {R}iemann—{H}ilbert problems with
  {L}ipschitz continuous boundary condition,'' \emph{{Rensselaer Soc. Eng.}},
  vol. 130, no.~04, pp. 793--800, 2000.

\bibitem{DBLP:conf/nips/ChenWH20}
X.~Chen, Z.~S. Wu, and M.~Hong, ``Understanding gradient clipping in private
  {SGD:} {A} geometric perspective,'' in \emph{Proc. {Annu. Conf. Neural Inf.
  Process. Syst. (NeurIPS'20)}, {\normalfont Virtual Event, December 6-12,
  2020}}, pp. 13\,773--13\,782.

\bibitem{DBLP:conf/nips/LobachevaKCMV21}
E.~Lobacheva, M.~Kodryan, N.~Chirkova, A.~Malinin, and D.~P. Vetrov, ``On the
  periodic behavior of neural network training with batch normalization and
  weight decay,'' in \emph{Proc. {Annu. Conf. Neural Inf. Process. Syst.
  (NeurIPS'21)}, {\normalfont Virtual Event, December 6-14, 2021}}, pp.
  21\,545--21\,556.

\bibitem{DBLP:journals/tie/DuCSWFK20}
Y.~Du, W.~Cao, J.~She, M.~Wu, M.~Fang, and S.~Kawata, ``Disturbance rejection
  and control system design using improved equivalent input disturbance
  approach,'' \emph{{IEEE} Trans. Ind. Electron.}, vol.~67, no.~4, pp.
  3013--3023, 2020.

\bibitem{DBLP:journals/access/DhilleswararaoB22}
P.~Dhilleswararao, S.~Boppu, M.~S. Manikandan, and L.~R. Cenkeramaddi,
  ``Efficient hardware architectures for accelerating deep neural networks:
  {S}urvey,'' \emph{{IEEE} Access}, vol.~10, pp. 131\,788--131\,828, 2022.

\bibitem{1315}
T.~Zhang, K.~Yang, and J.~Wei, ``Survey on detecting and defending adversarial
  examples for image data,'' \emph{J. Comput. Res. Dev.}, vol.~59, no.~6, pp.
  1315--1328, 2022.

\bibitem{DBLP:conf/iwcmc/UsamaALQA19}
M.~Usama, M.~Asim, S.~Latif, J.~Qadir, and A.~I.~A. Fuqaha, ``Generative
  adversarial networks for launching and thwarting adversarial attacks on
  network intrusion detection systems,'' in \emph{Proc. {Wirel. Commun. Mobile
  Comput. (IWCMC'19)}, {\normalfont Tangier, Morocco, June 24-28, 2019}}, pp.
  78--83.

\end{thebibliography}
\end{document}